%% file: main.tex
\documentclass[runningheads,orivec]{llncs}
\usepackage[table,dvipsnames]{xcolor}
\usepackage[super]{nth}
\usepackage[T1]{fontenc}
\usepackage{array}
\usepackage{multicol}
\usepackage{multirow}
\usepackage{amsmath}
\usepackage{booktabs}
\usepackage{rotating}
\usepackage{hyperref}
\usepackage{textcomp}
\usepackage{subcaption}
\usepackage{mathtools}
\usepackage{amssymb}
\usepackage{tabularx}
\usepackage{fontawesome5}
\usepackage{tikz}
\usepackage{graphicx}
\usepackage{siunitx}
\usepackage{cleveref}
\usepackage{pgfplots}
\usepackage{contour}
\usepackage{enumitem}
\usepackage{cleveref}
\usetikzlibrary{positioning,shapes,backgrounds,fit,calc,decorations.pathmorphing}
\input{common_styles}

\newcolumntype{C}[1]{>{\centering\arraybackslash}m{#1}}
\renewcommand\fbox{\fcolorbox{gray!50}{white}}
\newcommand{\gtext}[1]{\textcolor{gray}{#1}}
\newcommand{\raiseandrotate}[2]{\raisebox{#1\height}[0pt][0pt]{\rotatebox{90}{#2}}}

\pgfplotsset{compat=1.18}
\begin{document}
\title{Segmenting France Across Four Centuries}
\titlerunning{Segmenting France Across Four Centuries}
\author{Marta López-Rauhut\inst{1,2} \and Hongyu Zhou\inst{3,4} \and Mathieu Aubry\inst{1} \and Loic Landrieu\inst{1}}
\authorrunning{M. López-Rauhut et al.}
\institute{LIGM, ENPC, IP Paris, Univ Gustave Eiffel, CNRS, Marne-la-Vallée, France \\ \and Université Paris-Saclay, CNRS, LISN, Orsay, France\\ \and University of Bonn, Germany \\ \and Lamarr Institute, Germany}

\maketitle

\begin{abstract}
Historical maps offer an invaluable perspective into territory evolution across past centuries---long before satellite or remote sensing technologies existed. Deep learning methods have shown promising results in segmenting historical maps, but publicly available datasets typically focus on a single map type or period, require extensive and costly annotations, and are not suited for nationwide, long-term analyses.

In this paper, we introduce a new dataset of historical maps tailored for analyzing large-scale, long-term land use and land cover evolution with limited annotations. Spanning metropolitan France (548{,}305 km\textsuperscript{2}), our dataset contains three map collections from the \nth{18}, \nth{19}, and \nth{20} centuries. We provide both comprehensive modern labels and 22{,}878 km\textsuperscript{2} of manually annotated historical labels for the \nth{18} and \nth{19} century maps. Our dataset illustrates the complexity of the segmentation task, featuring stylistic inconsistencies, interpretive ambiguities, and significant landscape changes (e.g., marshlands disappearing in favor of forests).

We assess the difficulty of these challenges by benchmarking three approaches: a fully-supervised model trained with historical labels, and two weakly-supervised models that rely only on modern annotations. The latter either use the modern labels directly or first perform image-to-image translation to address the stylistic gap between historical and contemporary maps. Finally, we discuss how these methods can support long-term environment monitoring, offering insights into centuries of landscape transformation. Our official project repository is publicly available on \href{https://github.com/Archiel19/FRAx4.git}{GitHub}.

\end{abstract}

\input{1_introduction}
\input{2_related_work}
\input{3_dataset}
\input{4_method}
\input{5_experiments}
\input{6_conclusion}
\input{7_acknowledgments}

\bibliographystyle{splncs04}
\bibliography{references}

\end{document}

%% file: common_styles.tex
\definecolor{unet_color}{HTML}{E65651}
\definecolor{cyclegan_color}{HTML}{4BC1D1}
\definecolor{nice_orange}{HTML}{E39152}
\definecolor{nice_purple}{HTML}{BB52E3}
\definecolor{nice_brown}{HTML}{635C56}
\definecolor{nice_blue}{HTML}{4BC1D1}

\def\interdist{0.22}
\def\secondcol{3.8}

\def\mapsperblock{4}
\def\opacityscalefactor{1.15}
\def\depthscalefactor{-0.35}

\tikzset{
    pics/MapBlock/.style args={#1/#2}{code={
        \foreach \X in {#2,...,1}{
            \pgfmathsetmacro{\opacity}{1/(\opacityscalefactor^(\X-1))}
            \pgfmathsetmacro{\depth}{\depthscalefactor*(\X-1)}
            \node[draw,thick,opacity=\opacity] (_map_\X) at (0,0,\depth){\includegraphics[width=0.9cm]{cass\X_#1.jpeg}};
        }}
    },
    loss/.style={
        circle,very thick,draw=black!70,fill=white,minimum size=8mm
    },
    map_labels/.style={
        text width=2cm,align=flush center,font=\scriptsize
    },
    unet/.pic={
        \node [shape=isosceles triangle,draw=#1,fill=#1!60,very thick,minimum height=0.35cm,isosceles triangle apex angle=90] at (0,0) {};
        \node [shape=isosceles triangle,draw=#1,fill=#1!60,very thick,minimum height=0.35cm,isosceles triangle apex angle=90,shape border rotate=180] at (0.33cm,0) {};
    },
    cyclegan/.pic={
	\node (left) [shape=isosceles triangle,draw=cyclegan_color,fill=cyclegan_color!60,very thick,minimum height=0.35cm,isosceles triangle apex angle=90] at (0,0) {};
	\node (right) [shape=isosceles triangle,draw=cyclegan_color,fill=cyclegan_color!60,very thick,minimum height=0.35cm,isosceles triangle apex angle=90,shape border rotate=180] at (0.33cm,0) {};
	\node [fit=(left)(right)] (_net) {};
	\node [above=0.1 of _net] {\textcolor{cyclegan_color}{\faRedo}};
	\node [below=0.2 of _net,font=\scriptsize] {\textcolor{cyclegan_color}{\textbf{CycleGAN}}};
    }
}

\newcommand{\road}{\textcolor[HTML]{e9894a}{\faRoad}}

\newcommand{\water}{\textcolor[HTML]{31c1ec}{\faWater}}

\newcommand{\tree}{\textcolor[HTML]{99ec53}{\faTree}}

\newcommand{\building}{\textcolor[HTML]{ea0029}{\faBuilding}}

\newcommand{\tracks}[1]{
    \resizebox{#1}{!}{
        \begin{tikzpicture}
            \draw[line width=4mm] (-1.5,0) -- (-0.75,3); % 
            \draw[line width=4mm] (1.5,0) -- (0.75,3);  % 
            
            \foreach \y/\scale in {0.5/2.0, 1.5/1.6, 2.5/1.3} {
                \draw[line width=3mm] (-\scale,\y) -- (\scale,\y);
            }
        \end{tikzpicture}
    }
}

\newcommand{\bof}{{\contour{black}{$\sim$}}}

\newcommand{\yes}{\textcolor{ForestGreen!70}{\faCheck}}

\newcommand{\no}{\textcolor{red!75!black!60}{\faTimes}}

\newcommand{\req}{\footnotesize{\faEnvelope[regular]} \normalsize}

\newcommand{\auth}{\footnotesize {\faHandshake[regular]} \normalsize}

\crefname{section}{sec.}{secs.}
\Crefname{section}{Sec.}{Secs.}
\crefname{table}{tab.}{tabs.}
\Crefname{table}{Tab.}{Tabs.}
\crefname{figure}{fig.}{figs.}
\Crefname{figure}{Fig.}{Figs}

%% file: 1_introduction.tex
\section{Introduction}

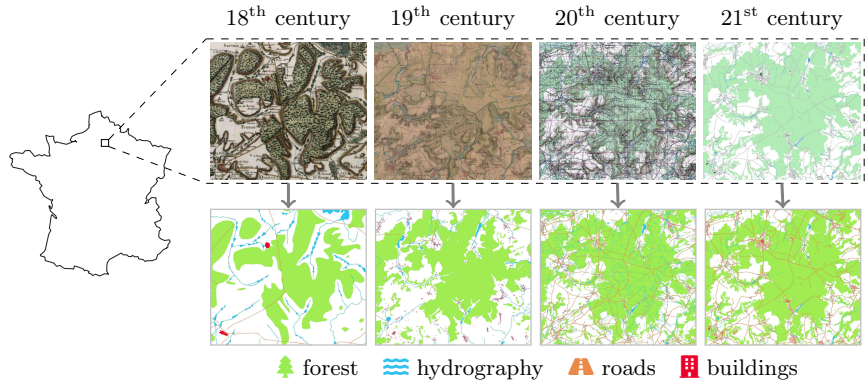
\begin{figure}
    \centering
    \resizebox{0.95\columnwidth}{!}{\input{teaser}}
    \caption{\textbf{Semantic Segmentation of Historical Maps of France Spanning Four Centuries.} \textit{Top row}: aligned input maps of the same area. The \nth{21}-century map is a vectorized present-day map. \textit{Bottom row}: semantic segmentation of each input map into four classes.}
    \label{fig:teaser}
\end{figure}

Historical maps are invaluable for studying long-term changes in landscapes, infrastructure, and land use. By analyzing cartographic data spanning centuries, we can quantify transformations such as deforestation, urbanization, and the evolution of transportation networks. However, direct comparisons across historical maps are hindered by stylistic variations, inconsistencies in drawing conventions, and the lack of standardized annotations. This underscores the need for automatic methods capable of extracting meaningful semantic information from heterogeneous historical maps.

In this work, we introduce a large-scale dataset for historical map analysis, covering four centuries of cartography over metropolitan France, {as shown in \cref{fig:teaser}}. Our dataset provides historical map scans alongside modern land cover annotations and sparse historical annotations. This enables comparing supervised approaches, trained with historical labels, and weakly-supervised approaches, trained with modern labels.

To establish a benchmark for future research, we propose three baseline approaches. The first applies direct segmentation using sparse historical labels. The second employs a weakly-supervised framework, training a model to segment historical maps using modern land cover labels. The third incorporates a CycleGAN-based image-to-image translation module to translate historical maps into a modern cartographic style before segmentation, reducing stylistic discrepancies.

Our results highlight the strengths and limitations of these approaches, demonstrating their effectiveness in segmenting historical maps while illustrating the broader challenges of the task. We further show how our models can be leveraged to analyze long-term land cover changes across centuries at a national scale. By releasing this benchmark dataset and systematically evaluating multiple baselines, we provide a foundation for future research in automated historical map analysis, facilitating the study of environmental and urban transformations.

%% file: teaser.tex
\begin{tikzpicture}[inner sep=0pt, label distance=5]

    % Maps
    \node (cass) [label=above:\nth{18} century] at (0,0) {\includegraphics[width=2.2cm]{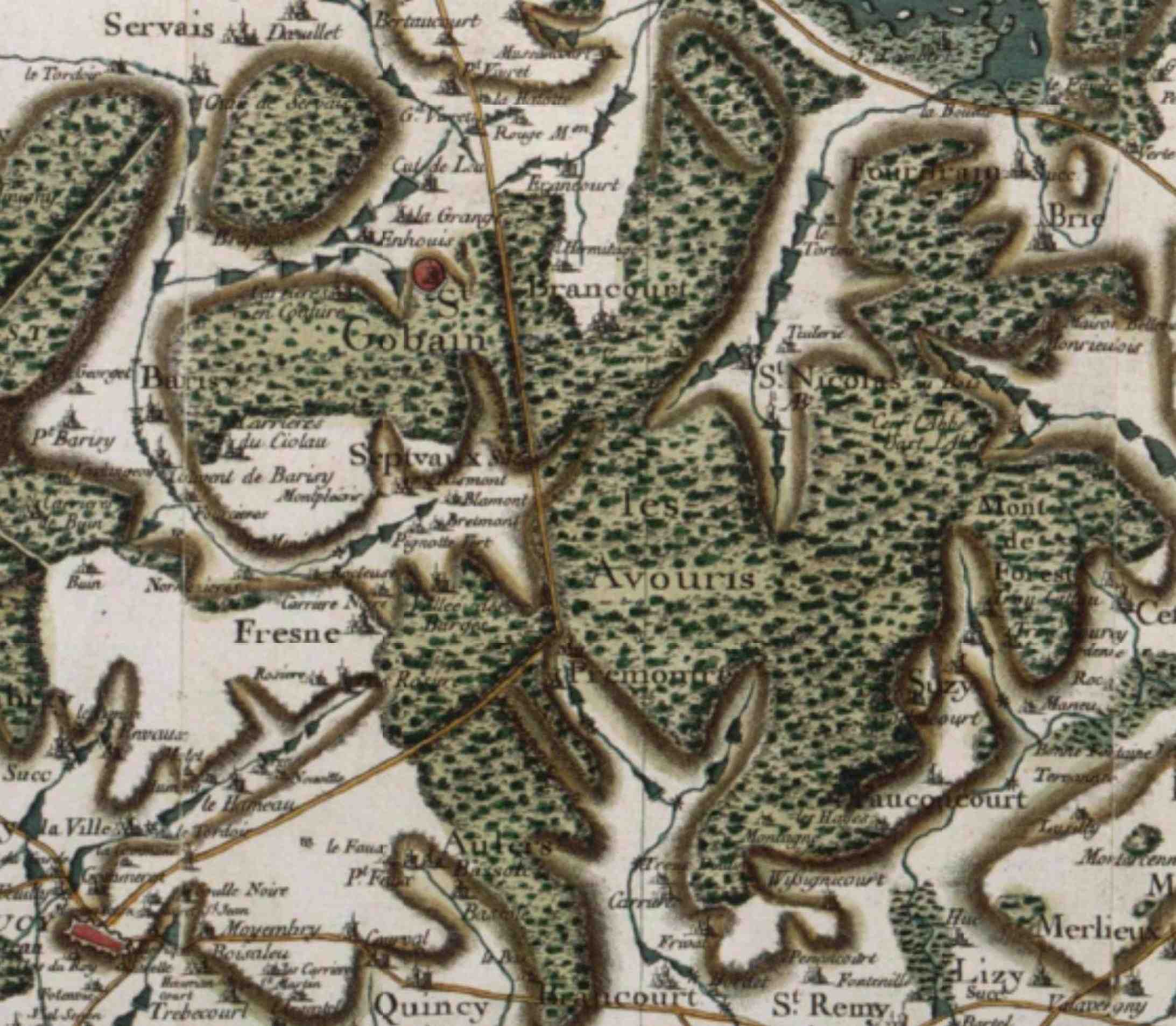}};
    \node (em) [label=above:\nth{19} century, right=0.1 of cass] {\includegraphics[width=2.2cm]{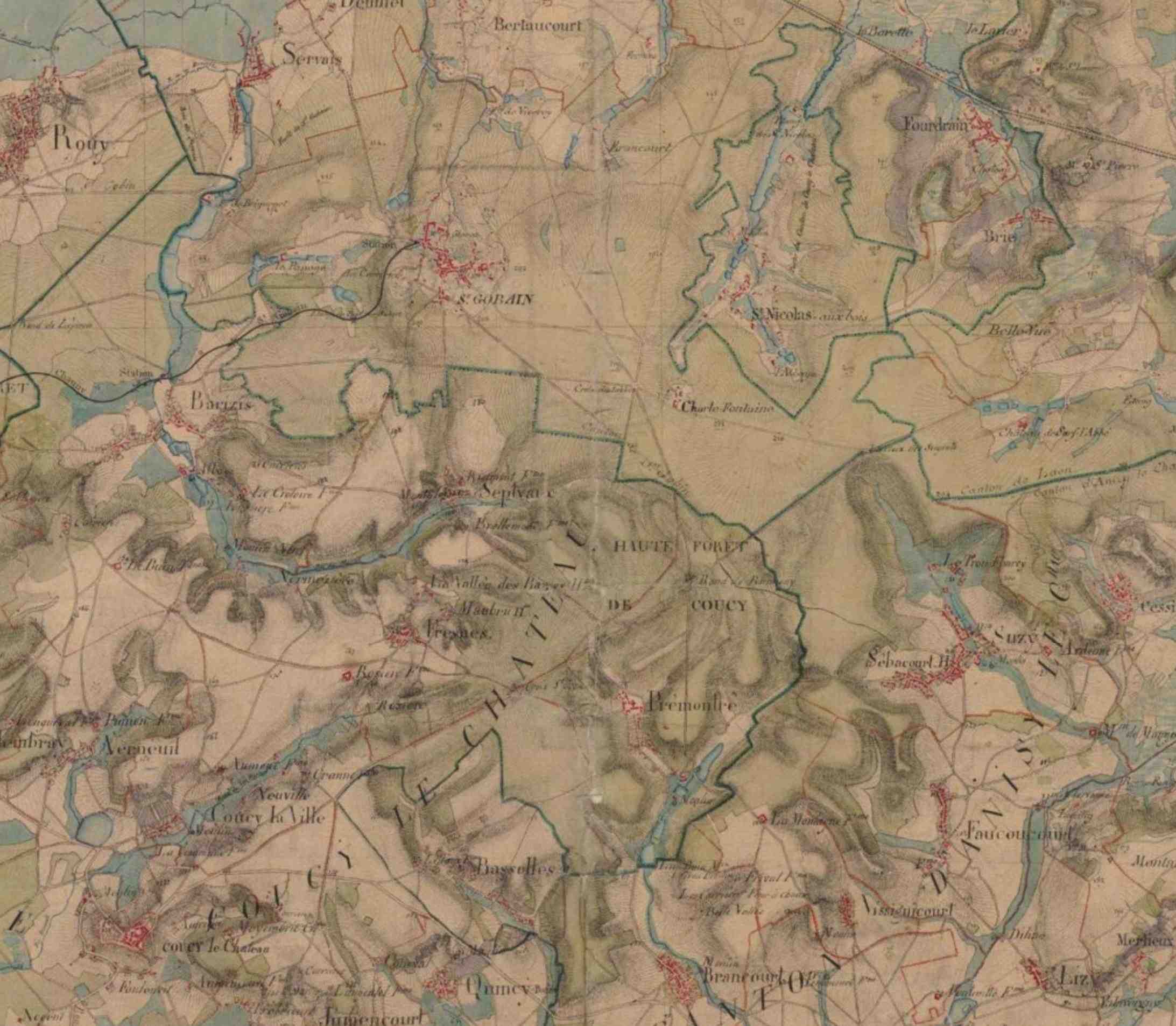}};
    \node (fift) [label=above:\nth{20} century, right=0.1 of em] {\includegraphics[width=2.2cm]{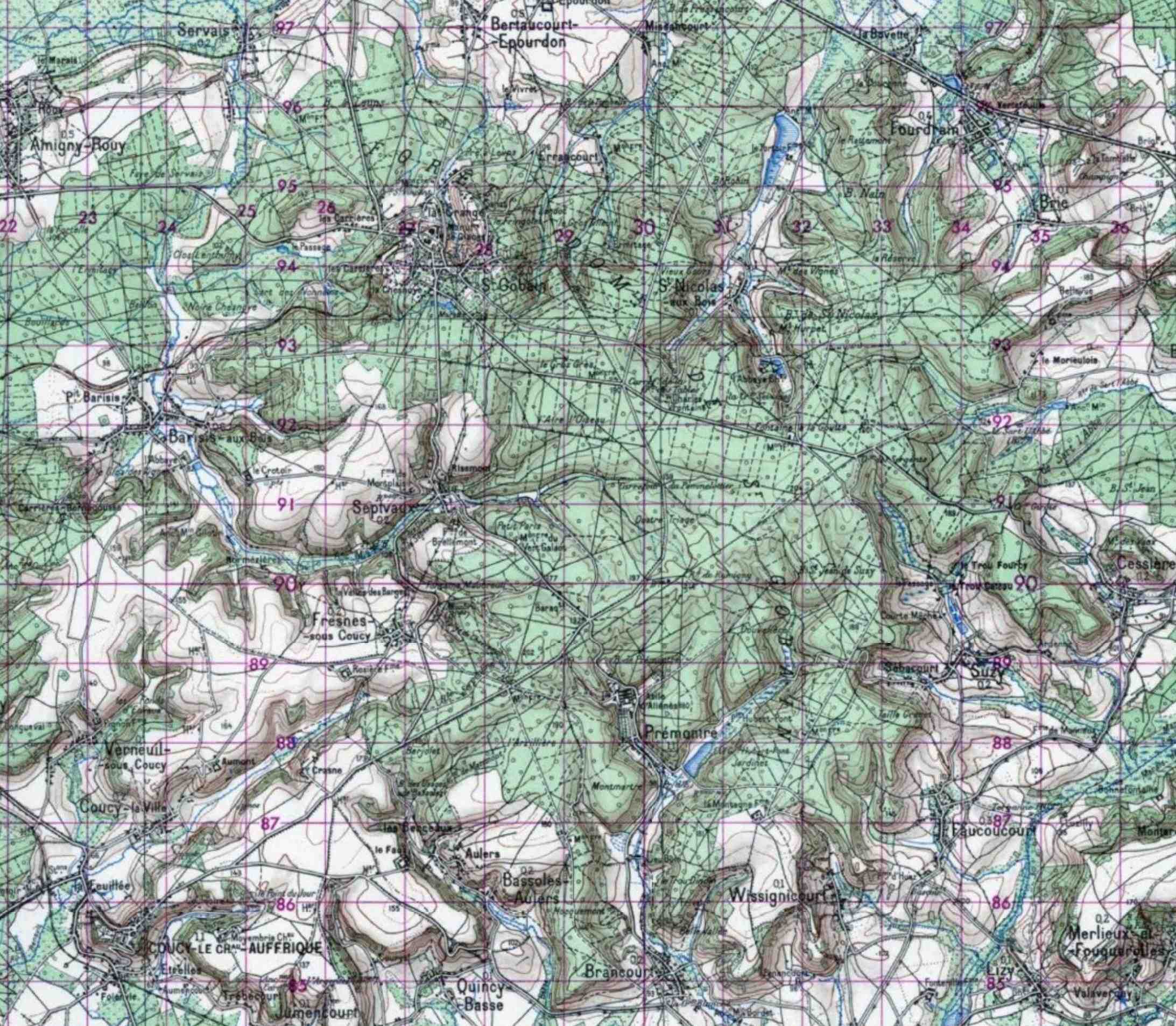}};
    \node (mod) [label=above:\nth{21} century, right=0.1 of fift] {\includegraphics[width=2.2cm]{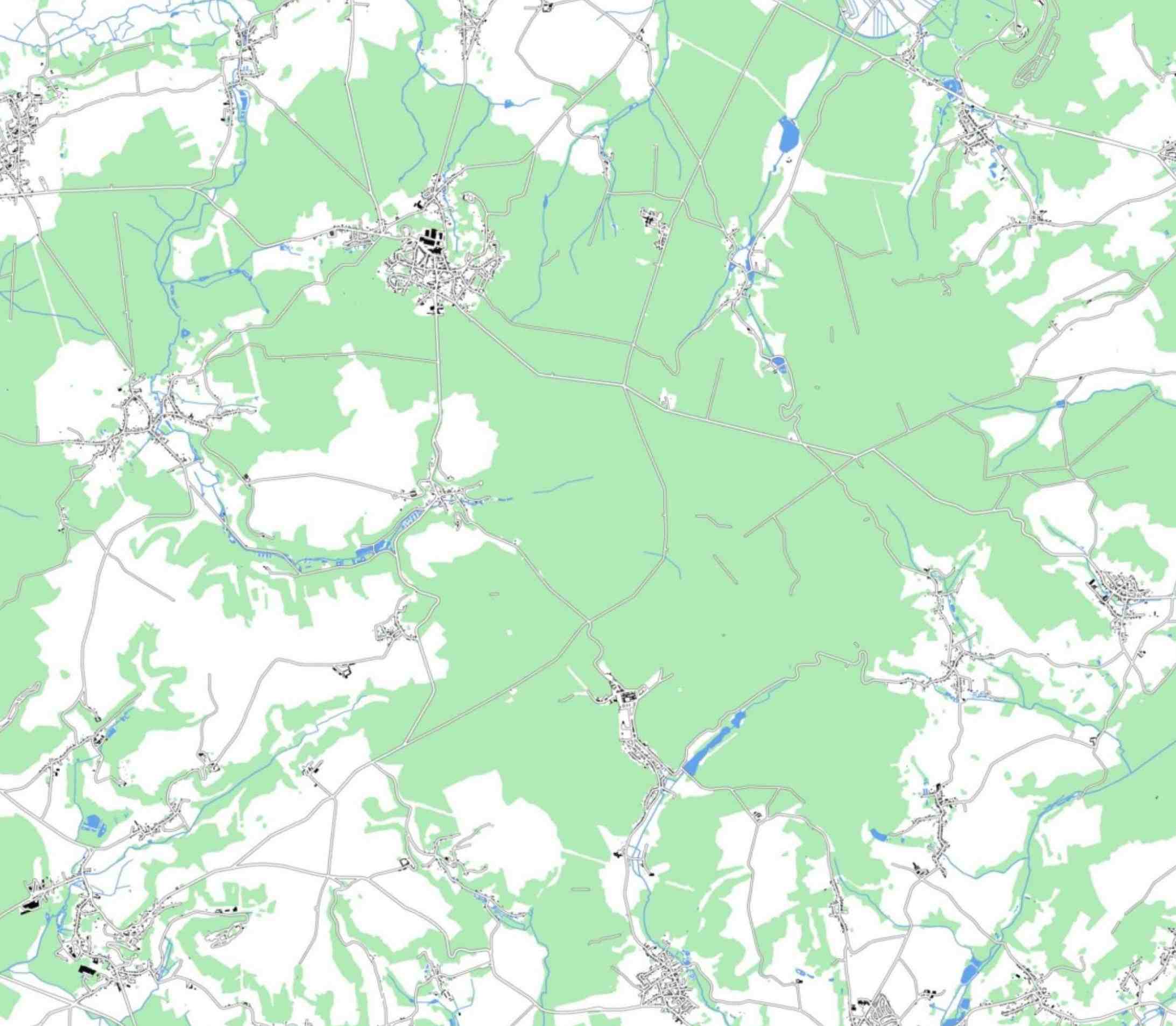}};

    % Labels
    \node (cass_lab) [below=0.4 of cass] {\fbox{\includegraphics[width=2.17cm]{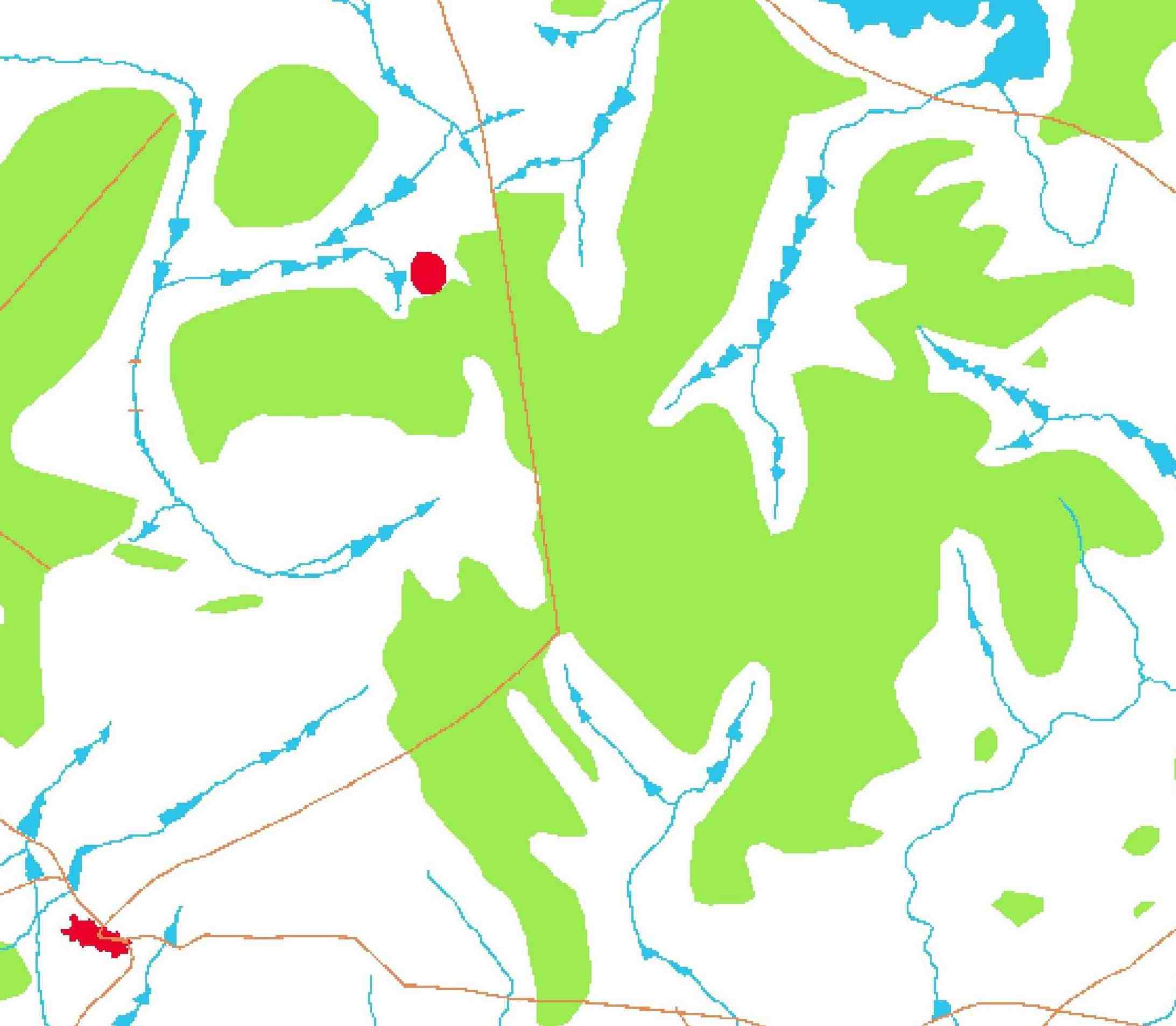}}};
    \node (em_lab) [right=0.1 of cass_lab] {\fbox{\includegraphics[width=2.17cm]{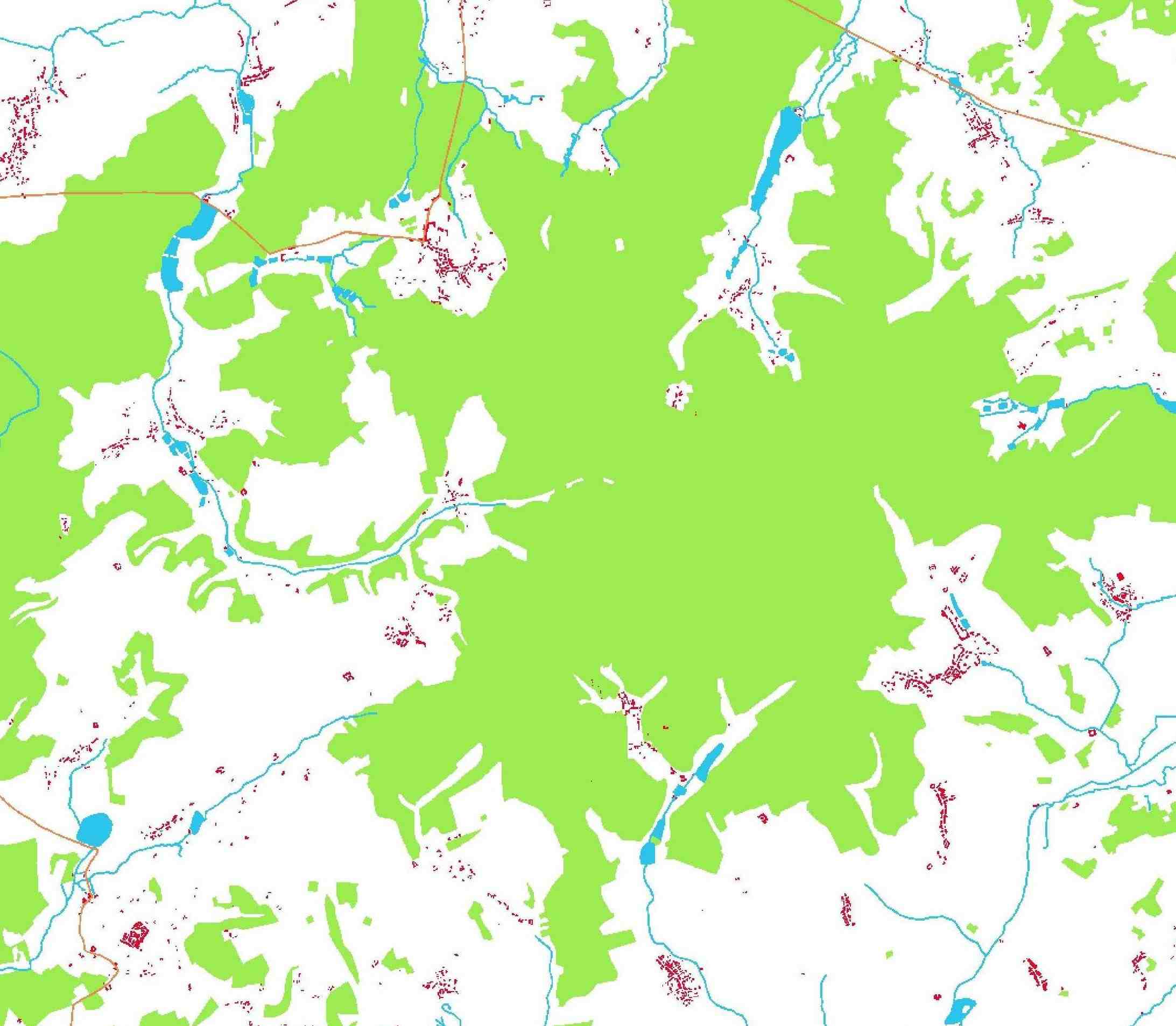}}};
    \node (fift_lab) [right=0.1 of em_lab] {\fbox{\includegraphics[width=2.17cm]{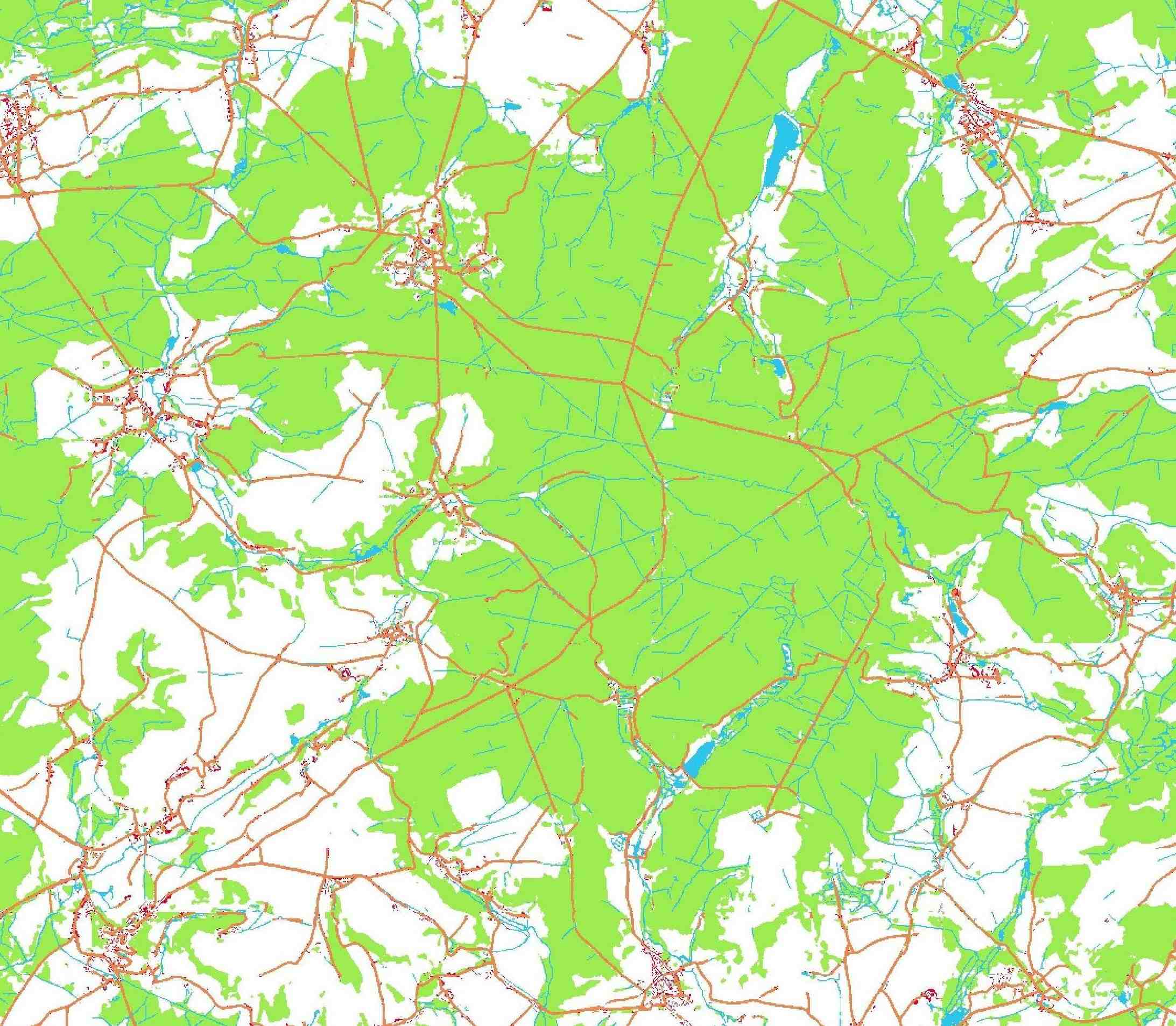}}};
    \node (mod_lab) [right=0.1 of fift_lab] {\fbox{\includegraphics[width=2.17cm]{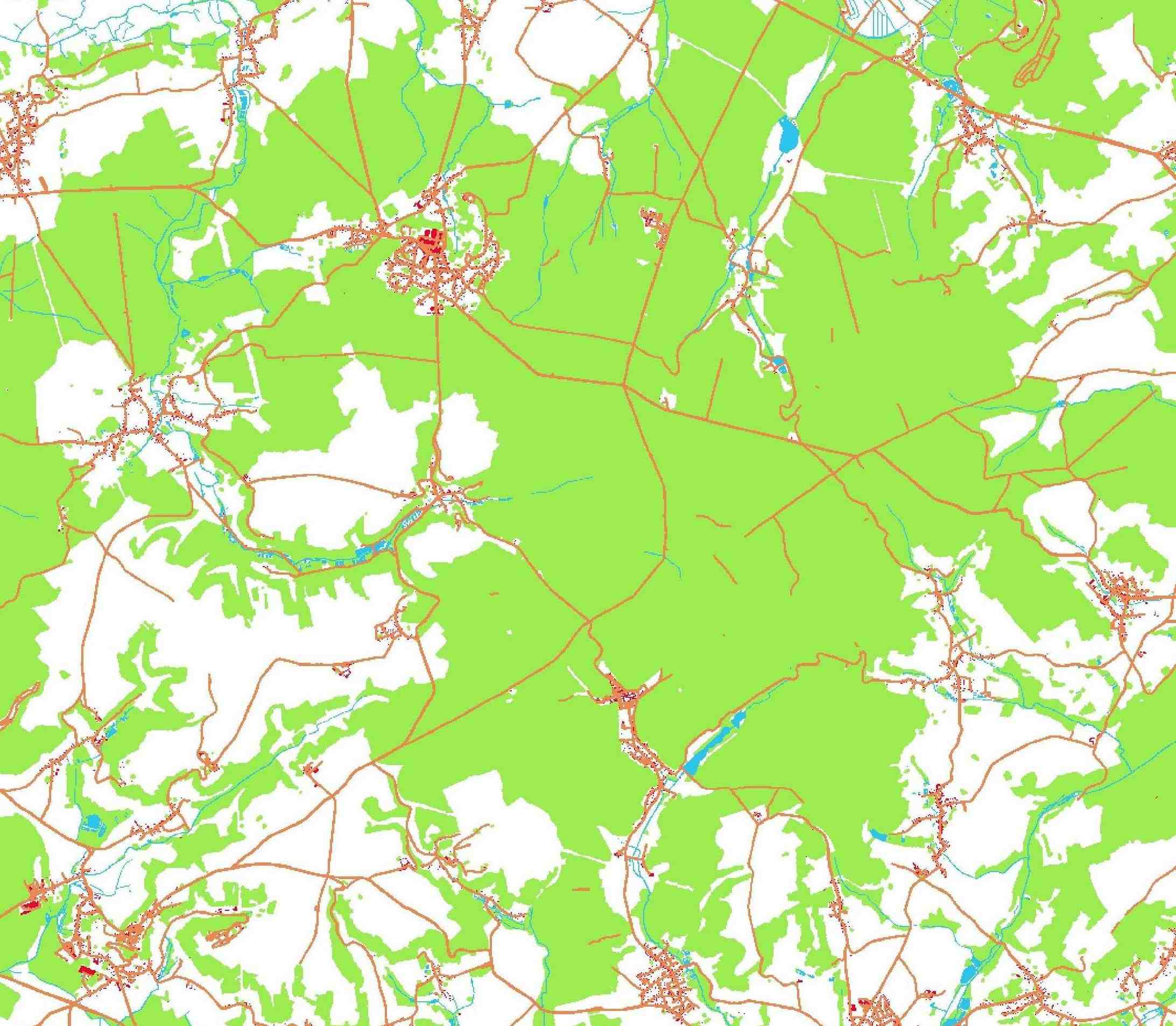}}};

    \draw[->, very thick,color=gray] ($(cass.south) - (0, 0.05)$) -- (cass_lab.north);
    \draw[->, very thick,color=gray] ($(em.south) - (0, 0.05)$) -- (em_lab.north);
    \draw[->, very thick,color=gray] ($(fift.south) - (0, 0.05)$) -- (fift_lab.north);
    \draw[->, very thick,color=gray] ($(mod.south) - (0, 0.05)$) -- (mod_lab.north);

    % Map of France
    \node (france) at ($(cass)!0.5!(cass_lab) - (2.8,0)$) {\includegraphics[width=2.5cm]{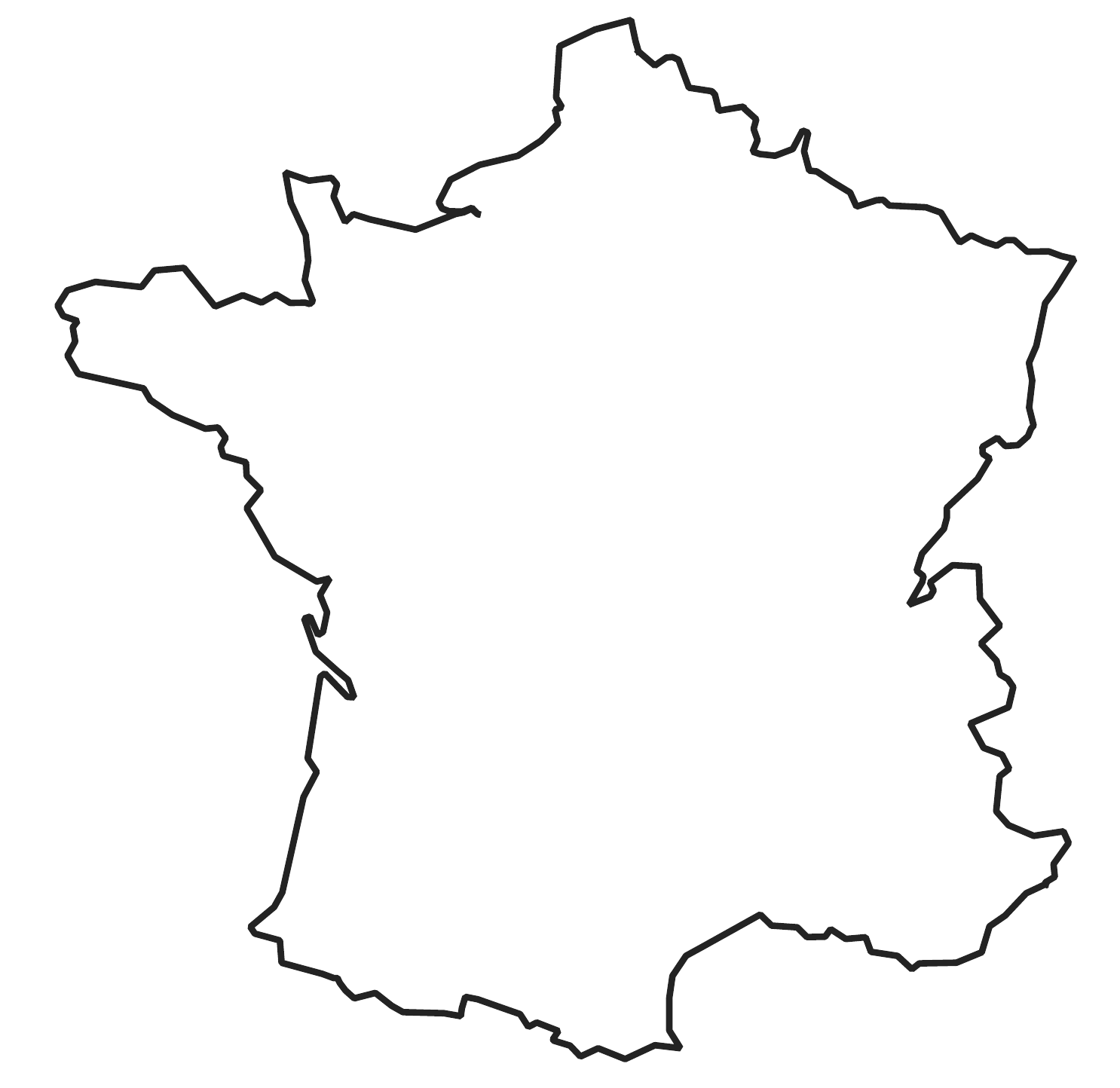}};

    \begin{scope}[x={(france.south east)},y={(france.north west)}]
        \coordinate (n1) at (0.22,0.565);
        \coordinate (n2) at (0.18,0.555);
        \coordinate (n3) at (0.22,0.54);
        \draw[draw=black] (n1) rectangle (n2);
    \end{scope}

    % Dashed outline
    \draw [draw=black,dashed] (n3) -- ($(cass.south west) + (-0.05, -0.05)$);
    \draw [draw=black,dashed] (n2) -- ($(cass.north west) + (-0.05, 0.05)$);
    \begin{scope}[on background layer, inner sep=1pt]
        \node (bg) [fit=(cass) (em) (fift) (mod)] {};
        \draw[dashed] ($(cass.north west) + (-0.05, 0.05)$) -- ($(mod.north east) + (0.05, 0.05)$);
        \draw[dashed] ($(cass.south west) + (-0.05, -0.05)$) -- ($(mod.south east) + (0.05, -0.05)$);
        \draw[dashed] ($(mod.north east) + (0.05, 0.05)$) -- ($(mod.south east) + (0.05, -0.05)$);
    \end{scope}

    % Label color legend
    \node (legend) at ($(em_lab)!0.5!(fift_lab) + (0.15, -1.3)$) {
        \begin{tabular}{C{0.3cm}@{\;}l@{\quad} C{0.4cm}@{\;}l@{\quad} C{0.4cm}@{\;}l@{\quad} C{0.3cm}@{\;}l@{\quad}}
            \tree & forest & \water & hydrography & \road & roads & \building & buildings 
        \end{tabular}
    };
    
\end{tikzpicture}

%% file: 2_related_work.tex
\section{Related work}

In this section, we review methods that have been applied to historical maps for semantic segmentation and image-to-image translation. In addition, we discuss how historical maps have been used to track and analyze long-term changes.

\subsection{Semantic Segmentation of Historical Maps}

While numerous studies address the segmentation of historical maps, most focus on isolating a single class or geographic feature: forests~\cite{herrault2013automatic}, railroads~\cite{chiang2020training}, bridges~\cite{wong2022semi}, roads~\cite{can2021automatic,ekim2021automatic,avci2022deep}, archaeological features~\cite{garcia2021potential}, mines~\cite{maxwell2020semantic}, cadastral parcels~\cite{oliveira2017machine,oliveira2019deep}, wetlands~\cite{jiao2020extracting,staahl2022identifying,o2024unleashing}, building footprints~\cite{hecht2018mapping,heitzler2020cartographic,laycock2011aligning} or human settlements~\cite{uhl2017extracting,uhl2018spatialising,uhl2019automated}. Here we focus on multi-class segmentation approaches and categorize them into two groups: traditional and deep-learning-based approaches.

\subsubsection{Traditional Approaches.}

Early attempts at segmenting historical maps primarily relied on color-based segmentation~\cite{leyk2010segmentation,chiang2013efficient}, line extraction techniques~\cite{khotanzad2003contour}, morphological operations~\cite{hecht2018mapping}, and watershed-based superpixels~\cite{miao2016guided}. These methods do not require annotated training data but involve complex, multi-step pipelines that must be customized for each specific historical map. Additionally, they depend heavily on clear and consistent color-coding and well-defined boundaries between shapes.

\subsubsection{Deep Learning Approaches.}

In recent years, deep learning has become the dominant approach for analyzing historical maps. While these methods reduce the need for manual engineering, they often require extensive annotated datasets, which can be costly to produce.

The majority of deep learning segmentation techniques that have been applied to historical maps are based on U-Net~\cite{ronneberger2015u,maxwell2020semantic,garcia2021potential,can2021automatic} or U-Net++~\cite{zhou2018unet,ekim2021automatic,martinez2023deep} architectures, which can be combined with Atrous Spatial Pyramid Pooling~\cite{wu2022leveraging} or attention~\cite{avci2022deep}. Some rely on other architectures such as region-based CNNs~\cite{saeedimoghaddam2020automatic}, simple FCNs~\cite{staahl2022identifying} or Deeplabv3+ and Ma-Net~\cite{avci2022deep}. Hybrid approaches, which combine deep learning with traditional techniques such as superpixels~\cite{miao2016guided} or morphological operations~\cite{liu2020superpixel}, have also shown promise.

To address the scarcity of annotated data, several strategies have been proposed, including transfer learning~\cite{chiang2020training,zhao2022building}, unsupervised domain adaptation~\cite{wu2023domain}, and weak supervision using contemporary~\cite{uhl2019automated,duan2020automatic,uhl2022towards} or multi-date~\cite{briand2024lulc} labels. Contemporary data can be leveraged more effectively by first homogenizing the visual style of historical and contemporary maps. This can be achieved through manual adaptation of symbology~\cite{jiao2022fast}, overpainting historical maps with standard symbols~\cite{muhlematter2024probabilistic}, or using image-to-image translation methods, which we discuss in the next subsection.

Following these works, our baselines rely on a variation of the U-net architecture. We also include in our dataset modern labels adapted with a suitable level of detail and color scheme for each historical map.

\subsection{Image-to-Image Translation on Maps}

Image-to-image translation for geospatial data is often used to translate satellite images to maps rather than maps to maps. These approaches often rely on CycleGAN~\cite{zhu2017unpaired}, but diffusion models have been used as well~\cite{tian2024mapgen}. Recent approaches improve the translated maps by leveraging color rendering information~\cite{li2020mapgan}, external geographic data~\cite{zhang2020enhanced}, cycle- and geometrical-consistency constraints~\cite{song2021mapgen}, spatial attention~\cite{song2022rsmt}, or a subset of closely-matched data~\cite{song2023semi}.

Previous works have already experimented on the image-to-image translation of historical maps, mostly employing Pix2Pix~\cite{andrade2022synthesis}, CycleGAN~\cite{li2021synthetic}, or both~\cite{arzoumanidis2024deep,christophe2022neural}, also in addition to CUT~\cite{wong2022semi}. Some of them have focused on generating synthetic data to train text detection~\cite{li2021synthetic} or semantic segmentation~\cite{arzoumanidis2024deep,wong2022semi} models.

We design a baseline based on CycleGAN, incorporating an additional translation loss accounting for weak annotations.

\subsection{Tracking Centuries of Change}

Historical maps are a valuable resource for understanding long-term dynamics in the social and natural sciences, especially when combined with aerial photography or remote sensing imagery~\cite{fuchs2015potential}. Observing historical land use and land cover evolution trends can support the development of sustainable policies for landscape, ecosystem, and heritage conservation. Existing methods rely on segmentation through manual vectorization or labeling~\cite{san2014urban,pindozzi2016using,picuno2019investigating}, multi-step color-based pipelines~\cite{hecht2018mapping,uhl2021combining}, object-based algorithms~\cite{liu2018integration,ulloa2020over,ettehadi2022integrated}, or deep learning~\cite{martinez2023deep,mayra2023utilizing}.

Although historical maps present unique challenges compared to remote sensing data~\cite{chiang2014survey}, the extensive literature on change detection in geospatial data offers valuable insights. In particular, when detecting changes between datasets captured by heterogeneous sensors or formats, a significant challenge lies in distinguishing actual semantic changes from variations in appearance. This problem has been addressed by translating both images into a common representation using conditional adversarial networks~\cite{niu2018conditional} or autoencoders~\cite{gong2019coupling,luppino2021deep}. Our dataset and baselines emphasize how these challenges translate to historical maps.

%% file: 3_dataset.tex
\section{Dataset}

\begin{table}[t]
    \caption{\textbf{Multi-Class Historical Map Datasets.} We report the characteristics of similar map collections. The column {\bf Labels} indicates the percentage of the map surface that is densely annotated. {\bf Classes} provides the nature of the labels, with possibly several subclasses per category. {\bf Multi-date} refers to whether the dataset provides \emph{aligned} maps of the same area belonging to different periods.}
    \label{tab:other_datasets}
    \resizebox{\columnwidth}{!}{
        \input{dataset_tab}
    }
\end{table}

We propose a new multi-date dataset of historical maps of metropolitan France spanning four centuries. We provide full contemporary annotations, along with partial historical annotations, making it ideal for evaluating weakly- or semi-supervised learning approaches. As shown in \cref{tab:other_datasets}, our dataset exceeds comparable collections in both spatial and temporal coverage, while also offering more diverse annotations.

\subsubsection{Composition.} Our dataset covers 548,305 km\textsuperscript{2} and is comprised of four aligned tile sets of 10{,}952 tiles, each tile being at least $1000\times1000$ pixels in size. These tile sets were sourced from four map collections, examples of which can be seen in \cref{fig:teaser}:

\begin{itemize} 
    \item {\bf Cassini (\nth{18} c.).} Commissioned by Louis XIV, the Cassini maps charted the entire Kingdom of France from 1750 to 1815 using geodesic triangulation~\cite{perret2015roads}. Their remarkable accuracy allows near-perfect alignment with modern maps. 
    \item {\bf Carte d'État-Major (\nth{19} c.).} Produced between 1820 and 1866 for the French military, these updates to the Cassini maps include topographic relief and individual buildings~\cite{huguenin1948historique}. 
    \item {\bf SCAN50 (\nth{20} c.).} Compiled in 1950 from map sheets produced during the first half of the 1900s, forming a cohesive mid-century map of France~\cite{scan50}. 
    \item {\bf Modern maps (\nth{21} c.).} Contemporary multi-scale vector maps obtained from IGN, the French Mapping Agency.
\end{itemize}
    
\href{https://www.ign.fr/institut/identity-card}{IGN} distributes these maps freely and in a common georeferenced coordinate system. We downloaded the historical maps at a scale of 1:40{,}000 and 150 DPI, corresponding to approximately 6.77 meters per pixel. We split the maps into sheets according to the Cassini mosaic. We selected the areas  covered by all three historical maps, corresponding to 175 out of 181 sheets. In the rest of the paper, we refer to the four map collections as \emph{Cassini}, \emph{État-Major}, \emph{SCAN50}, and \emph{modern maps}.

\subsubsection{Historical Labels.} 
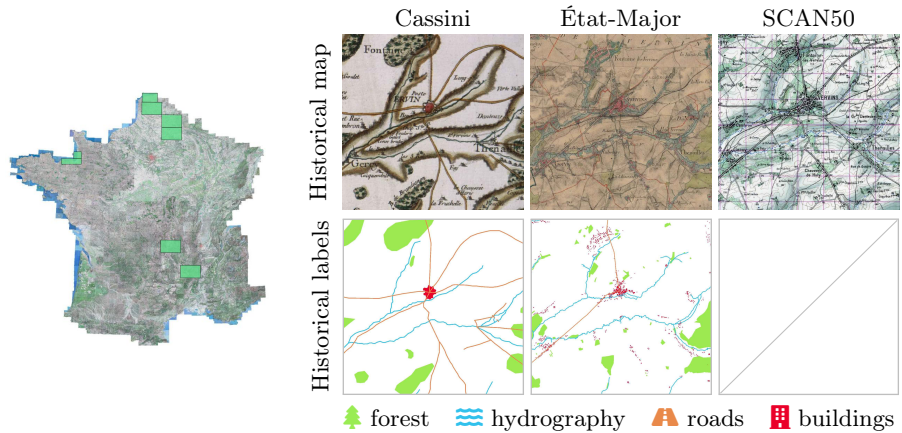
\begin{figure}[t]
    \begin{minipage}{0.29\textwidth}
        \centering
        \includegraphics[width=\linewidth]{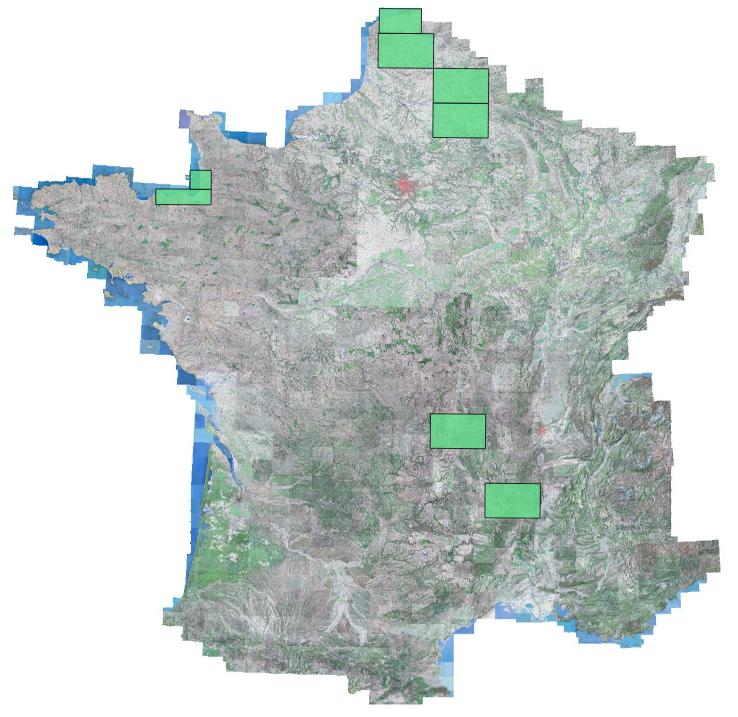}
    \end{minipage}
    \hfill
    \begin{minipage}{0.7\textwidth}
        \centering
        \begin{tabular}{C{0.4cm}C{2.4cm}C{2.4cm}C{2.4cm}}
            & Cassini & État-Major & SCAN50 \\
            \rotatebox{90}{Historical map} &
            \includegraphics[width=\linewidth]{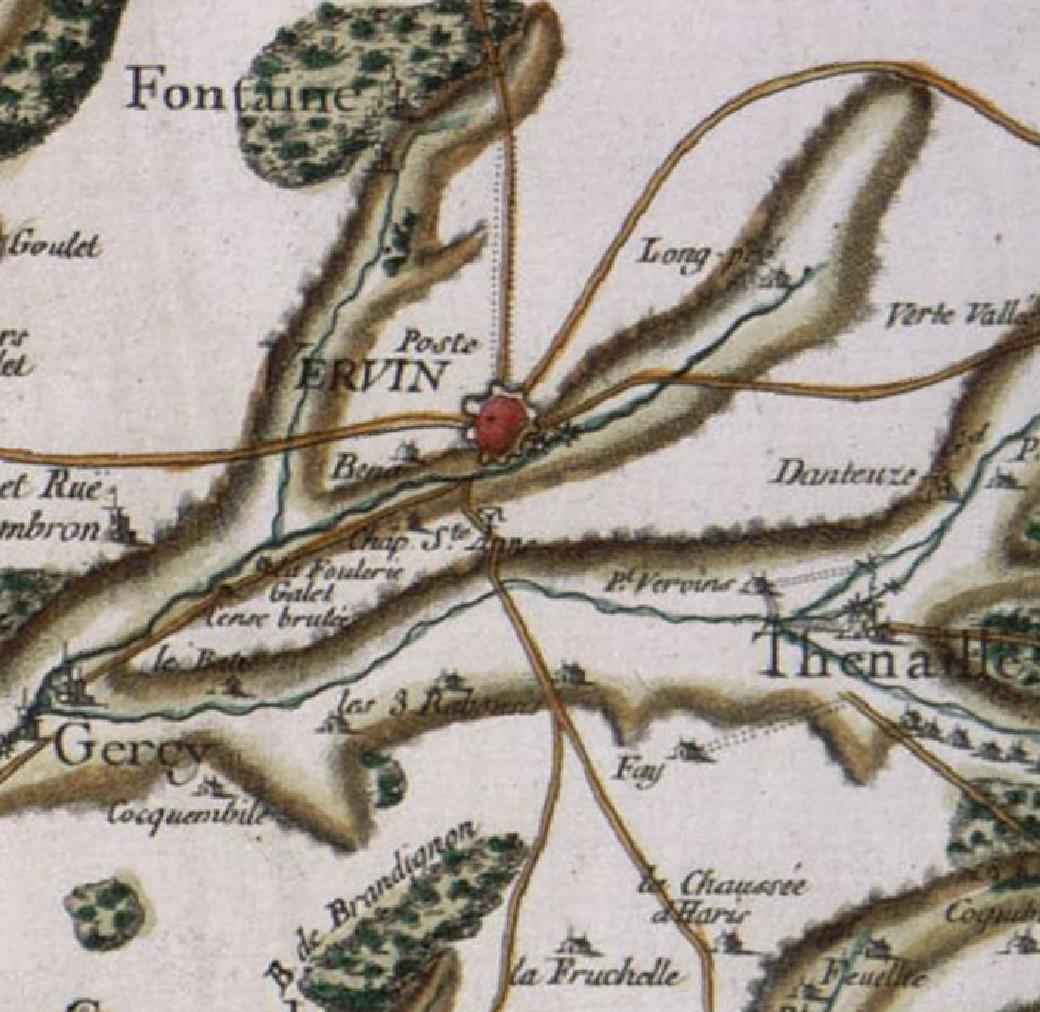} &
            \includegraphics[width=\linewidth]{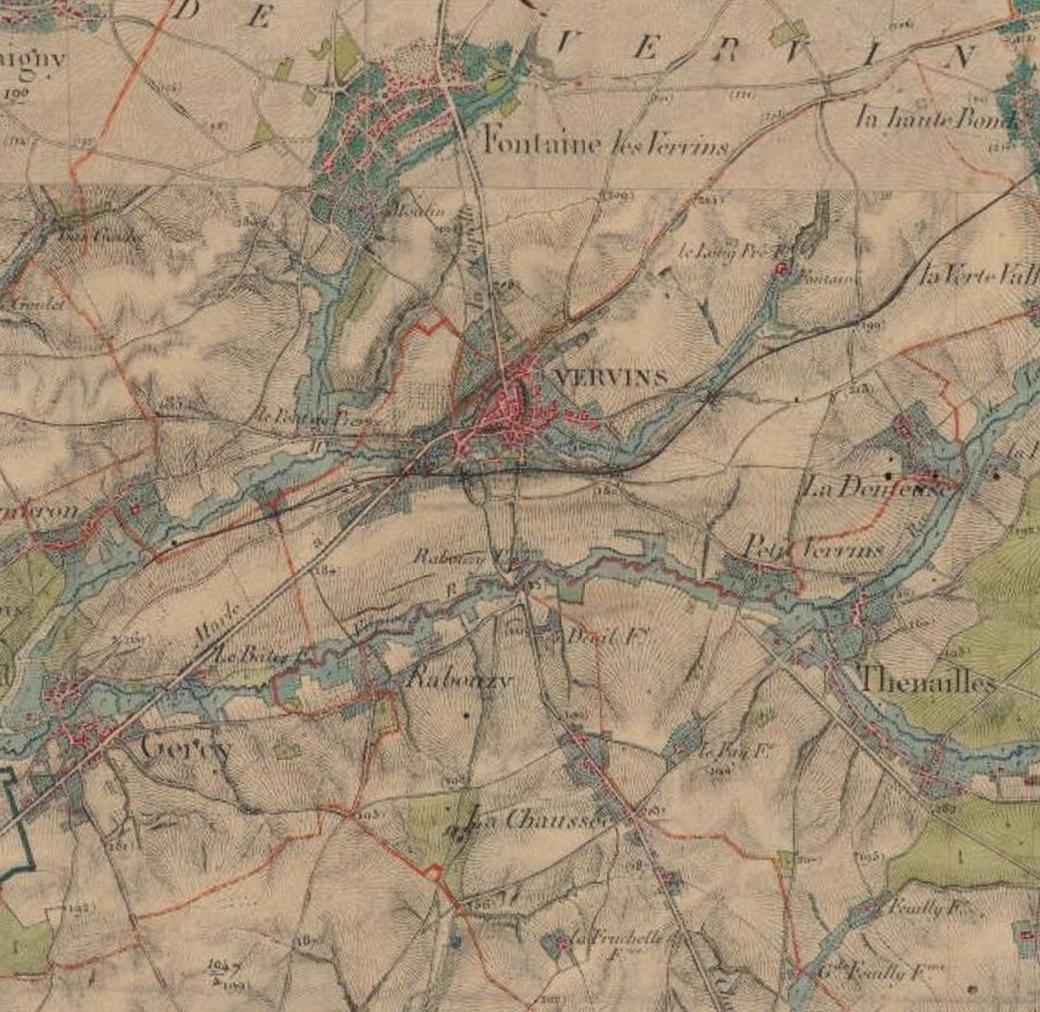} &
            \includegraphics[width=\linewidth]{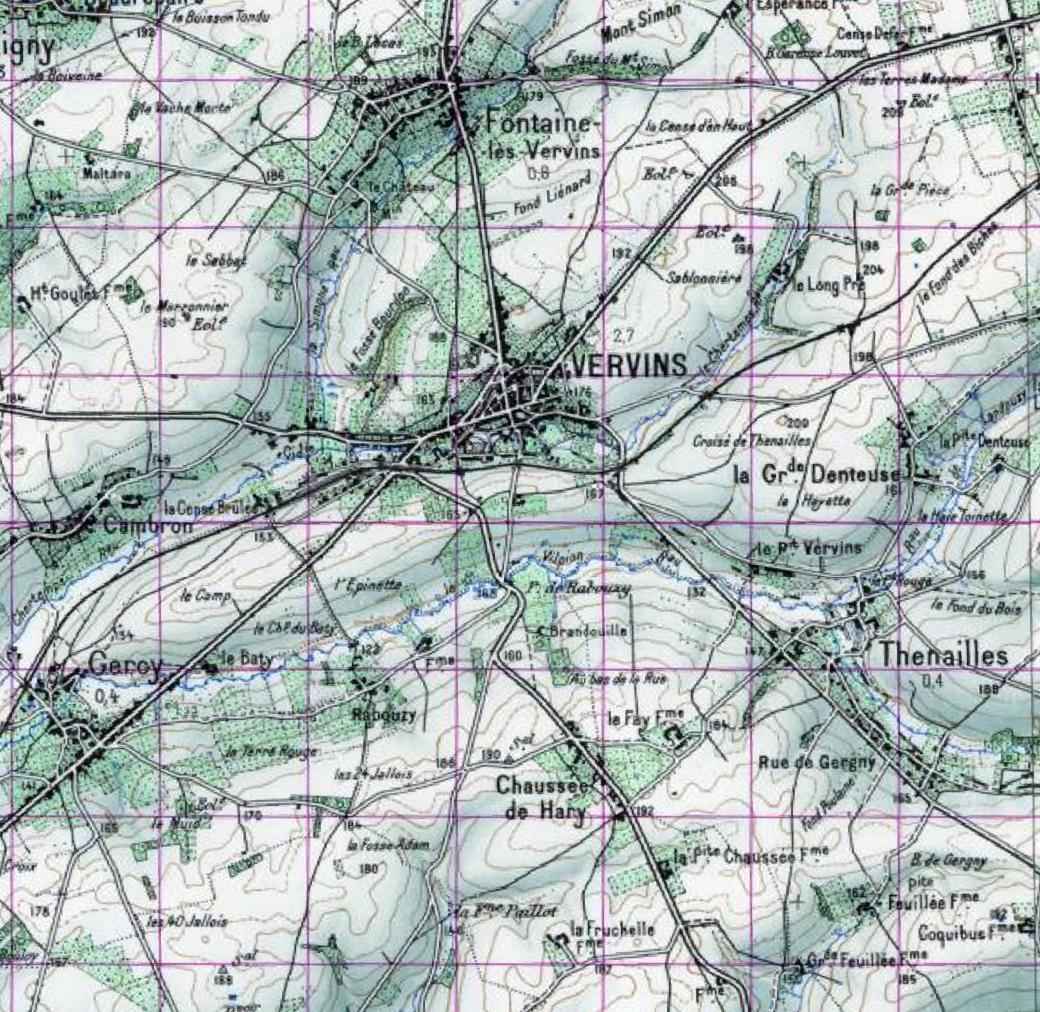} \\
            \rotatebox{90}{Historical labels} &
            \fbox{\includegraphics[width=0.99\linewidth]{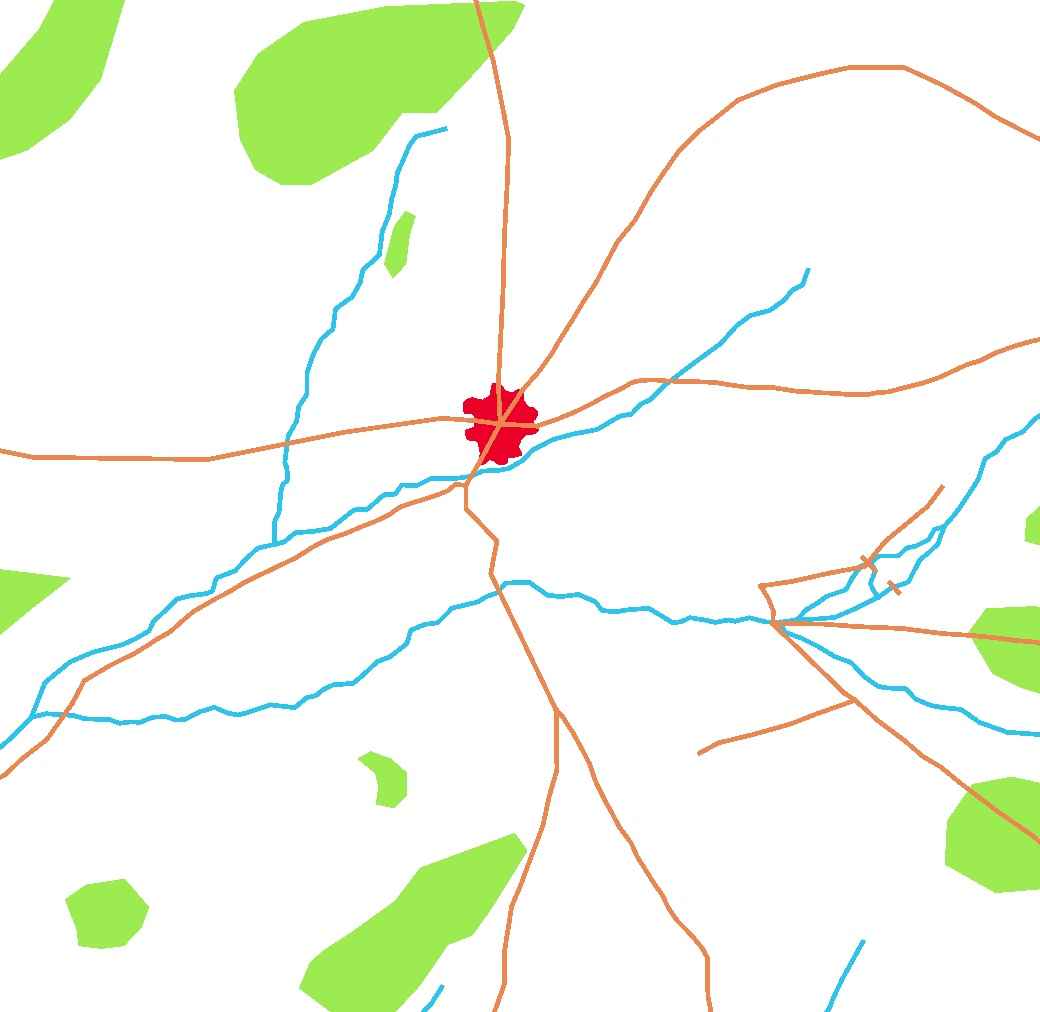}} & 
            \fbox{\includegraphics[width=0.99\linewidth]{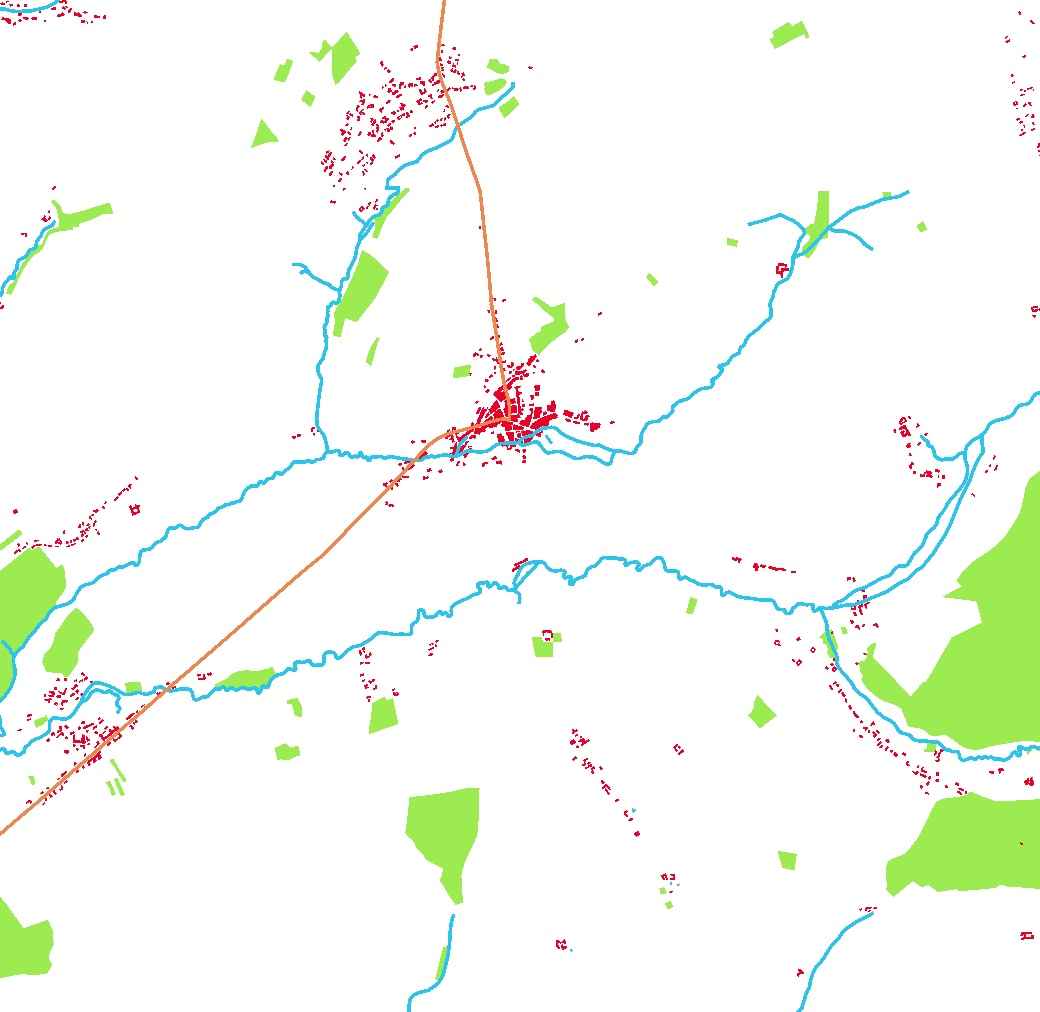}} & 
            \fbox{\begin{tikzpicture}[inner sep=0pt]
                \draw [draw=gray!50] (0,0) -- (2.35,2.3);
            \end{tikzpicture}} \\
        \end{tabular}
        \begin{tabular}{C{0.6cm}C{0.3cm}@{\;}l@{\quad}C{0.4cm}@{\;}l@{\quad}C{0.4cm}@{\;}l@{\quad}C{0.3cm}@{\;}l@{\quad}}
            & \tree & forest & \water & hydrography & \road & roads & \building & buildings 
        \end{tabular}
    \end{minipage}
    \caption{\textbf{Historical Maps and Labels.} Our dataset covers three collections of historical maps. We provide manual annotations of the highlighted regions of France for the Cassini and État-Major collections.}
    \label{fig:dataset_sample}
\end{figure}

To evaluate our baselines, we selected 470 tiles (covering 22{,}878 km\textsuperscript{2}) based on the extent of pre-existing annotations and terrain diversity. We obtained initial labels for Cassini from the GeoHistoricalData Project~\cite{geohistoricaldata} and manually corrected them for registration issues, missing annotations, and minor errors using five labels: \emph{forest}, \emph{hydrography}, \emph{roads}, \emph{buildings}, and \emph{background} (\cref{fig:dataset_sample}). For État-Major, we completed partial labels provided by IGN. As no initial labels were available for SCAN50, we do not provide them. In total, the additional manual labeling process required approximately 160 hours.

%% file: dataset_tab.tex
\begin{tabular}{l@{\;\;}c@{\;\;}l@{\;\;}c@{\;\;}l@{\;\;}r@{\;\;}r@{\;\;}r@{\;\;}r@{\;\;}l}
    \toprule
    \multirow{2}{*}{\bf{Dataset}} & 
    \multirow{1}{*}{\bf{Multi-}} &
    \multirow{2}{*}{\textbf{Collection}} &
    \multirow{2}{*}{\textbf{Access}} & 
    \multirow{2}{*}{\textbf{Location}} &
    \multirow{2}{*}{\textbf{Period}} &
    \multirow{2}{*}{\textbf{Scale}} &
    \multirow{1}{*}{\textbf{Extent}} &
    \multirow{1}{*}{\textbf{Labels}} &
    \multirow{2}{*}{\textbf{Classes}} \\
    & \multirow{1}{*}{\bf{date}} & & & & & & \;in km\textsuperscript{2} & \;\;in \% & \\ \midrule
    Wu et al. \cite{wu2022leveraging} & \faTimes & Siegfried & \req \auth & CH & 1880 &  1:25,000 & 23,048 & 100 & \water \\
    Martínez et al. \cite{martinez2023deep} & \faTimes & Cassini & \req & FR & 1750-1815 & 1:86,400 & 10,000  & 100 & \tree\;\water \\
    Can et al. \cite{can2021automatic} & \faTimes & Generalkarte & \href{https://doi.org/10.5281/zenodo.7073746}{\faDownload} & AT-HU & 1884-1918 & 1:200,000 & 321,146 & 100 & \road \\
    Ekim et al. \cite{ekim2021automatic} & \faTimes & DHK Turkey & \href{https://doi.org/10.5281/zenodo.7072623}{\faDownload} & TR & 1941-1943 & 1:200,000 & 132,970 & 100 & \road \\
    Hosseini et al. \cite{hosseini2022mapreader} & \faTimes & OS & \href{https://doi.org/10.5281/zenodo.7147906}{\faDownload} & GB & 1888-1913 & 1:10,560 & 620 & 100 & \building  \!\raisebox{-0.5mm}{\tracks{4mm}} \\
    Paris \cite{petitpierre2021generic} & \bof& Maps of Paris & \href{https://doi.org/10.5281/zenodo.5513639}{\faDownload}  & FR & 1765-1994 & $\geq$1:2,000 & $\leq$105 & 100 & \building\;\water\;\road \\
    World \cite{petitpierre2021generic} &\bof & 32 collections & \href{https://doi.org/10.5281/zenodo.5513639}{\faDownload} & 90 countries & 1720-1950 & - & - & 100 & \building\;\water\;\road \\ \midrule
    
    \multirow{4}{*}{\textbf{Ours}} & \multirow{4}{*}{\faCheck} & Cassini & \faDownload & FR & 1750-1815 & 1:86,400 & 548,305 & 4.2 & \tree\;\water\;\road\;\building \\
    && État-Major & \faDownload & FR & 1820-1866 & 1:40,000 & 548,305 & 4.2 & \tree\;\water\;\road\;\building \\
    && SCAN50 & \faDownload & FR & 1889-1922 & 1:50,000 & 548,305 & 0.0 & \tree\;\water\;\road\;\building \\
    && Modern maps & \faDownload & FR & 2019 & Vectorized & 548,305 & 100 & \tree\;\water\;\road\;\building \\ \bottomrule
    \\ [-1mm]
    \multicolumn{10}{c}{
        \begin{tabular}{rl@{\quad}rl@{\quad}rl@{\quad}}
            \faCheck & \emph{aligned} maps from multiple periods &
            \bof & \emph{unaligned} maps from multiple periods &
            \faTimes & maps from a single period 
        \end{tabular}
    } \\[1mm]
    \multicolumn{10}{c}{
        \begin{tabular}{rl@{\quad}rl@{\quad}rl}
            \req & requires request & 
            \auth & requires providers' authorization &
            \faDownload & open-access + direct download
        \end{tabular}
    } \\[1mm]
    \multicolumn{10}{c}{
        \begin{tabular}{rl@{\quad}rl@{\quad}rl@{\quad}rl@{\quad}rl}
            \water & hydrography &
            \tree & vegetation &
            \raisebox{-0.5mm}{\tracks{4mm}}\!\! & rail &
            \road & roads &
            \building & buildings 
        \end{tabular}
    }
\end{tabular}

%% file: 4_method.tex
\section{Method}
\label{sec:methods}

\begin{figure}[tb]
    \centering
    \begin{subfigure}[t]{0.49\textwidth}
        \centering
        \resizebox{0.95\columnwidth}{!}{\input{baselineA}}
        \caption{\textbf{Supervised Segmentation:} Trains on rare historical labels.}
        \label{fig:baselineA}
    \end{subfigure}
    \hfill
    \begin{subfigure}[t]{0.49\textwidth}
        \centering
        \resizebox{0.95\columnwidth}{!}{\input{baselineB}}
        \caption{\textbf{Direct Weakly-Supervised Segmentation:} Trains on abundant modern labels.}
        \label{fig:baselineB}
    \end{subfigure}%
    \vspace{0.1cm}
    \begin{subfigure}[t]{\textwidth}
        \centering
        \resizebox{0.95\columnwidth}{!}{\input{baselineC}}
        \caption{\textbf{Translation + Segmentation.} Learns to convert historical maps to a modern style, and to segment modern maps into label maps. For inference, performs these actions sequentially.}
        \label{fig:baselineC}
    \end{subfigure}
    
    \caption{\textbf{Historical Map Segmentation Baselines.} We train three semantic segmentation baselines to predict historical map labels, one supervised (\cref{fig:baselineA}) and two weakly-supervised (\cref{fig:baselineB} and \cref{fig:baselineC}).}
    \label{fig:baseline_training}
\end{figure}
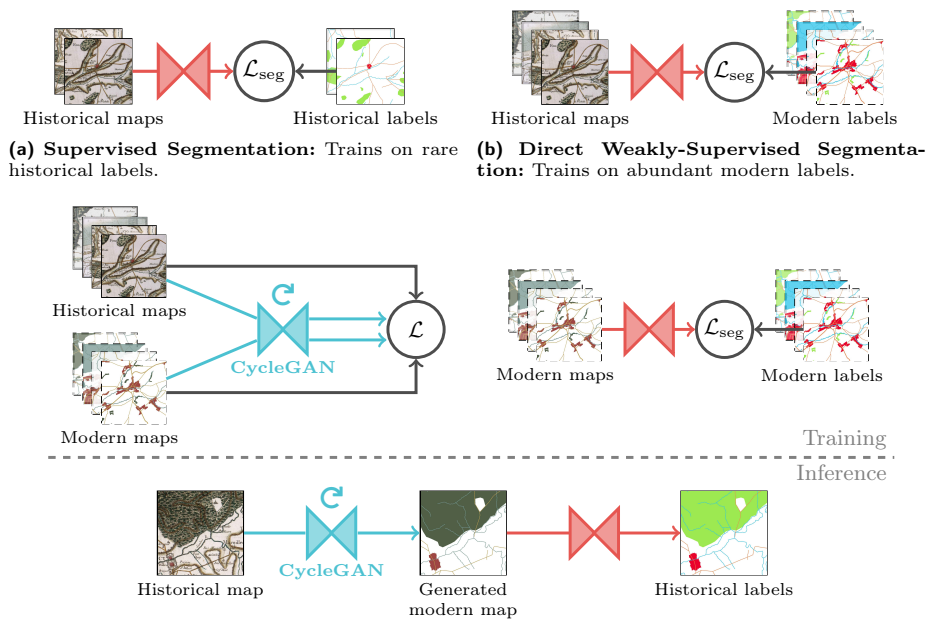

We introduce three baseline approaches for segmenting historical maps, illustrated in \cref{fig:baseline_training}. Our first approach is a fully-supervised model trained using scarce historical annotations (\cref{baselineA}). We also propose two weakly-supervised baselines that leverage abundant modern land cover labels (\cref{baselineB}). The first approach trains a segmentation model to directly map historical maps to modern land cover labels. The second approach first learns an image-to-image translation model to translate historical maps into a modern style, and then applies a segmentation model from modern maps to modern labels. All methods are evaluated on the same set of 470 tiles annotated for Cassini and État-Major. 

\subsection{Supervised Segmentation}\label{baselineA}

This method leverages our manually annotated historical labels to train a model using a standard supervised approach. As it relies on historical annotations, this baseline can only be applied when significant annotation efforts can be made, but it can be expected to provide the best results.

To evaluate this approach, we split the 470 annotated tiles into seven spatially distinct folds and perform 7-fold cross-validation. For each fold, we train a U-Net~\cite{ronneberger2015u} using six folds and evaluate it on the held-out fold. The performance we report corresponds to the aggregation of the results on left-out folds. During training, we further divide the six folds into a training and validation set (80/20 split), using the validation set to select the best-performing epoch.

\begin{figure}
    \centering
    \resizebox{0.95\columnwidth}{!}{\input{dataset_fig}}
    \caption{\textbf{Modern Maps and Labels Workflow.} We tailor the level of detail and appearance of the modern vector map labels to each historical map collection, and then derive a synthetic modern map by colorizing the labels.}
    \label{fig:modern_maps_workflow}
\end{figure}
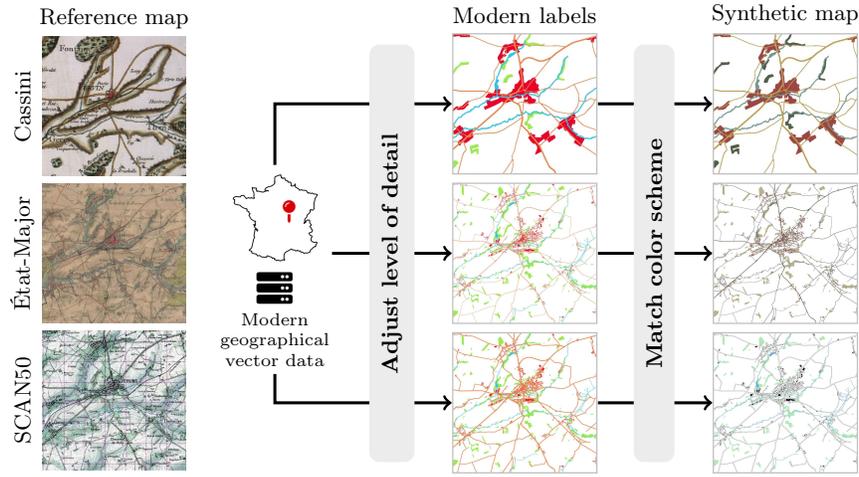

\subsection{Weakly-Supervised Segmentation}\label{baselineB}

Manually annotating historical maps is labor-intensive, making it impractical for large-scale studies. In contrast, modern land cover labels can be freely obtained from open access vector-based GIS databases.

\subsubsection{Direct Weakly-Supervised Segmentation.}

To leverage this, our first baseline is to train a U-Net to directly map historical maps to modern land cover labels. We exclude the 470 manually annotated tiles used for evaluation and split the remaining 10{,}482 historical tiles into 9{,}096 training tiles and 1{,}386 validation tiles. To obtain modern labels for each historical map, we select a level of detail consistent with its content, as visualized in the middle part of \cref{fig:modern_maps_workflow}.

Naturally, this task is highly challenging---asking the model, in essence, to  ``see'' centuries into the future. However, we hypothesize that certain permanent landscape features, such as old-growth forests, major transportation corridors, and waterways, provide a supervisory signal, allowing the model to learn meaningful correspondences despite temporal discrepancies. However, training a segmentation model directly on weakly-aligned data becomes particularly challenging when the temporal gap between historical maps and modern labels is large---for instance, in the case of Cassini maps, dating back to the \nth{18} century.

\subsubsection{Translation + Segmentation.}

To try and address this, our second weakly-supervised baseline follows a two-step pipeline:
\begin{itemize}
    \item {\bf Image-to-Image Translation.} We first train a model to translate historical maps into a modern cartographic style, reducing stylistic discrepancies while maintaining content consistency with the historical map, thanks to the cycle consistency loss.
    \item {\bf Segmentation.} We then train a segmentation model on these translated maps to predict modern land cover labels.
\end{itemize}
To facilitate image-to-image translation, we generate synthetic modern maps by coloring modern label maps with a palette that mimics the appearance of historical maps, as visualized in the right part of \cref{fig:modern_maps_workflow}. This preprocessing step helps stabilize the translation training process.

For the image translation step, we employ a modified CycleGAN model, which we detail in the next section. Once translated, the historical maps adopt a visual style closer to modern maps, making them more compatible with segmentation models. Finally, we train a U-Net to segment these transformed maps into land cover labels.

\subsection{Implementation Details}

\subsubsection{U-Net.}

We use an 8-stage nnU-Net~\cite{isensee2021nnu} architecture modified to enable mirror padding. Each stage applies a 3$\times$3 convolution with stride 2, except for the first stage, which uses stride 1. Our training loop follows nnU-Net’s default pipeline except we only use random resized crops as data augmentation. We train for 200 epochs with a batch size of 32. The cropping size is adjusted to match the scale of each historical map collection: 1000$\times$1000 pixels for Cassini, and 500$\times$500 pixels for État-Major and SCAN50.

\subsubsection{CycleGAN.} \label{cyclegan_details}

Given a source domain $\mathcal{X}$ and a target domain $\mathcal{Y}$, CycleGAN~\cite{zhu2017unpaired} learns an image translation function (or \emph{generator}) $G_{\mathcal{X}\mapsto \mathcal{Y}} : \mathcal{X} \mapsto \mathcal{Y}$ that transforms images $x \in \mathcal{X}$ to the style of domain $\mathcal{Y}$. The introduction of a reverse mapping $G_{\mathcal{Y}\mapsto \mathcal{X}} : \mathcal{Y} \mapsto \mathcal{X}$ enables training on unaligned image pairs. Two domain \textit{discriminators} $D_{\mathcal{X}}$ and $D_{\mathcal{Y}}$ distinguish between real and generated images. The total loss consists of three terms:
\begin{itemize}
    \item The \textit{adversarial loss}~\cite{goodfellow2020generative} $\mathcal{L}_{\text{GAN}}$ encourages the generators to produce realistic outputs, and the discriminators to discriminate fake images from real ones.
    \item The \textit{cycle consistency} loss $\mathcal{L}_{\text{cyc}}$  ensures that translating an image to the opposite domain and back preserves its content.
    \item The \textit{identity} loss $\mathcal{L}_{\text{id}}$ regularizes the generators to maintain color consistency.
\end{itemize}

Barring temporal discrepancies and alignment errors, the content of historical and modern maps has some consistency. To leverage this weak alignment, we introduce an additional translation loss $\mathcal{L}_{\text{tran}}$:
\begin{align}
    \mathcal{L}_{\text{tran}} = \mathbb{E}_{x \sim \mathcal{X}}\| G_{\mathcal{Y}\mapsto\mathcal{X}}(y) - x \|_1 + \mathbb{E}_{y \sim \mathcal{Y}}\| G_{\mathcal{X}\mapsto\mathcal{Y}}(x) - y \|_1~.
\end{align}

The final objectives are:
\begin{align}
    \mathcal{L}(G_{\mathcal{X} \mapsto \mathcal{Y}}, G_{\mathcal{Y} \mapsto \mathcal{X}}) &= \mathcal{L}_{\text{GAN}} + \lambda_{\text{cyc}} \mathcal{L}_{\text{cyc}} + \lambda_{\text{id}} \mathcal{L}_{\text{id}} + \lambda_{\text{tran}} \mathcal{L}_{\text{tran}}, \\
    \mathcal{L}(D_{\mathcal{X}}, D_{\mathcal{Y}}) &= -\mathcal{L}_{\text{GAN}},
\end{align}
with $\lambda_{\text{cyc}}$, $\lambda_{\text{id}}$ and $\lambda_{\text{tran}}$ non-negative hyperparameters. We set $\lambda_{\text{cyc}}=1.0$, $\lambda_{\text{id}}=0.5$ and $\lambda_{\text{tran}}=0.5$ throughout all experiments.

We use ResNet-based generators with 9 blocks, while the discriminators are PatchGANs~\cite{isola2017image}. We train the networks for 100 epochs with a batch size of 1, a constant learning rate of \num{2e-4} and apply the same random cropping and resizing strategy as for U-Net segmentation. Opposite to CycleGAN, which uses an image buffer to stabilize training by sampling fakes from previous iterations, we only consider fakes generated from the images in each batch, taking further advantage of the weak alignment.

%% file: baselineA.tex
\begin{tikzpicture}[inner sep=0pt,z={(3.85mm, -3.85mm)}]
    \path (0,0) -- (1,0.9); % Spacing

    \pic (histA) [local bounding box=histA] at (0,0) {MapBlock=hist/2};
    \node (histA_label) [below=0.1 of histA,map_labels] {Historical maps};
    \pic (unetA) [right=1.5*\interdist of histA_map_1,local bounding box=unetA] {unet=unet_color};
    \node (lcA) [loss,right=1.5*\interdist of unetA] {$\mathcal{L}_{\text{seg}}$};
    \pic (hist_annot) at (\secondcol-0.1,0) [local bounding box=hist_annot] {MapBlock=hist_annot/2};
    \node (hist_annot_label) [below=0.1 of hist_annot,map_labels] {Historical labels};

    \draw [very thick,color=unet_color](histA_map_1) to (unetA);
    \draw [->,very thick, color=unet_color] (unetA) to (lcA);
    \draw [<-,very thick,color=black!70] (lcA) to (hist_annot_map_1);
\end{tikzpicture}

%% file: baselineB.tex
\begin{tikzpicture}[inner sep=0pt,z={(3.85mm, -3.85mm)}]

    \pic (histB) [local bounding box=histB] at (0,0) {MapBlock=hist/\mapsperblock};
    \node (histB_label) [below=0.1 of histB,map_labels] {Historical maps};
    \pic (unetB) [right=1.4*\interdist of histB_map_1,local bounding box=unetB] {unet=unet_color};
    \node (lcB) [loss,right=1.4*\interdist of unetB] {$\mathcal{L}_{\text{seg}}$};
    \pic (mod_annot) at (\secondcol,0) [local bounding box=mod_annot,densely dashed] {MapBlock=mod_annot/\mapsperblock};
    \node (mod_annot_label) [below=0.1 of mod_annot,map_labels] {Modern labels};

    \draw [very thick,color=unet_color] (histB_map_1) to (unetB);
    \draw [->, very thick,color=unet_color] (unetB) to (lcB);
    \draw [<-, very thick,color=black!70] (lcB) to (mod_annot_map_1);

\end{tikzpicture}

%% file: baselineC.tex
\def\secondrow{-1.8}
\def\thirdrow{-3.8}
\begin{tikzpicture}[inner sep=0pt,z={(3.85mm, -3.85mm)}]
    % CycleGAN
    \pic (histC) [local bounding box=histC] at (0,0) {MapBlock=hist/\mapsperblock};
    \node (histC_label) [below=0.1 of histC,map_labels] {Historical maps};
    \pic (mod) at (0,\secondrow) [local bounding box=mod,densely dashed] {MapBlock=mod/\mapsperblock};
    \node [below=0.1 of mod, map_labels] {Modern maps};
    \pic (cyclegan) [right=8*\interdist of $(histC_map_1)!0.5!(mod_map_1)$, local bounding box=cyclegan] {cyclegan};
    \node (lgan) [loss,right=5*\interdist of cyclegan_net] {$\mathcal{L}$};
    
    \draw [very thick,color=cyclegan_color](histC_map_1) to (cyclegan_net);
    \draw [very thick,color=cyclegan_color](mod_map_1) to (cyclegan_net);
    \draw [->, very thick,color=cyclegan_color] ([yshift=1.5mm]cyclegan_net.east) to ([yshift=1.5mm]lgan.west);
    \draw [->, very thick,color=cyclegan_color] ([yshift=-1.5mm]cyclegan_net.east) to ([yshift=-1.5mm]lgan.west);
    \draw [<-, very thick,color=black!70] (lgan) to (lgan |- mod_map_1) to (mod_map_1);
    \draw [<-, very thick,color=black!70] (lgan) to (lgan |- histC_map_1) to (histC_map_1);

    % Segmentation
    \pic (modC) [local bounding box=modC,densely dashed] at (6.2, \secondrow / 2.0) {MapBlock=mod/\mapsperblock};
    \node (modC_label) [below=0.1 of modC,map_labels] {Modern maps};
    \pic (unetC) [right=1.4*\interdist of modC_map_1,local bounding box=unetC] {unet=unet_color};
    \node (lcC) [loss,right=1.4*\interdist of unetC] {$\mathcal{L}_{\text{seg}}$};
    \pic (modC_annot) at (\secondcol + 6.2,\secondrow / 2.0) [local bounding box=modC_annot,densely dashed] {MapBlock=mod_annot/\mapsperblock};
    \node [very thick] (modC_annot_label) [below=0.1 of modC_annot,map_labels] {Modern labels};
    
    \draw [very thick,color=unet_color] (modC_map_1) to (unetC);
    \draw [->, very thick,color=unet_color] (unetC) to (lcC);
    \draw [<-, very thick,color=black!70] (lcC) to (modC_annot_map_1);

    % INFERENCE
    \node[draw,very thick] (hist_inf) at (\secondcol/4,\thirdrow) {\includegraphics[width=1.2cm]{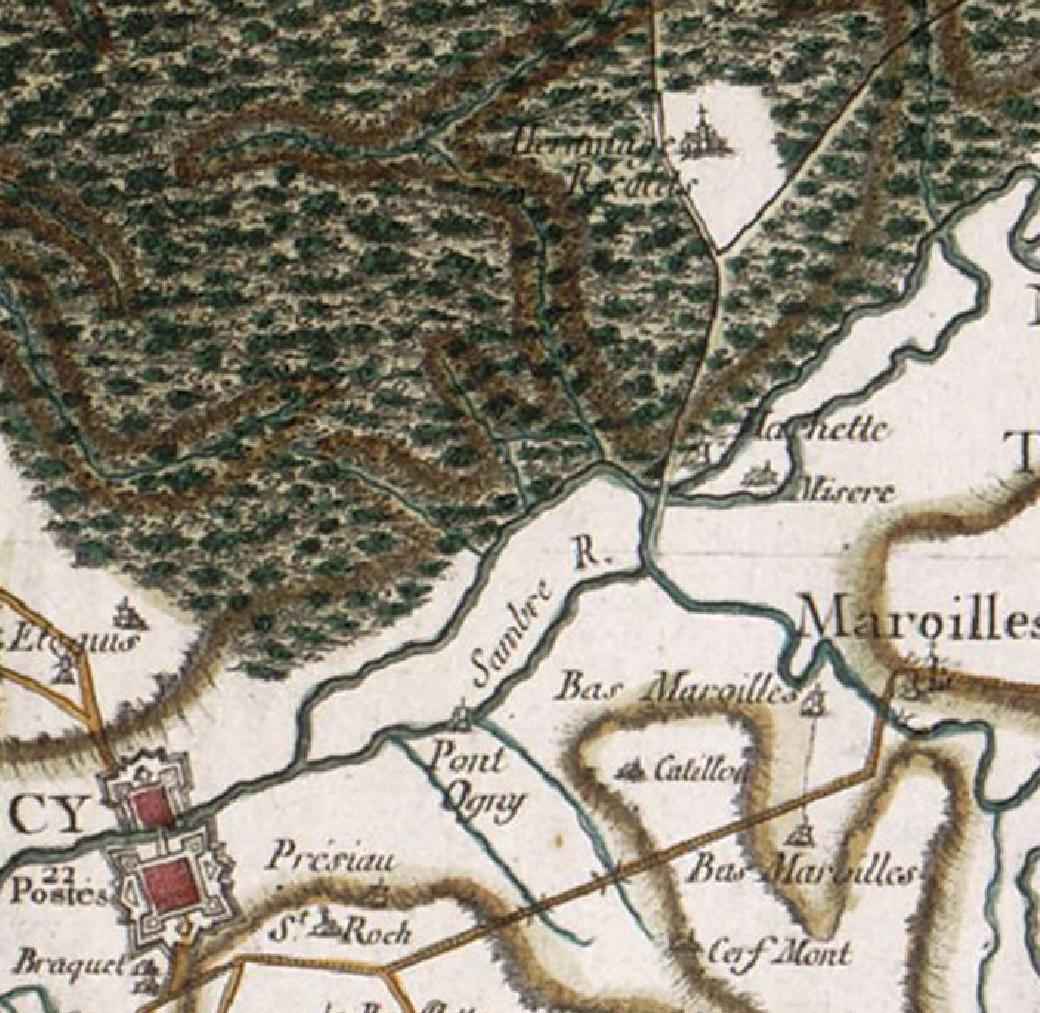}};
    \node (hist_inf_label) [below=0.1 of hist_inf,map_labels] {Historical map};

    % CycleGAN
    \pic (cyclegan_inf) [right=4*\interdist of hist_inf, local bounding box=cyclegan_inf] {cyclegan};
    \node[draw,very thick,color=black,right=4*\interdist of cyclegan_inf_net] (fake_mod) {\includegraphics[width=1.2cm]{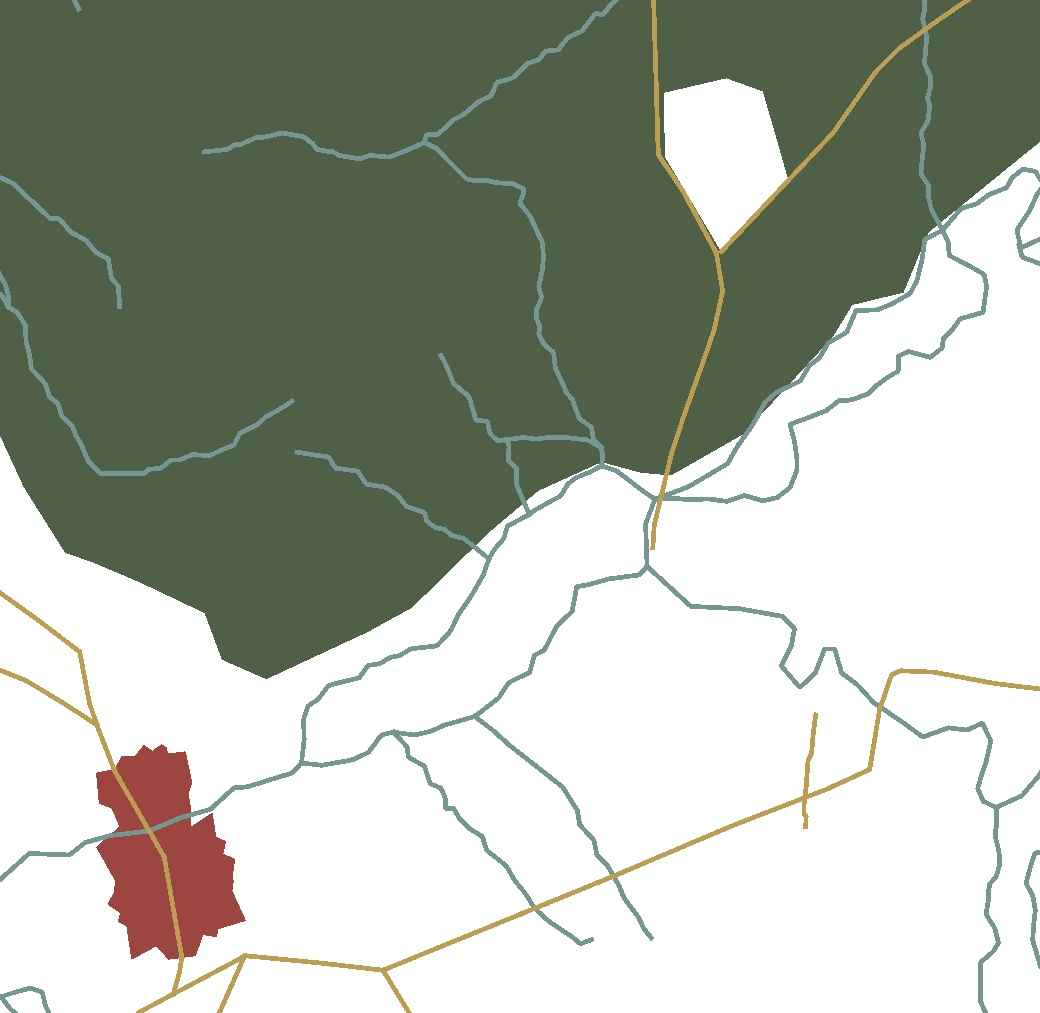}};
    \node (fake_mod_label) [below=0.1 of fake_mod, map_labels] {Generated modern map};

    % Segmentation
    \pic (unetC_inf) [right=4*\interdist of fake_mod,local bounding box=unetC_inf] {unet=unet_color};
    \node[draw,very thick,right=4*\interdist of unetC_inf] (hist_annot_inf) {\includegraphics[width=1.2cm]{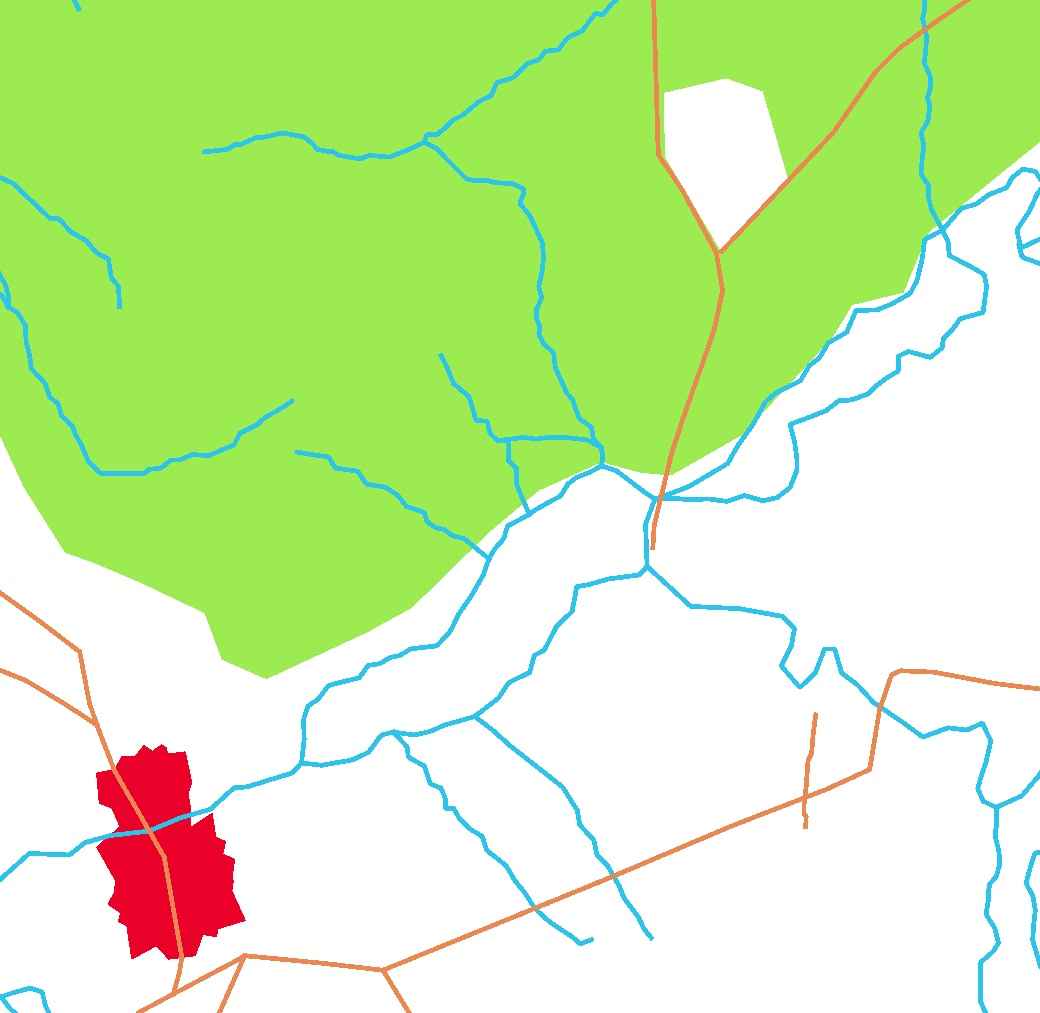}};
    \node (hist_annot_inf_label) [below=0.1 of hist_annot_inf,map_labels] {Historical labels};
    
    \draw [very thick,color=cyclegan_color] (hist_inf) to (cyclegan_inf_net);
    \draw [->, very thick,color=cyclegan_color] (cyclegan_inf_net) to (fake_mod);
    \draw [very thick,color=unet_color](fake_mod) to (unetC_inf);
    \draw [->, very thick,color=unet_color] (unetC_inf) to (hist_annot_inf);

    \begin{scope}[on background layer]
        \node (bg) [fit=(histC) (histC_label) (modC_annot_label)(hist_annot_inf_label)] {};
        \draw [dashed,draw=black!50,very thick] ([yshift=-0.8cm]bg.west) -- ([yshift=-0.8cm]bg.east);
        \node [below left=0.9cm and 0.1 of bg.east] {\textcolor{black!50}{Inference}};
        \node [below left=0.4cm and 0.1 of bg.east] {\textcolor{black!50}{Training}};
    \end{scope}
\end{tikzpicture}

%% file: dataset_fig.tex
\begin{tikzpicture}
    \node (server) at (0, 0) {\Large \faServer};
    \node (france) [above=-0.1 of server] {\includegraphics[width=1.3cm]{france_outline_svg-tex.pdf}};
    \node at ($(france) + (0.2, 0.1)$) {\color{red!85!black} \faMapPin};
    \node (server_label) [below=-0.1 of server,text width=2cm,align=center,font=\scriptsize] {Modern geographical vector data};

    % Reference maps
    \node (em_ref) at ($(france)!0.5!(server) - (2.25cm, 0)$) [inner sep=0] {\includegraphics[width=2cm]{em1_hist.jpeg}};
    \node (cass_ref) [above=0.1 of em_ref,inner sep=0] {\includegraphics[width=2cm]{cass1_hist.jpeg}};
    \node (fift_ref) [below=0.1 of em_ref,inner sep=0] {\includegraphics[width=2cm]{fift1_hist.jpeg}};
    \node (reference_name) at ($(cass_ref.north) + (0,0.25)$) {Reference map};

    \node (em_name) [anchor=east] at (em_ref.west) {\rotatebox{90}{État-Major}};
    \node (cass_name) [anchor=east] at ($(cass_ref.west) + (-0.05,0)$) {\rotatebox{90}{Cassini}};
    \node (fift_name) [anchor=east] at ($(fift_ref.west) + (-0.05,0)$) {\rotatebox{90}{SCAN50}};

    % Modern labels
    \node (em_labels) at ($(france)!0.5!(server) + (3.5cm, 0)$) [inner sep=0] {\fbox{\includegraphics[width=2cm]{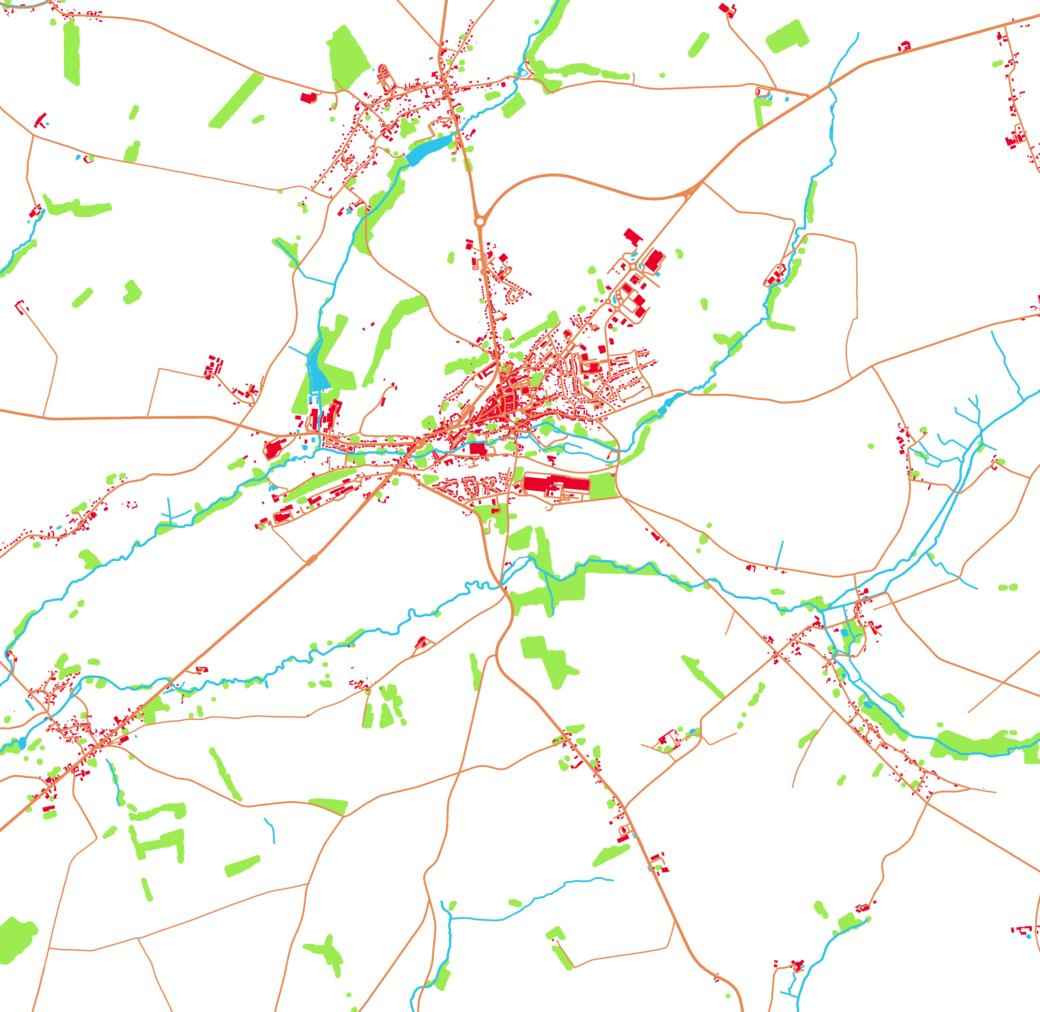}}};
    \node (cass_labels) [above=0.1 of em_labels,inner sep=0] {\fbox{\includegraphics[width=2cm]{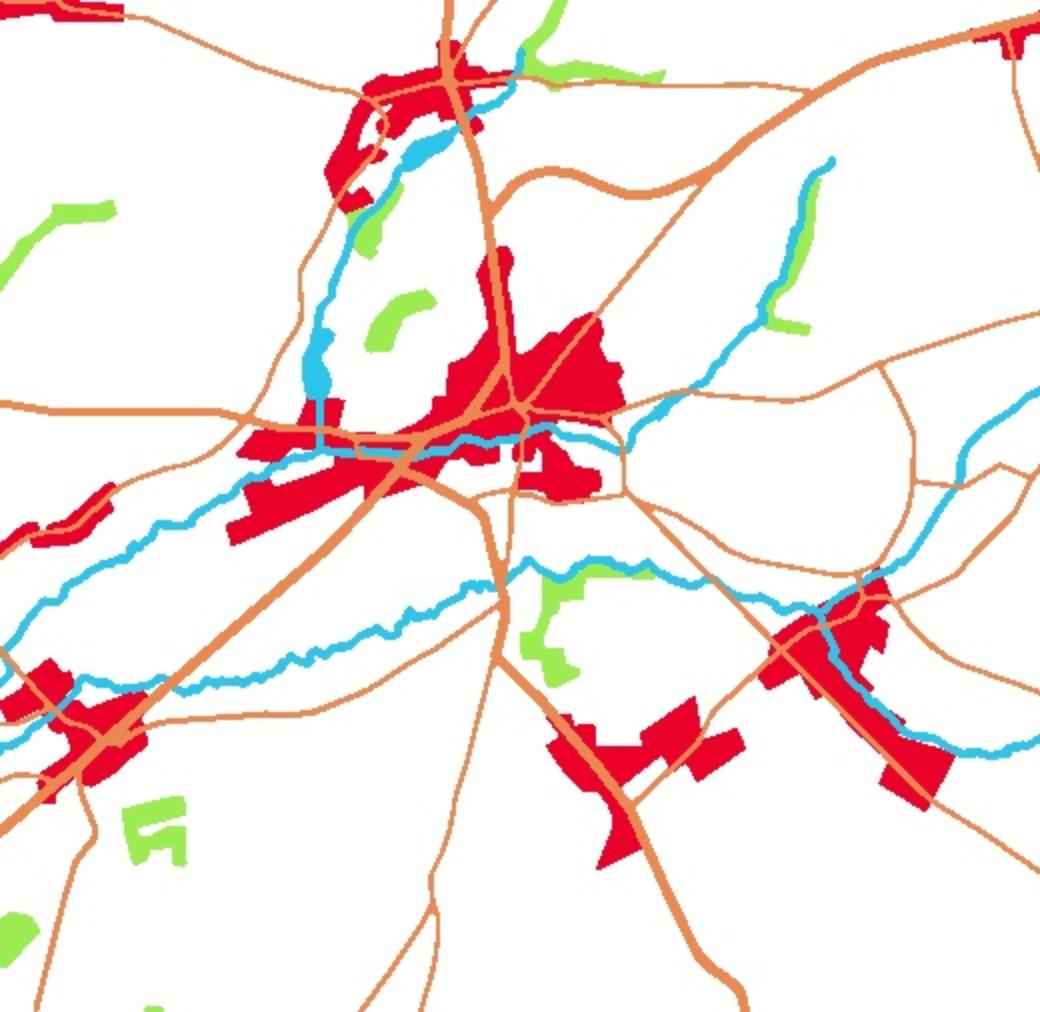}}};
    \node (fift_labels) [below=0.1 of em_labels,inner sep=0] {\fbox{\includegraphics[width=2cm]{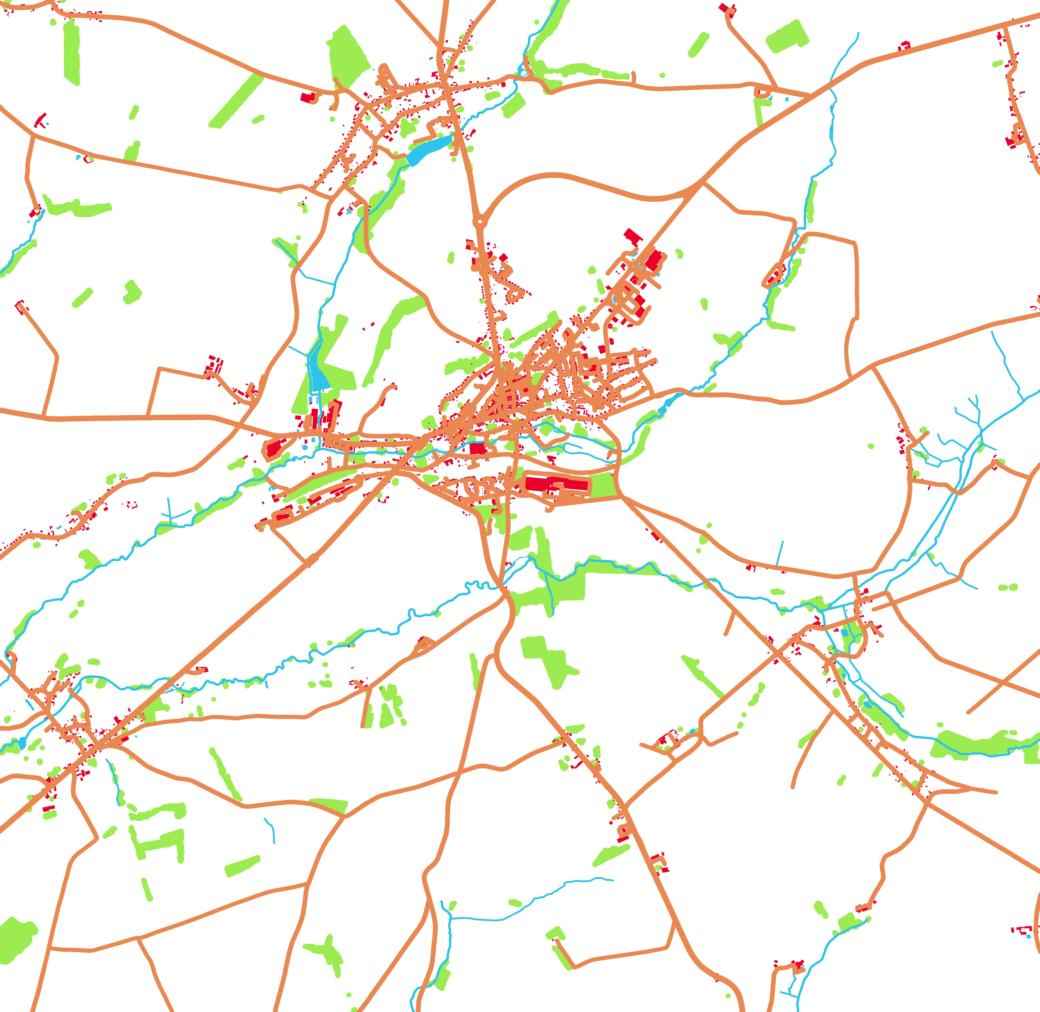}}};
    \node (mod_labels_name) at ($(cass_labels.north) + (0,0.25)$) {Modern labels};

    \draw[->, very thick] (france) |- (cass_labels);
    \draw[->, very thick] (server_label) |- (fift_labels);
    \draw[->, very thick] (em_ref -| france) ++(0.8, 0) -- (em_labels) node[align=center,rotate=90,anchor=north,pos=0.3,fill=gray!15, text width=5.5cm, rounded corners, inner sep=5] {\bf Adjust level of detail};

    % Modern maps
    \node (cass_mod) [right=1.6cm of cass_labels,inner sep=0] {\fbox{\includegraphics[width=2cm]{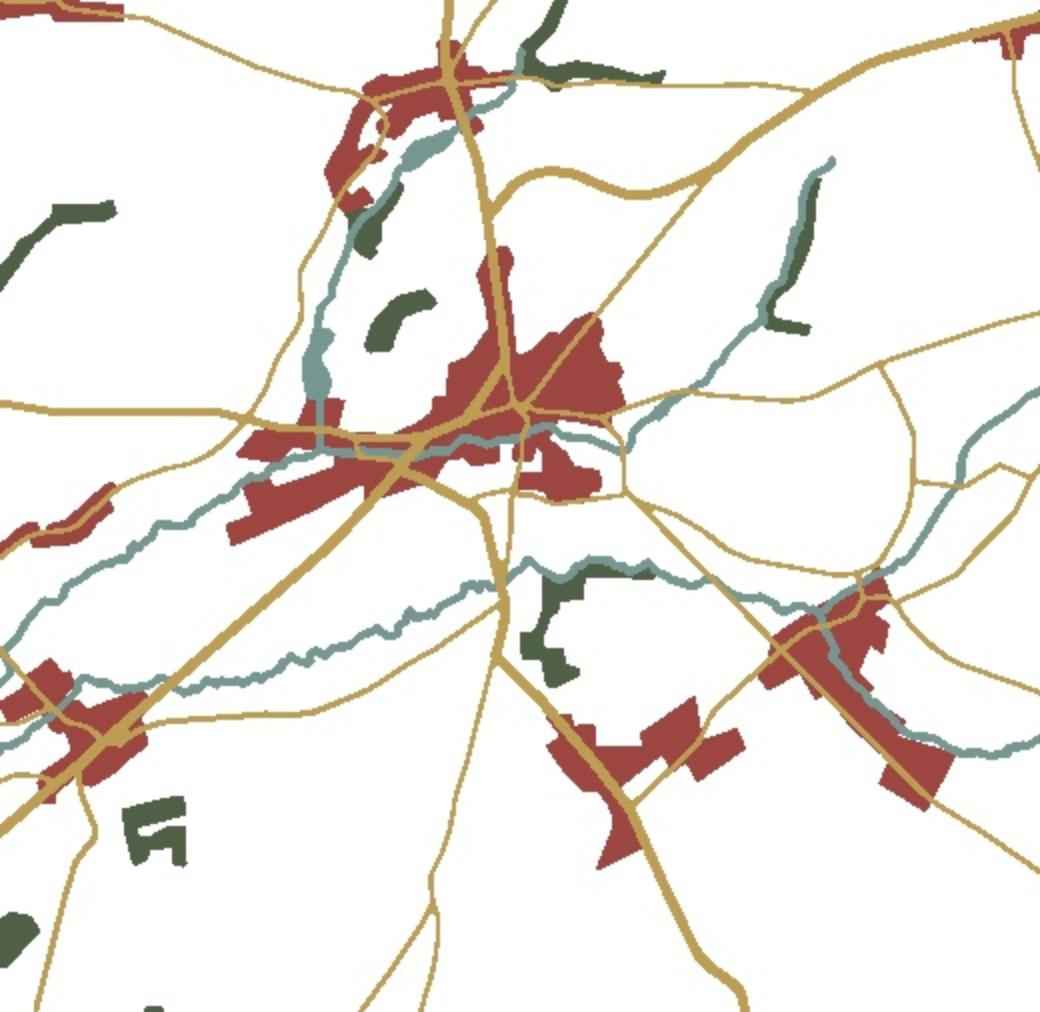}}};
    \node (em_mod) [below=0.1 of cass_mod,inner sep=0] {\fbox{\includegraphics[width=2cm]{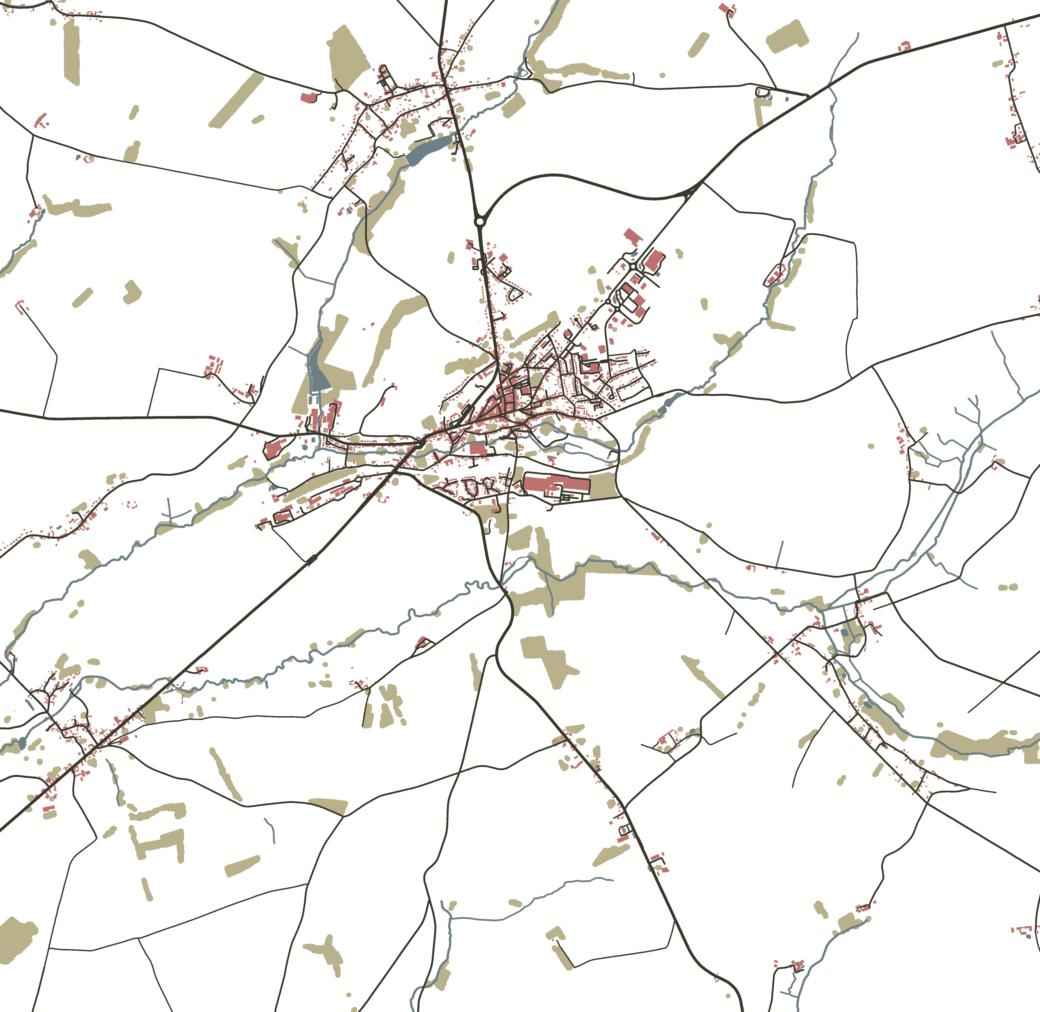}}};
    \node (fift_mod) [below=0.1 of em_mod,inner sep=0] {\fbox{\includegraphics[width=2cm]{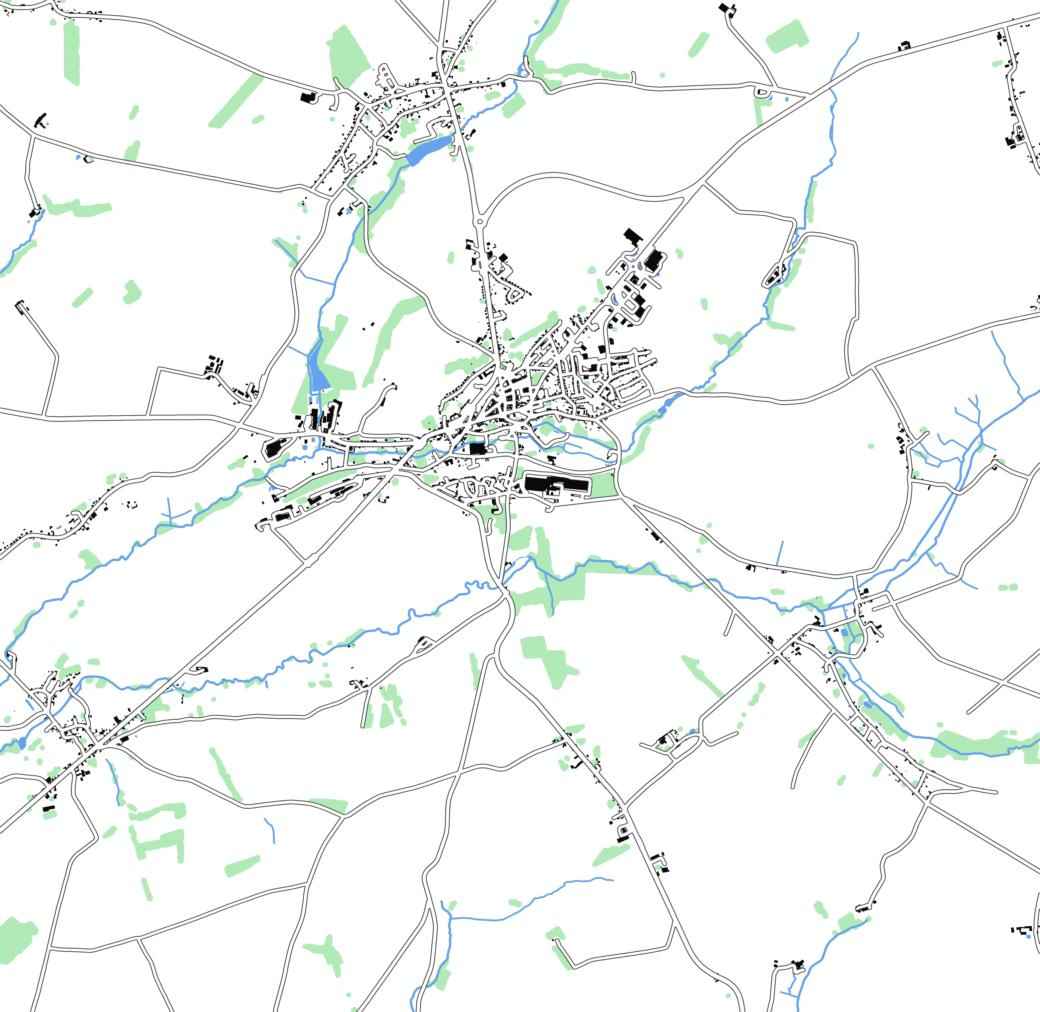}}};
    \node (synthetic_name) at ($(cass_mod.north) + (0,0.25)$) {Synthetic map};

    \draw[->, very thick] (cass_labels) -- (cass_mod);
    \draw[->, very thick] (fift_labels) -- (fift_mod);
    \draw[->, very thick] (em_labels) -- (em_mod) node[align=center,rotate=90,anchor=north,pos=0.3,fill=gray!15, text width=5.5cm,rounded corners,inner sep=5] {\bf Match color scheme};

\end{tikzpicture}

%% file: 5_experiments.tex
\section{Experiments}

In this section, we first present our metrics and evaluation setting, then discuss qualitative and quantitative results, and finally present an application to long-term forest monitoring.

\subsection{Evaluation and Metrics}

We assess the baselines presented in \cref{sec:methods} using the 470 tiles annotated for the Cassini and État-Major maps. The evaluation is based on two metrics:

\begin{table}[t]
    \centering
    \caption{\textbf{Quantitative Semantic Segmentation Results.} Overall Accuracy (OA) and dilated Intersection over Union (dIoU) for our three baselines and the two maps with available historical annotations.}
    \input{seg_quantitative}
    \label{tab:seg_quantitative}
\end{table}

\subsubsection{Overall Accuracy.}

This metric measures the proportion of correctly classified pixels across the entire test set. While this value provides a general assessment of segmentation quality, it can be biased by dominant classes, such as background pixels, and does not necessarily reflect performance on target classes.

\subsubsection{Dilated IoU.}

This metric modifies the standard Intersection over Union (IoU) to account for slight misalignments between the prediction and the ground truth. Otherwise, small shifts disproportionally impact the precision measured for thin structures like roads and rivers. Standard IoU for a given class $k$ is defined as:

\begin{equation}
    \text{IoU}_k(P_k, T_k)=\frac{|P_k \cap T_k|}{|P_k \cup T_k|}~,
\end{equation}

where $T_k$ (respectively, $P_k$) denotes the {set of pixels labeled} (respectively, predicted) as class $k$, and $\vert \cdot \vert$ denotes the size of a set. We add a tolerance to small misalignments by applying a dilation operator~\cite{serra1983image} that we write as $\text{Dil}$, which expands a set of pixels by a margin $w$ in all directions. The dilated IoU (dIoU)~\cite{cheng2021boundary} is then computed as:

\begin{equation}
    \text{dIoU}_k(P_k, T_k)=\frac{|(\text{Dil}(P_k) \cap T_k) \cup (P_k \cap \text{Dil}(T_k))|}{|P_k \cup T_k|}~.
\end{equation}

For all experiments, we use $w = 3$ pixels. When computing the classwise average of dIoUs, we exclude the background class.

\begin{figure}
    \centering
    \resizebox{0.95\columnwidth}{!}{
        \input{seg_qualitative}
    }
    \caption{\textbf{Qualitative Semantic Segmentation Results.} Predicted segmentation of our three baselines and  SAM~\cite{kirillov2023segment}, on the same area across three centuries.}
    \label{fig:seg_qualitative}
\end{figure}
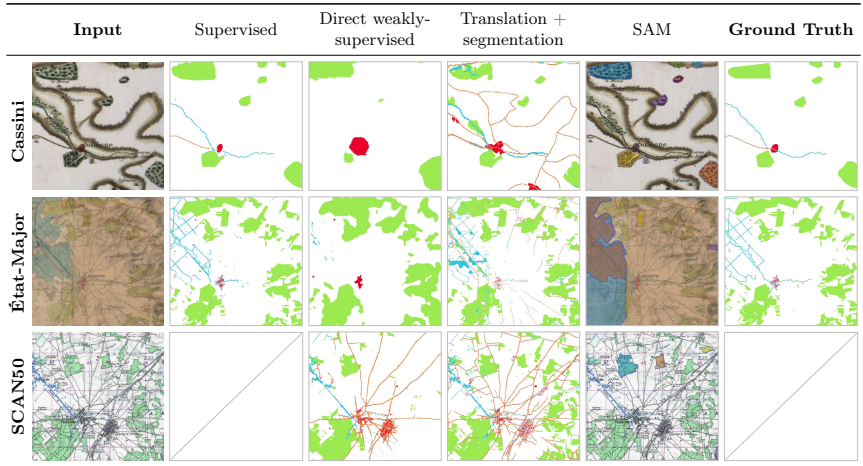

\subsubsection{Evaluation Setting.}

To segment large tiles, we split them into patches of the same size used during training, without overlaps. As the performance of our translation + segmentation approach varies more across training runs than for the other models, we report the average of three models trained independently. The standard deviation for OA and mean dIoU is below 1.3 for Cassini, and 4.3 for État-Major.

\subsection{Results and Analysis}

We report quantitative and qualitative results for the semantic segmentation performance, a qualitative analysis of the image-to-image translation module and an ablation study.

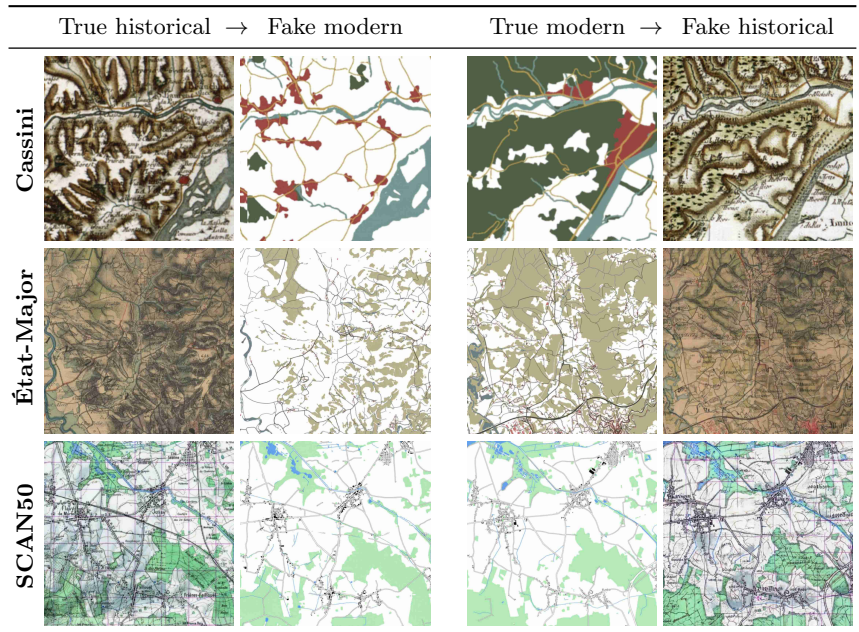
\begin{figure}[t]
    \centering
    \resizebox{0.95\columnwidth}{!}{
        \input{im2im_translation}
    }
    \caption{\textbf{Image-to-Image Translation.} Examples of historical-to-modern and modern-to-historical generation with our image translation model (\cref{cyclegan_details}).}
    \label{fig:im2im_translation}
\end{figure}

\subsubsection{Semantic Segmentation.}
We report in \cref{tab:seg_quantitative} the performance of our baselines on the Cassini and État-Major maps, and illustrate their predictions in \cref{fig:seg_qualitative}. As expected, the fully-supervised method yields the best performance.

On the Cassini map, the translation + segmentation baseline consistently outperforms the other weakly-supervised approach, with forests, rivers, and roads all showing better recall. However, the relief lines of the Cassini maps are often misinterpreted as roads, causing lower precision, and the model also hallucinates roads and buildings. This behavior is unsurprising given that it is trained on modern labels, where roads and urban areas are more prevalent.

For the État-Major maps, direct weakly-supervised segmentation exhibits higher numerical performance, but qualitatively the translation + segmentation baseline delineates individual buildings, forest boundaries, roads, and rivers more accurately. Because the annotations for these maps only include main roads, predicting secondary roads is penalized even when they are correct. 

Our analysis of the SCAN50 map is purely qualitative, as we do not have access to historical labels. Yet the results appear of high quality, likely because the smaller time gap between SCAN50 and modern reference data makes the modern labels more relevant. The translation + segmentation approach also produces more granular predictions for SCAN50, even detecting individual buildings.

Additionally, we illustrate in \cref{fig:seg_qualitative} the segmentation predicted by the foundation model SAM~\cite{kirillov2023segment} in a zero-shot setting. On the Cassini map, SAM accurately delineates forests and towns but fails to retrieve roads and rivers. Its predictions on the État-Major and SCAN50 maps are significantly worse. This suggests that the model, trained on natural images, struggles to generalize to historical maps, whose cartographic styles differ significantly from the data it was trained on.

In summary, our translation + segmentation baseline is arguably the most promising weakly-supervised approach. We thus analyze it further in the next paragraphs.

\subsubsection{Qualitative Image-to-Image Translation Analysis.}

In \cref{fig:im2im_translation}, we illustrate examples of image-to-image translation for all datasets in both directions: historical-to-modern and modern-to-historical. The generated historical and modern maps appear visually credible, capturing the characteristic stylistic elements of their target domain, such as shading and color schemes. For SCAN50, the content is generally accurately translated, but this is not the case for Cassini and État-Major. We believe this is mainly due to the strong differences in the represented features and the strong distribution shift in classes. For example, most marshes from the \nth{18} century have disappeared, while road networks and urban areas have greatly expanded, leading to frequent hallucinations in the generated maps.

\subsubsection{Ablation Study.}

We assess the impact of two factors on our translation + segmentation baseline: (i) the color and level of detail adaptation of the synthetic modern maps to each historical map (\cref{fig:modern_maps_workflow}), and (ii) the addition of the translation loss $\mathcal{L}_{\text{tran}}$ {(\cref{cyclegan_details})}. The corresponding quantitative and qualitative segmentation results are collected in \cref{tab:ablation_seg_quantitative} and \cref{fig:ablation_seg_qualitative} respectively. In general, qualitative results highlight the benefits of combining data adaptation with the translation loss on all datasets.

More in detail, data adaptation consistently helps on the Cassini maps except for building segmentation. On État-Major, adapting color and detail improves hydrography and roads but degrades other classes. The increased level of detail allows the model to segment thinner roads and smaller towns, but also introduces more false positives. Incorporating a translation loss improves the forest and hydrography segmentation for Cassini and État-Major, with \cref{tab:ablation_seg_quantitative} showing a notable increase in dIoU and \cref{fig:ablation_seg_qualitative} highlighting the improved forest precision for these two maps. For SCAN50, the impact of the translation loss is more limited as data adaptation appears sufficient to bridge the style gap.

\begin{table}[t]
    \centering
    \caption{\textbf{Quantitative Ablation Study.} Impact of the color and level of detail matching of the modern maps (``Match'', see \cref{fig:modern_maps_workflow}), and the translation loss ($\mathcal{L}_{\text{tran}}$) on segmentation quality for our translation + segmentation baseline.}
    \input{ablation_seg_quantitative}
    \label{tab:ablation_seg_quantitative}
\end{table}

\begin{figure}[htbp]
    \centering
    \resizebox{0.95\columnwidth}{!}{
        \input{ablation_seg_qualitative}
    }
    \caption{\textbf{Qualitative Ablation Study.} Impact of  color and level of detail matching of the modern maps (``Match'', see \cref{fig:modern_maps_workflow}), and translation loss ($\mathcal{L}_{\text{tran}}$) on segmentation quality for our translation + segmentation baseline.}
    \label{fig:ablation_seg_qualitative}
\end{figure}

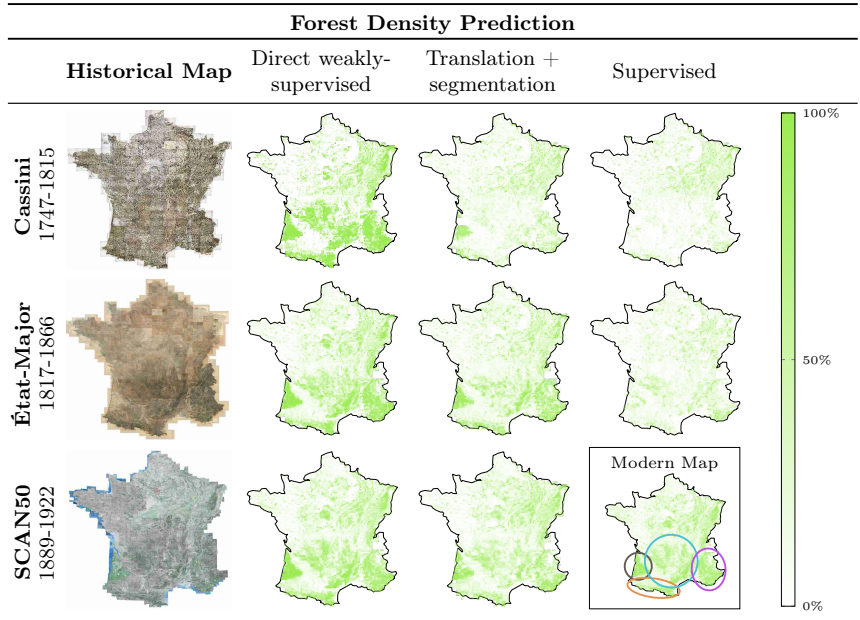
\begin{figure}[t]
    \centering
    \resizebox{0.95\columnwidth}{!}{
        \input{forest_density}
    }
    \caption{\textbf{Predicted Forest Cover over France Across Four Centuries.} The framed map at the bottom right is the current forest density derived from contemporary sources. The colored ovals highlight key areas in our analysis: the Landes\;\textcolor{nice_brown}{\faCircle[regular]}, Pyrenees\;\textcolor{nice_orange}{\faCircle[regular]}, Massif Central\;\textcolor{nice_blue}{\faCircle[regular]}, and Provence-Alps region\;\textcolor{nice_purple}{\faCircle[regular]}.}
    \label{fig:forest_density}
\end{figure}

\subsection{Application: Long-Term Forest Monitoring}

We evaluate the potential of our approach for long-term forest monitoring by tracking changes in predicted forest cover across metropolitan France from the \nth{18} century to the present day. For each baseline and map collection, we use the predicted forest pixels to compute density maps and plot the results in \cref{fig:forest_density}. Since the supervised baseline achieves the highest scores in \cref{tab:seg_quantitative}, its predictions can be regarded as the most reliable estimate of historical forest extent.

While the translation + segmentation baseline appears more precise than the direct weakly-supervised segmentation baseline in our experiments, it still over-predicts forest in the Landes region (\textcolor{nice_brown}{\faCircle[regular]}) and mountainous areas (\textcolor{nice_orange}{\faCircle[regular]}, \textcolor{nice_blue}{\faCircle[regular]}, \textcolor{nice_purple}{\faCircle[regular]}). This comes from often confusing marshes and relief shading with tree cover—an effect of relying on modern labels, where forests are more abundant.

Despite these caveats, our approach enables the reconstruction of three centuries of forest cover evolution across metropolitan France. Using the most reliable models (supervised baseline for the Cassini and État-Major collections, and translation + segmentation baseline for SCAN50), we identify trends that broadly align with historical records.

\subsubsection{An Historical Perspective.}

From the Cassini maps, we see particularly dense woodlands in the northeast during the \nth{18} century, which agrees with historical accounts of valuable forests in Franche-Comté~\cite{matteson2015forests}. Overall, forest cover has steadily increased over the last centuries, paralleling the rise of conservation policies in the \nth{19} century—most notably, the founding of the first French National School of Forestry in Nancy in 1824, the Forestry Code of 1827, and subsequent reforestation laws in 1860 and 1882 targetting the Massif Central, the Pyrenees and the Alps~\cite{scheifley1920depleted}. Improved farmland productivity and the shift from wood to fossil fuels accelerated this trend~\cite{cinotti1996evolution}. In the southwest, we observe a marked increase in forested areas, reflecting the long-term transformation of marshlands in the Landes region from the \nth{18} to the \nth{20} centuries~\cite{ford2023environmental}.

%% file: seg_quantitative.tex
\begin{tabular}{l@{\;\;}l@{\;\;}c@{\;\;}c@{\;\;}c@{\;\;}c@{\;\;}c@{\;\;}c}
    \toprule
    \multirow{2}{*}{\textbf{Map}} &
    \multirow{2}{*}{\textbf{Baseline}} &
    \multirow{2}{*}{\textbf{OA}} &
    \textbf{Mean} &
    \multicolumn{4}{c}{\textbf{Per-class dIoU}} \\ \cline{5-8}
    & & & \textbf{dIoU}
    & \raisebox{-0.7mm}{\tree}
    & \raisebox{-0.7mm}{\building}
    & \raisebox{-0.7mm}{\water}
    & \raisebox{-0.7mm}{\road} \\ \midrule
    \multirow{3}{*}{\textbf{Cassini}}
    & \gtext{Supervised} & \gtext{96.7} & \gtext{76.4} & \gtext{88.3} & \gtext{69.8} & \gtext{85.0} & \gtext{62.6} \\
    \arrayrulecolor{gray}\cmidrule{2-8}\arrayrulecolor{black}
    & Direct weakly-supervised & 79.8 & 26.1 & 35.8 & \textbf{5.3} & 63.1 & 0.0 \\
    & Translation + segmentation & \textbf{85.3} & \textbf{36.1} & \textbf{56.1} & 4.7 & \textbf{70.9} & \textbf{12.7} \\ \midrule
    
    \multirow{3}{*}{\textbf{État-Major}} 
    & \gtext{Supervised} & \gtext{91.3} & \gtext{61.2} & \gtext{77.4} & \gtext{56.3} & \gtext{65.4} & \gtext{46.0}  \\
    \arrayrulecolor{gray}\cmidrule{2-8}\arrayrulecolor{black}
    & Direct weakly-supervised & \textbf{83.3} & \textbf{38.6} & \textbf{49.6} & \textbf{39.6} & \textbf{58.6} & \textbf{6.3} \\
    & Translation + segmentation & 78.4 & 28.7  & 43.0  & 15.6  & 51.0  & 5.3  \\ \bottomrule
\end{tabular}

%% file: seg_qualitative.tex
\begin{tabular}{ccccccc}
    \toprule
    & \multirow{2}{*}{\textbf{Input}} 
    & \multirow{2}{*}{Supervised} 
    & Direct weakly- 
    & Translation + 
    & \multirow{2}{*}{SAM}
    & \multirow{2}{*}{\textbf{Ground Truth}} \\ 
    & & & supervised & segmentation & & \\ \midrule
    \raiseandrotate{.425}{\textbf{Cassini}} & 
    \includegraphics[width=0.2\linewidth]{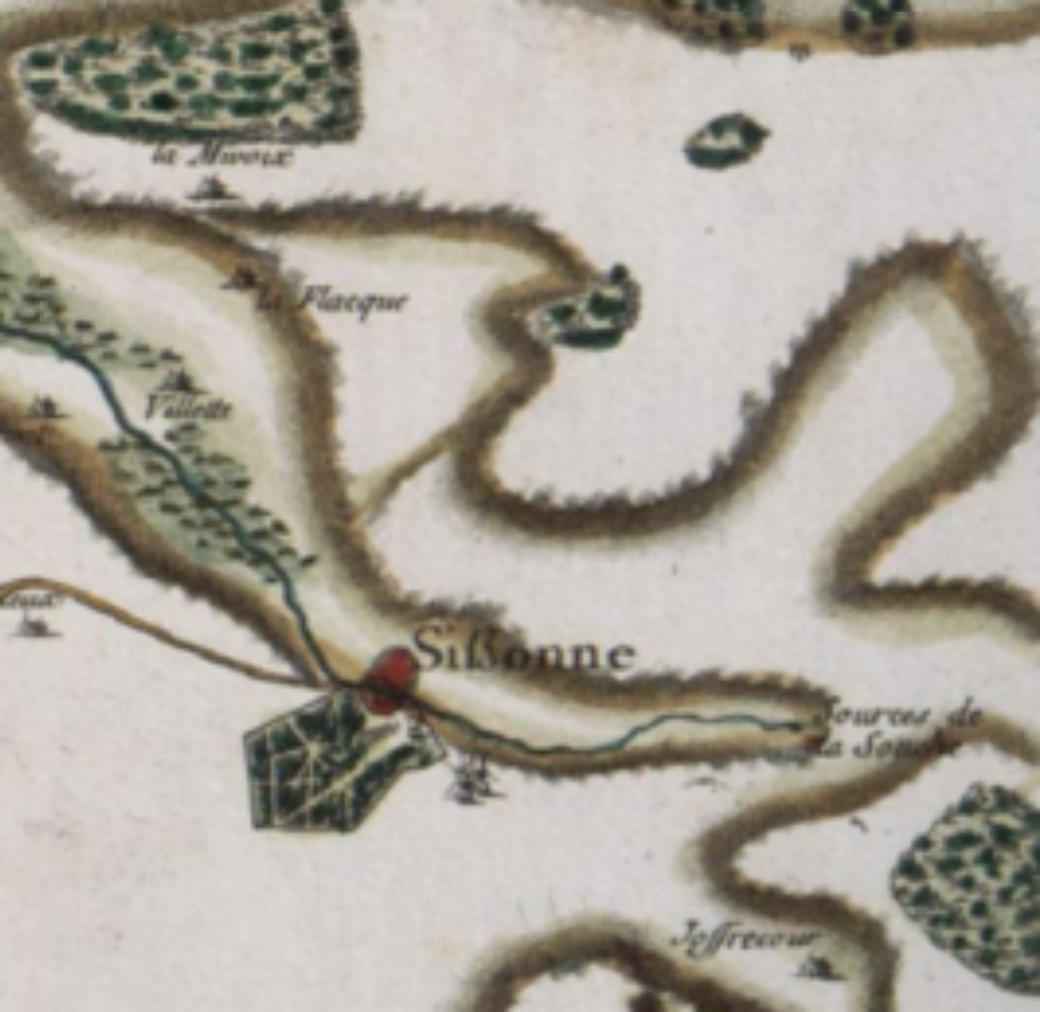} &
    \fbox{\includegraphics[width=0.2\linewidth]{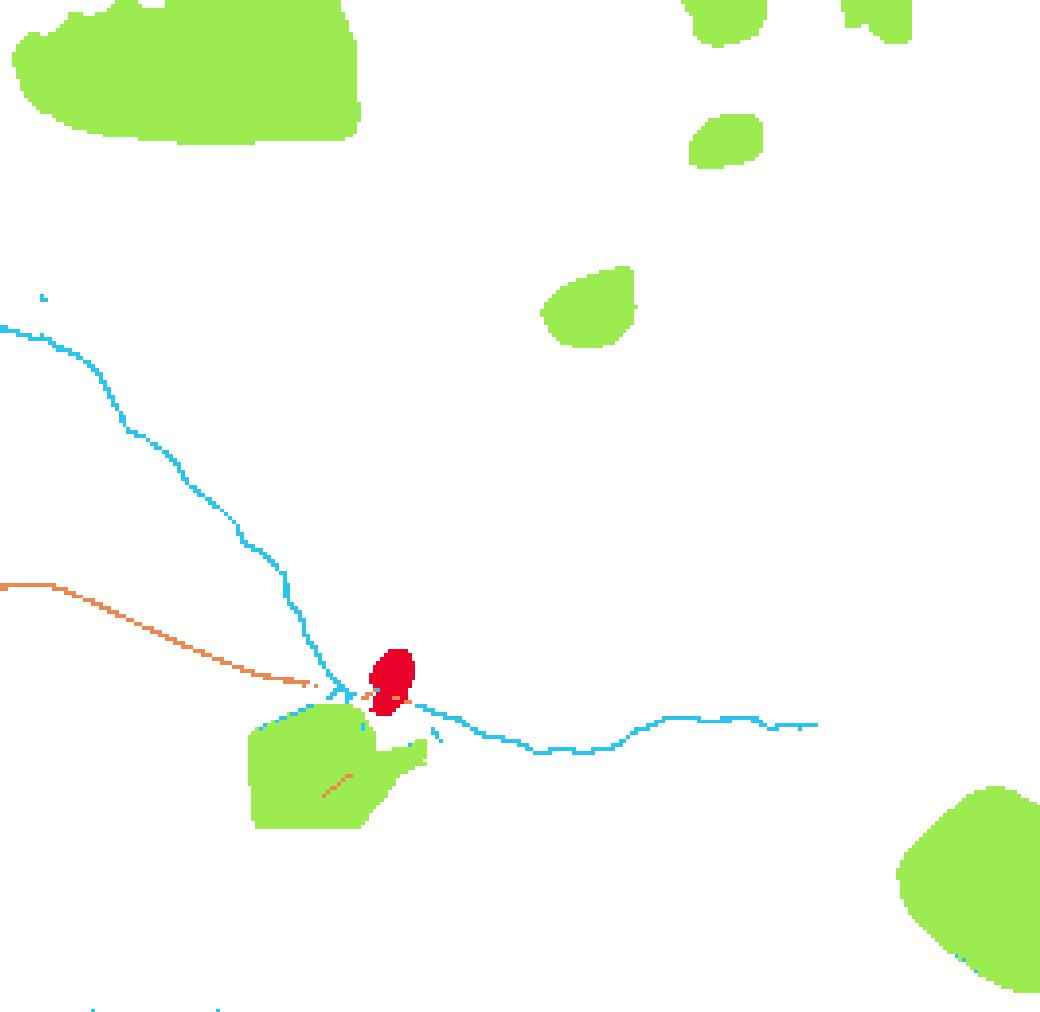}} &
    \fbox{\includegraphics[width=0.2\linewidth]{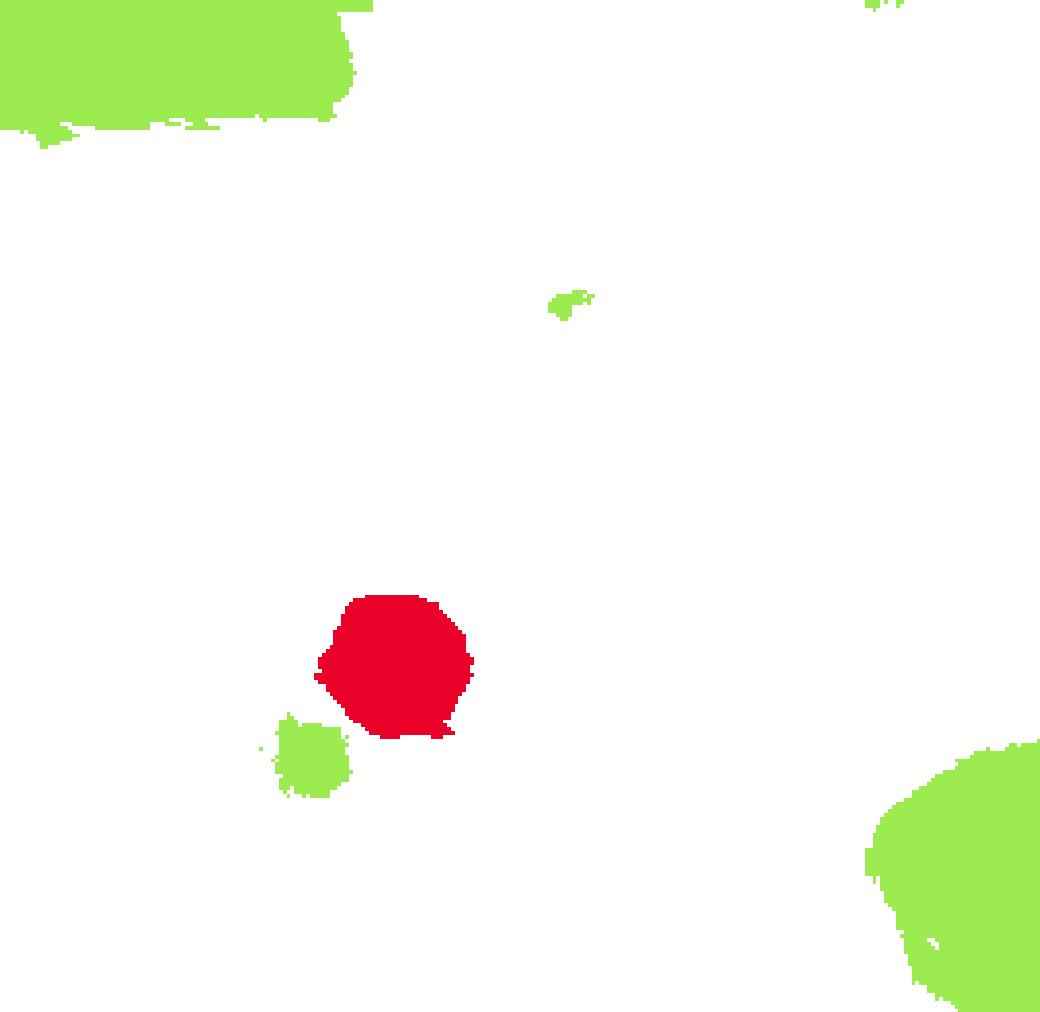}} & 
    \fbox{\includegraphics[width=0.2\linewidth]{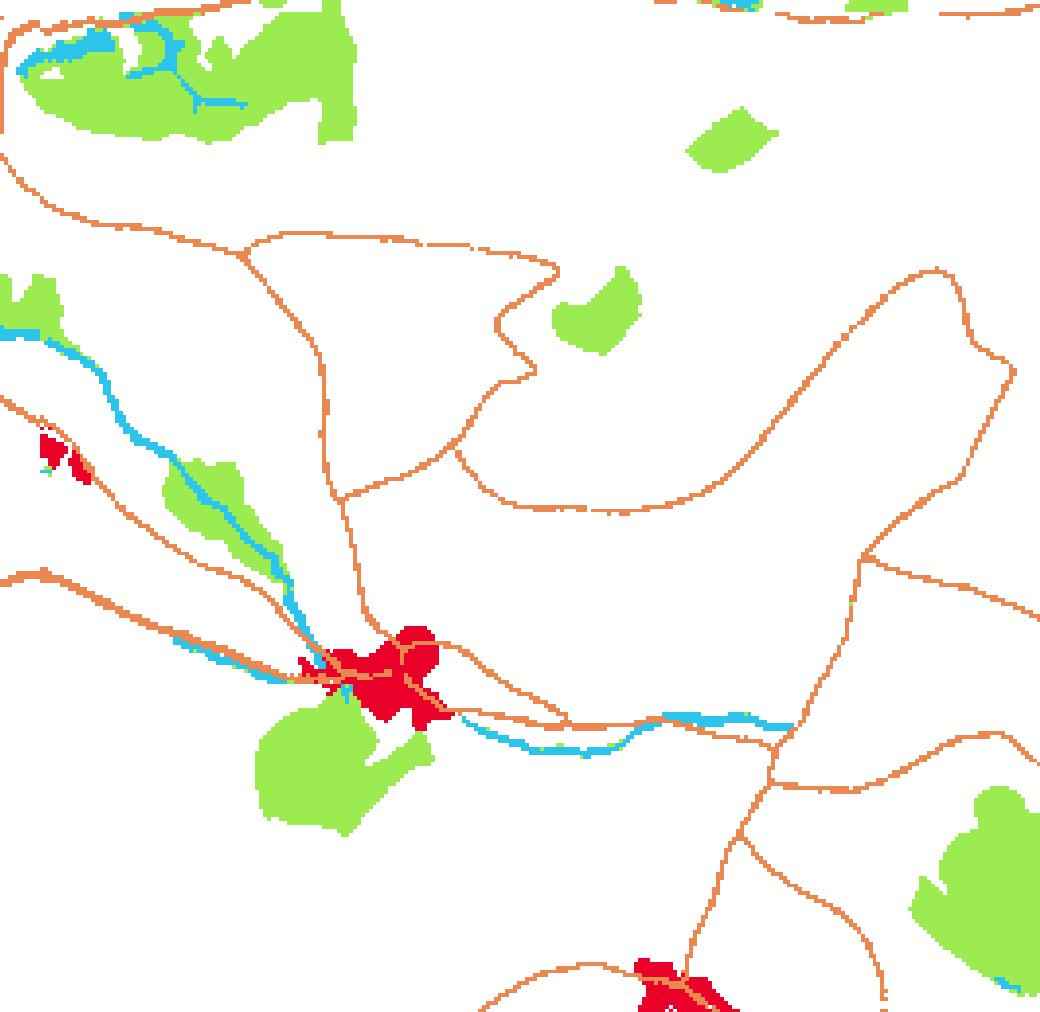}} &
    \fbox{\includegraphics[width=0.2\linewidth]{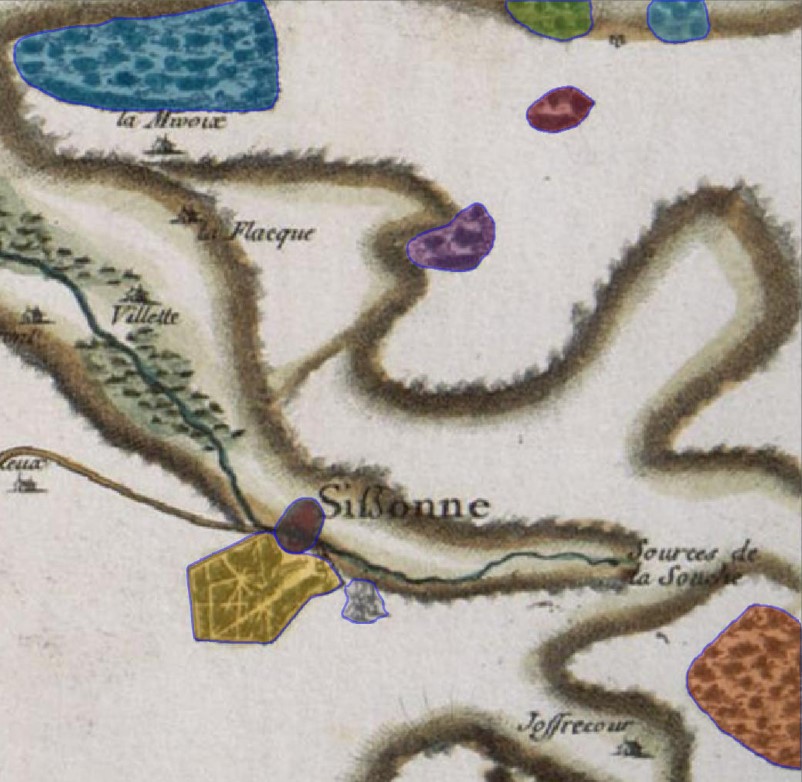}} &
    \fbox{\includegraphics[width=0.2\linewidth]{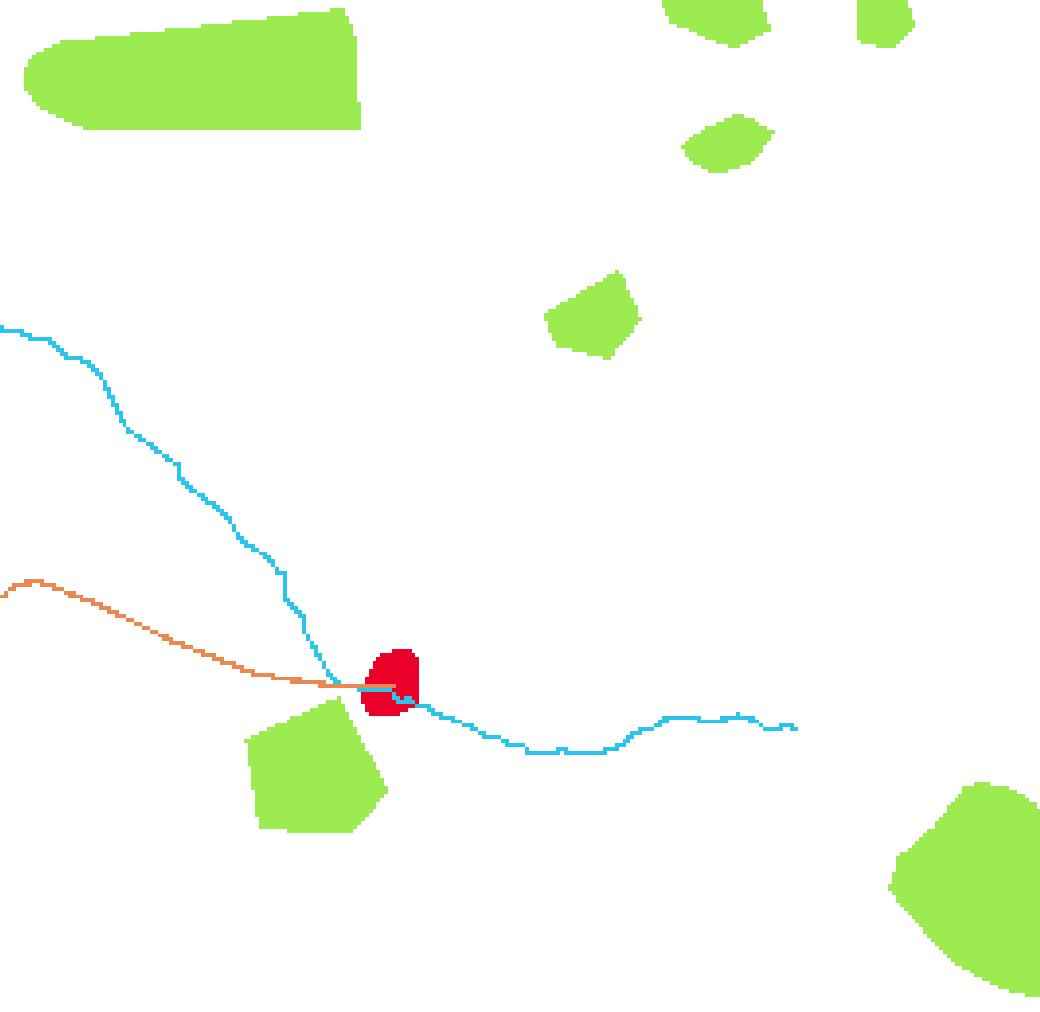}} \\
    
    \raiseandrotate{.125}{\textbf{État-Major}} &
    \includegraphics[width=0.2\linewidth]{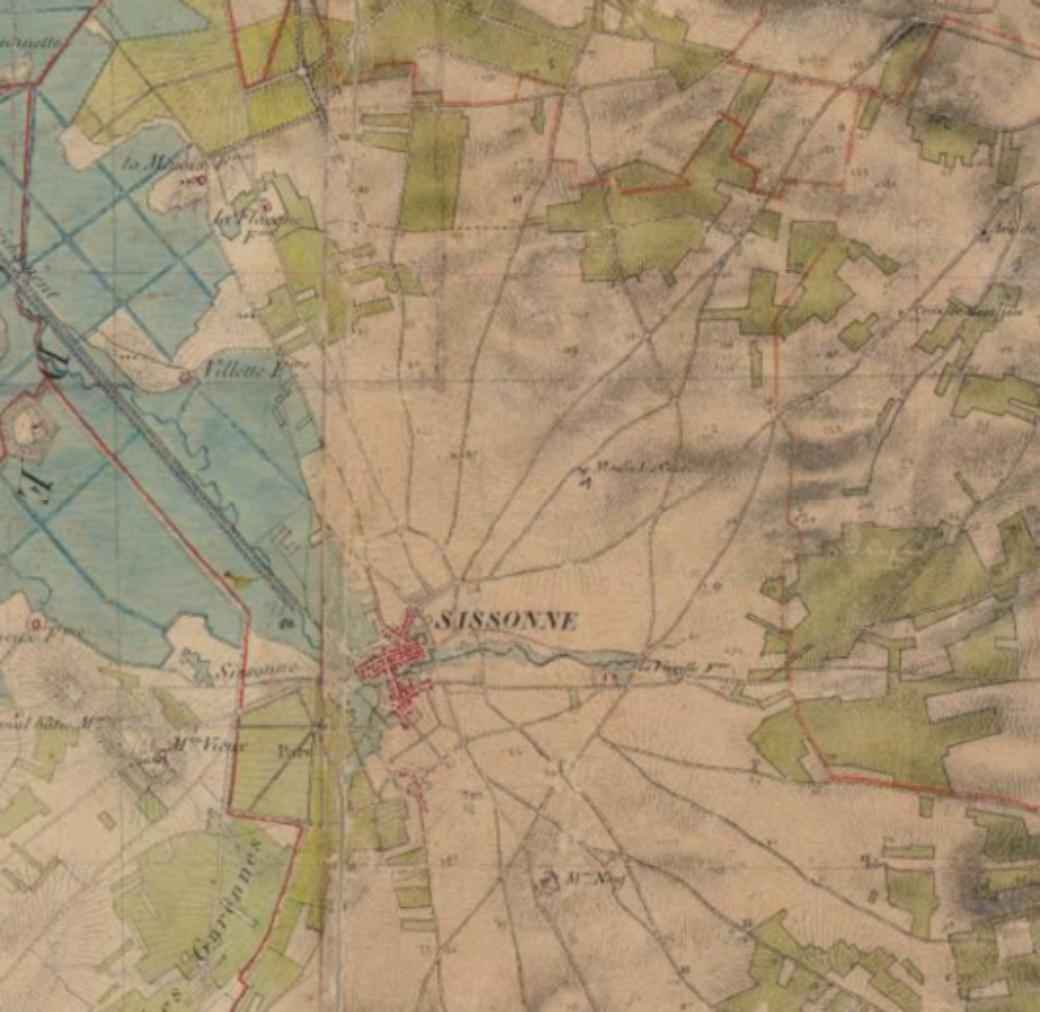} & 
    \fbox{\includegraphics[width=0.2\linewidth]{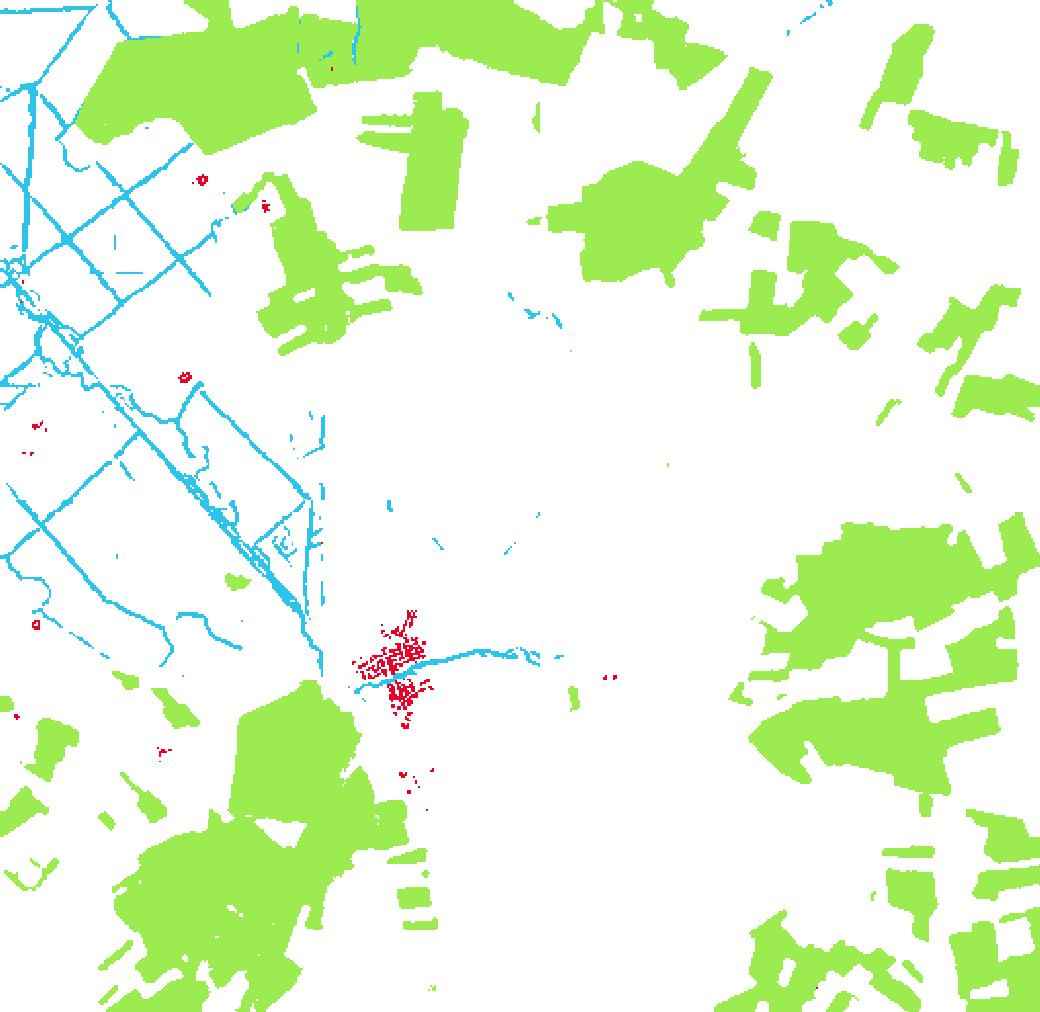}} &
    \fbox{\includegraphics[width=0.2\linewidth]{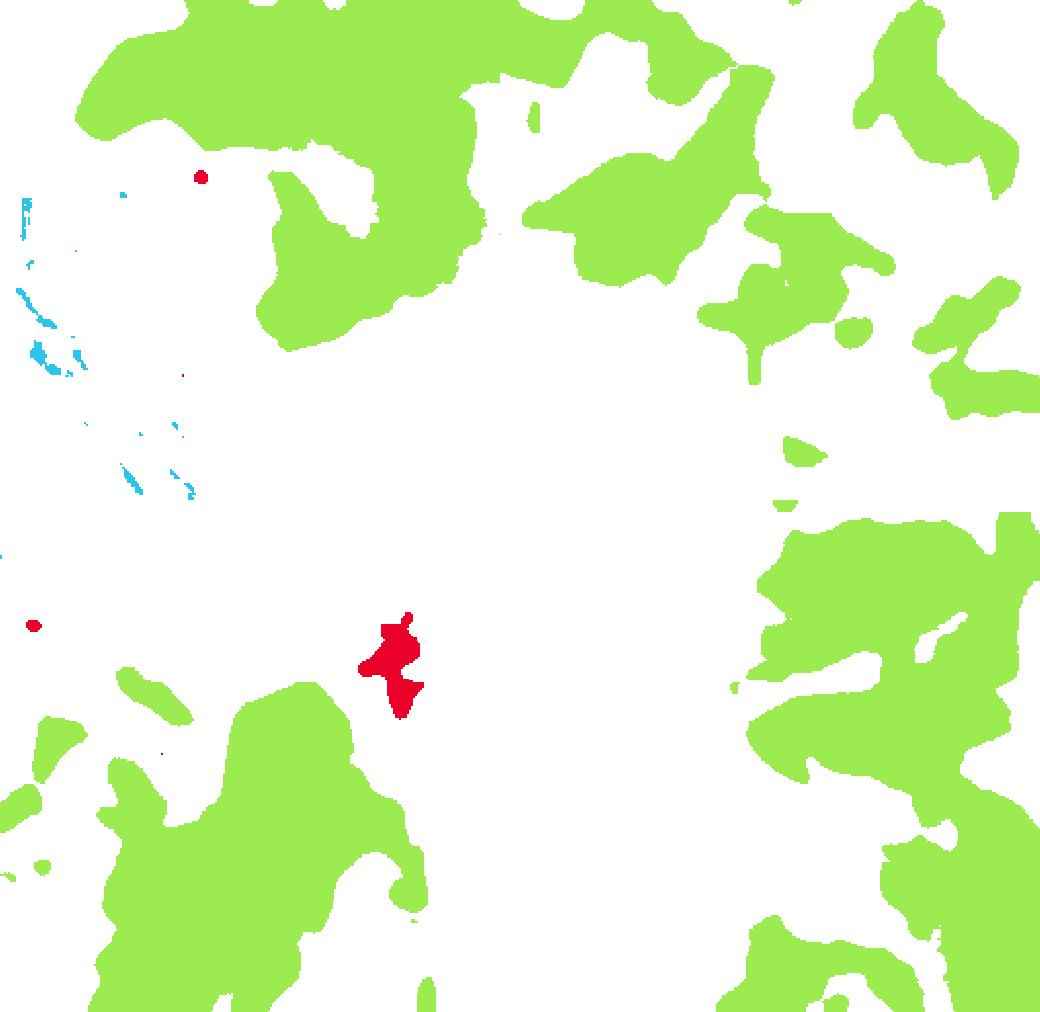}} &
    \fbox{\includegraphics[width=0.2\linewidth]{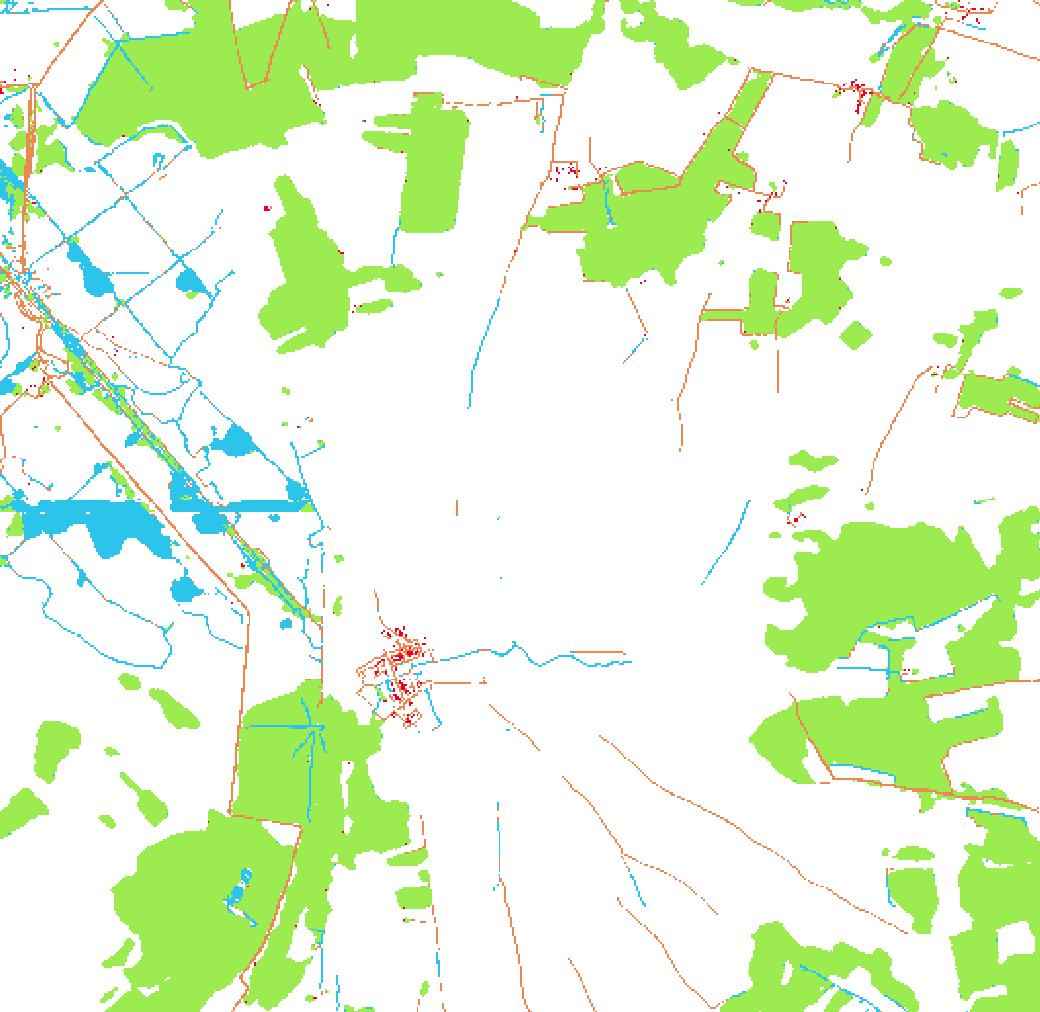}} &
    \fbox{\includegraphics[width=0.2\linewidth]{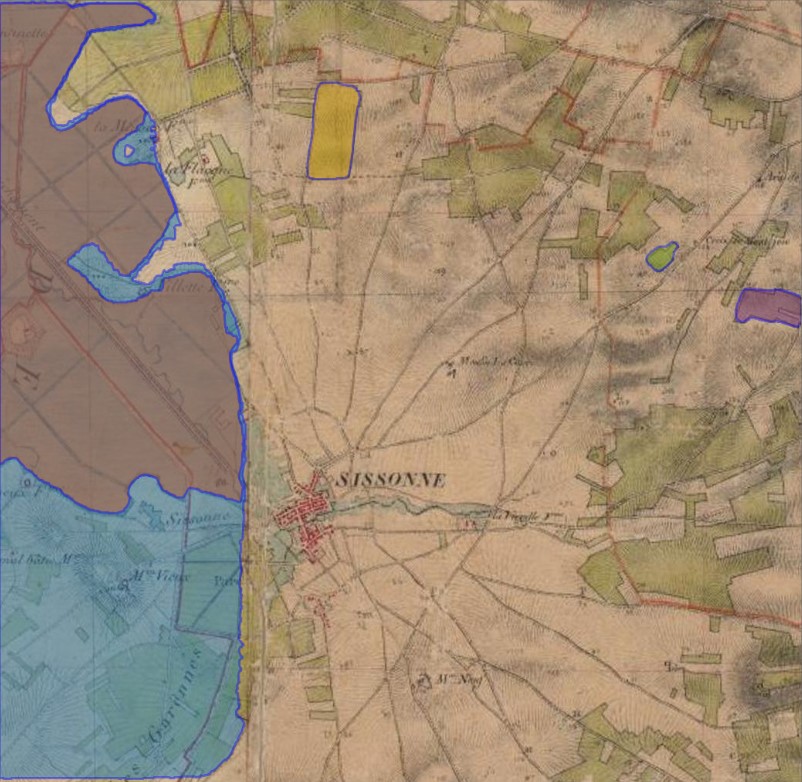}} &
    \fbox{\includegraphics[width=0.2\linewidth]{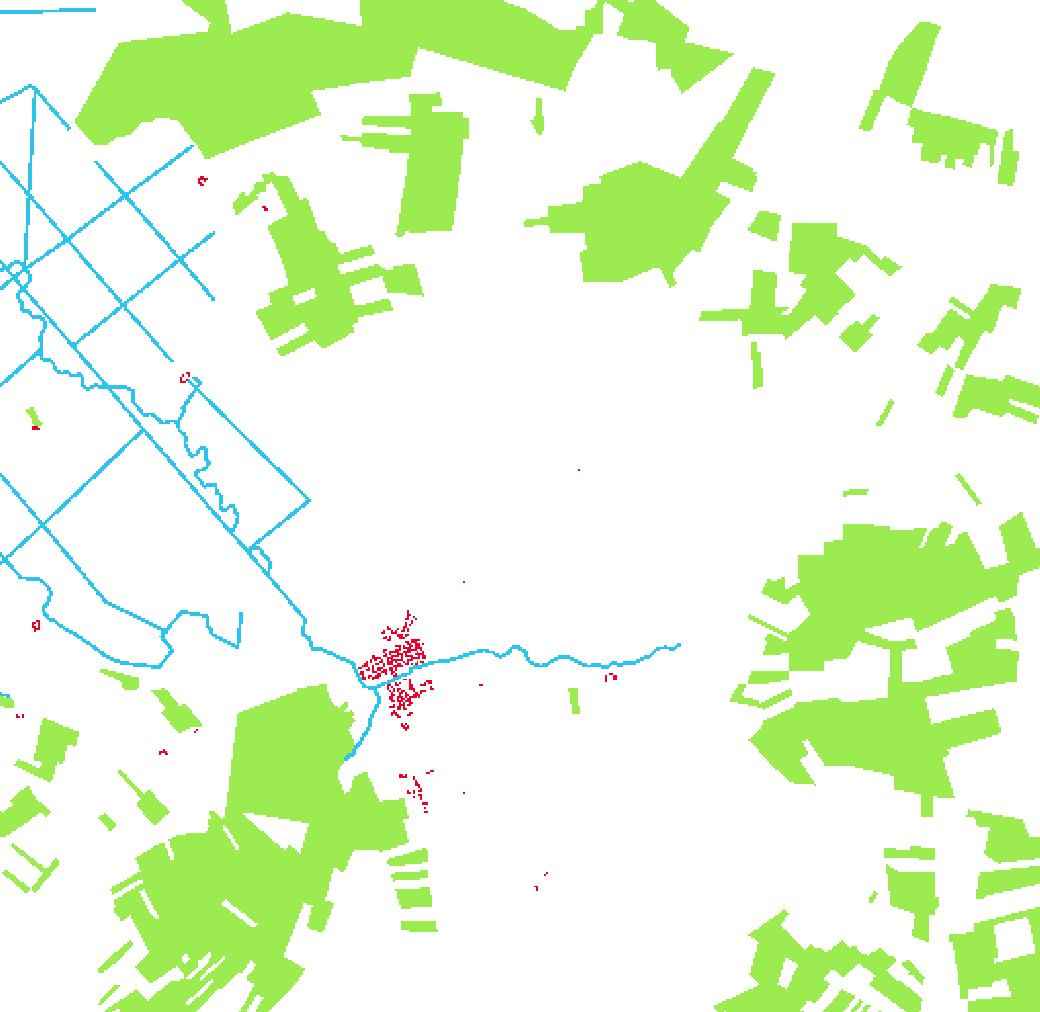}}\\
    
    \raiseandrotate{.325}{\textbf{SCAN50}} &
    \includegraphics[width=0.2\linewidth]{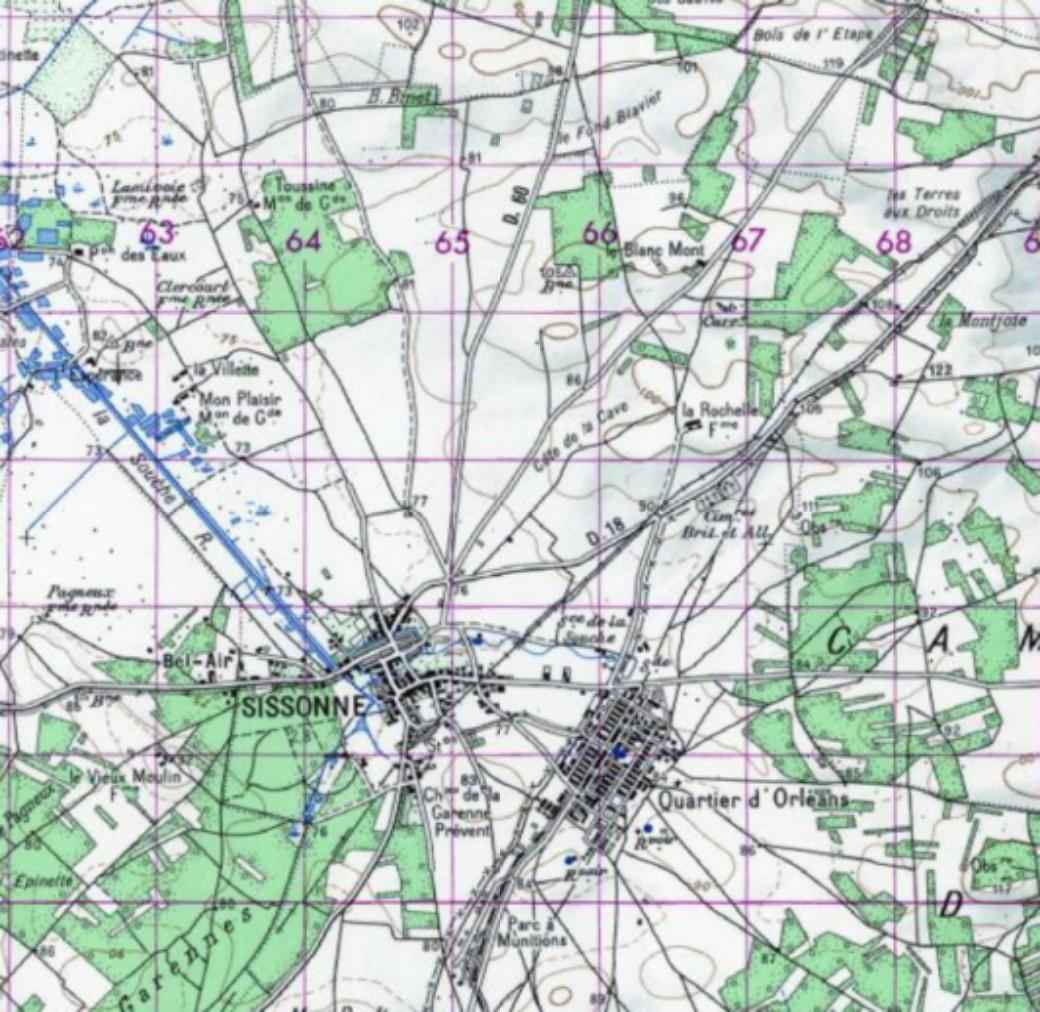} &
    \fbox{\begin{tikzpicture}[inner sep=0pt]
        \draw [draw=gray!50] (0,0) -- (2.44,2.35);
    \end{tikzpicture}} & 
    \fbox{\includegraphics[width=0.2\linewidth]{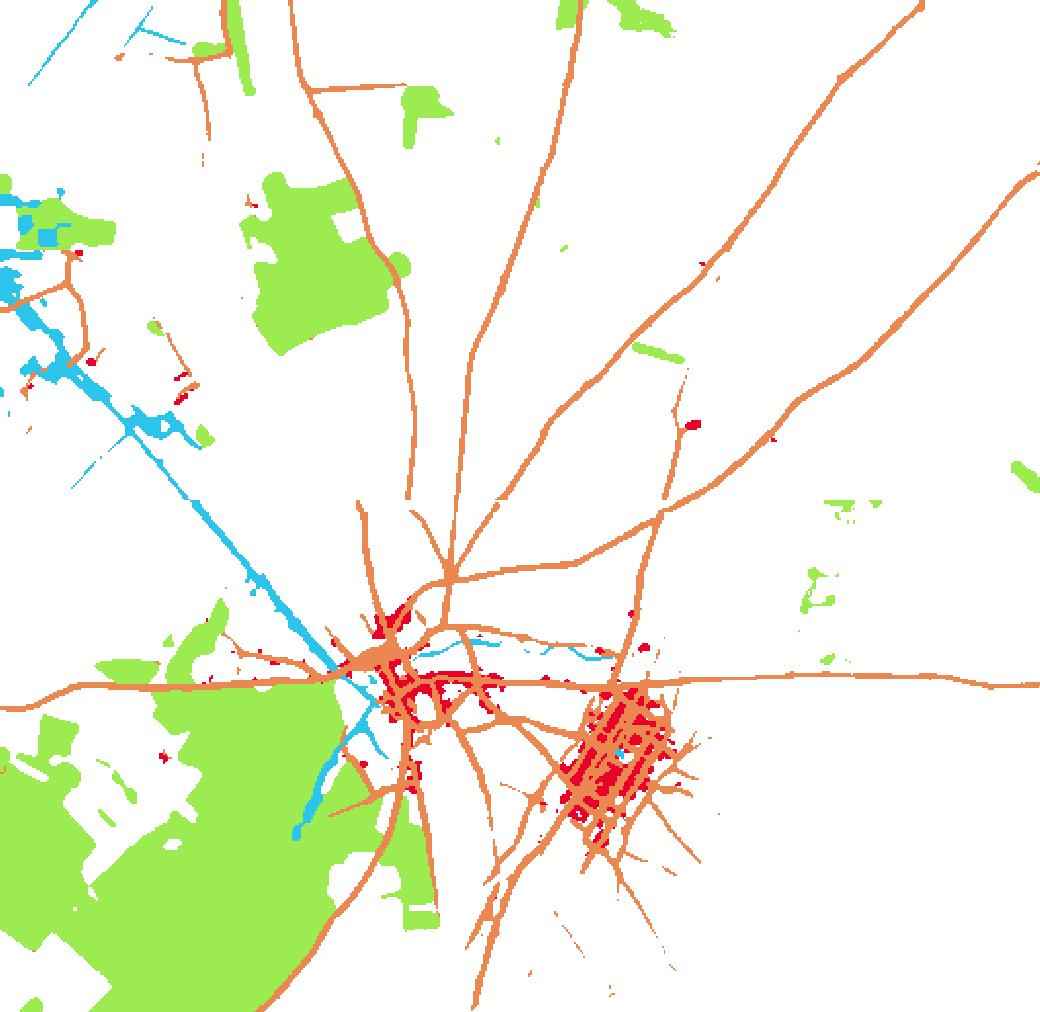}} &
    \fbox{\includegraphics[width=0.2\linewidth]{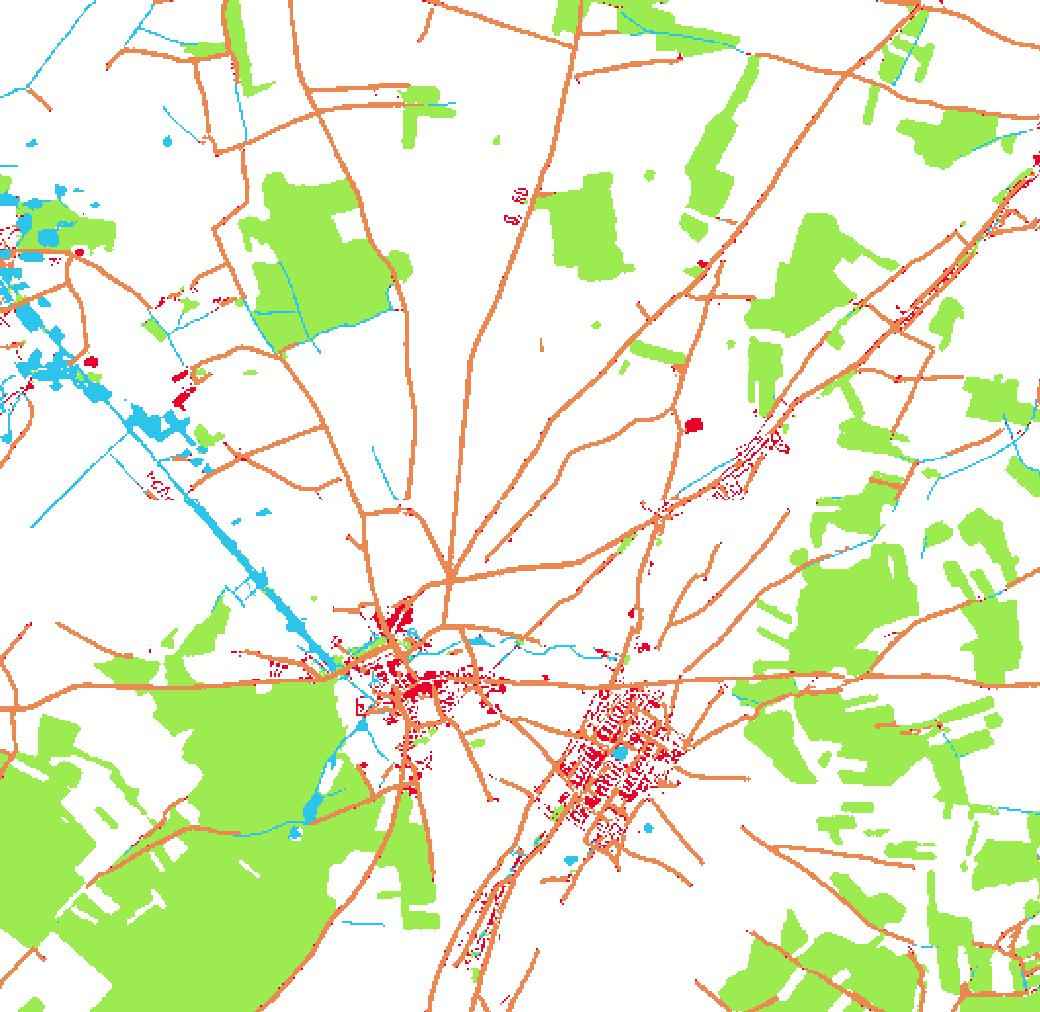}} &
    \fbox{\includegraphics[width=0.2\linewidth]{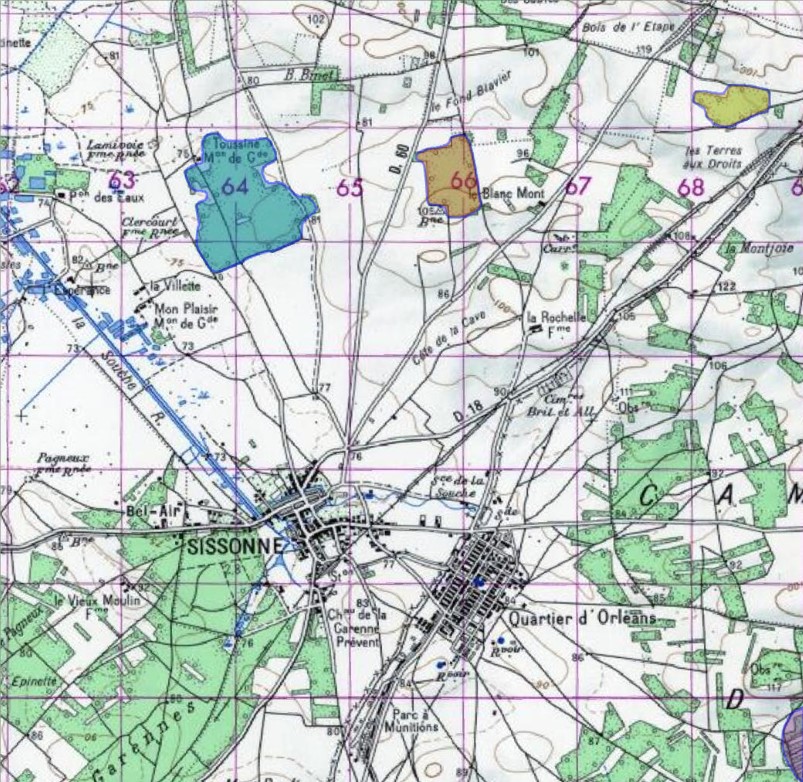}} &
    \fbox{\begin{tikzpicture}[inner sep=0pt]
        \draw [draw=gray!50] (0,0) -- (2.44,2.35);
    \end{tikzpicture}} 
    \\\bottomrule
\end{tabular}

%% file: im2im_translation.tex
\begin{tabular}{ccc@{\hspace{0.5cm}}cc}
    \toprule
    & True historical \makebox[0pt][l]{\:$\rightarrow$}
    & Fake modern
    & True modern \makebox[0pt][l]{\:$\rightarrow$}
    & Fake historical \\ \midrule
    \raiseandrotate{.55}{\textbf{Cassini}} & 
    \includegraphics[width=0.21\linewidth]{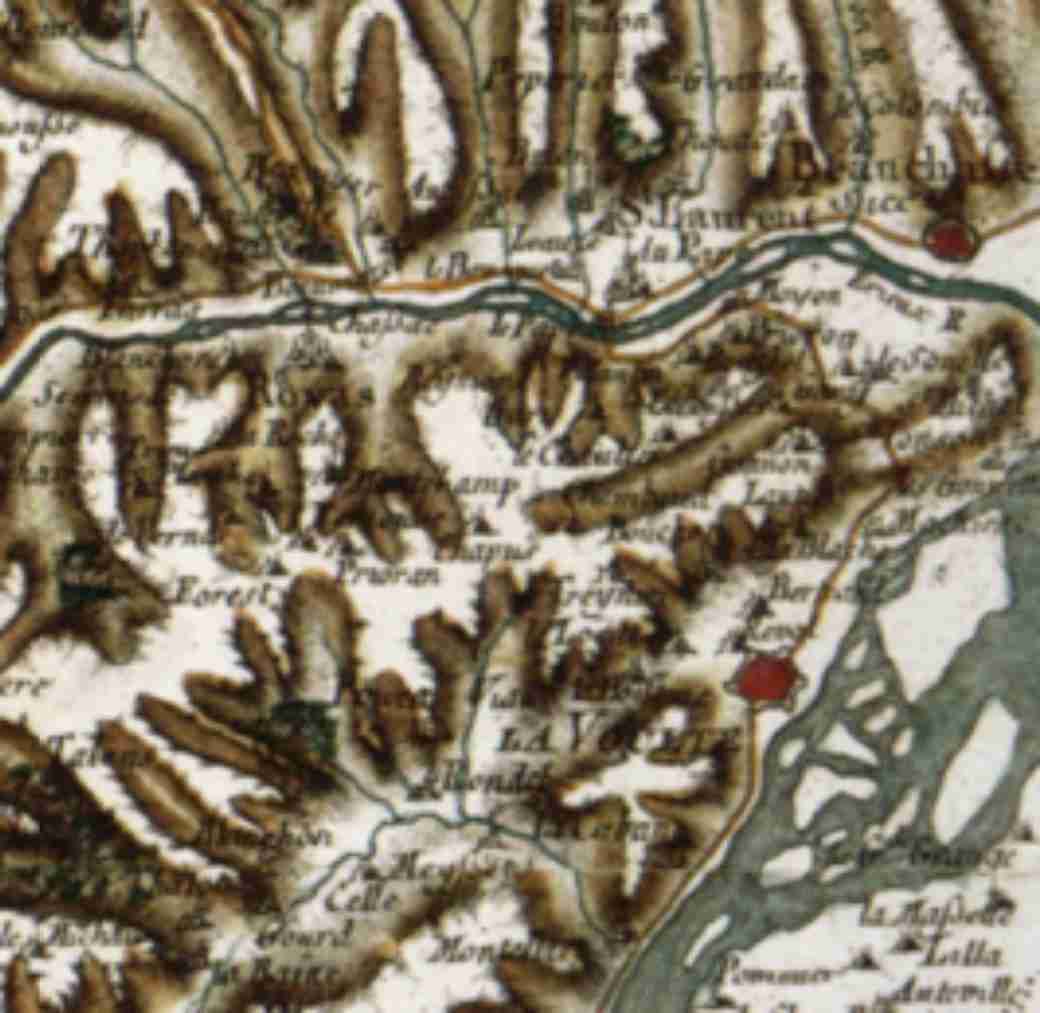} &
    \includegraphics[width=0.21\linewidth]{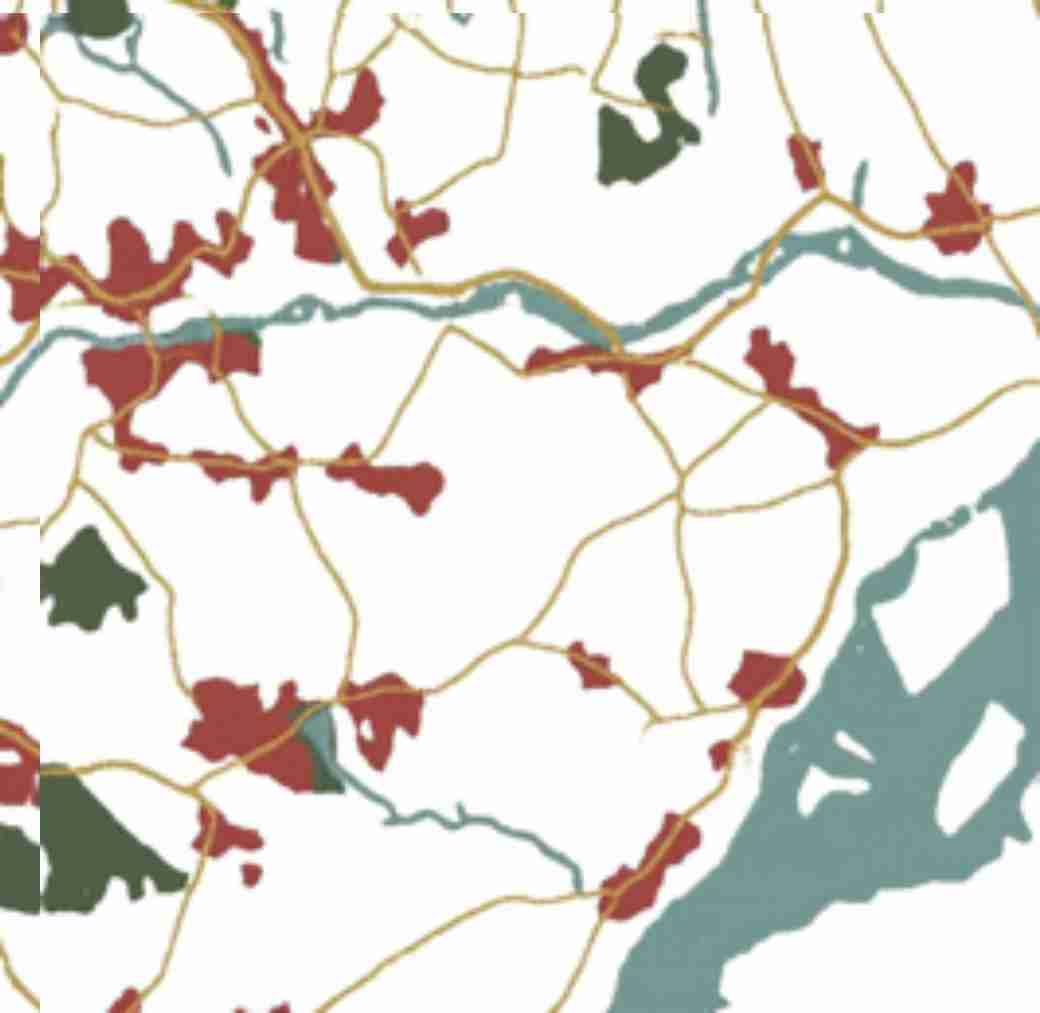} &
    \includegraphics[width=0.21\linewidth]{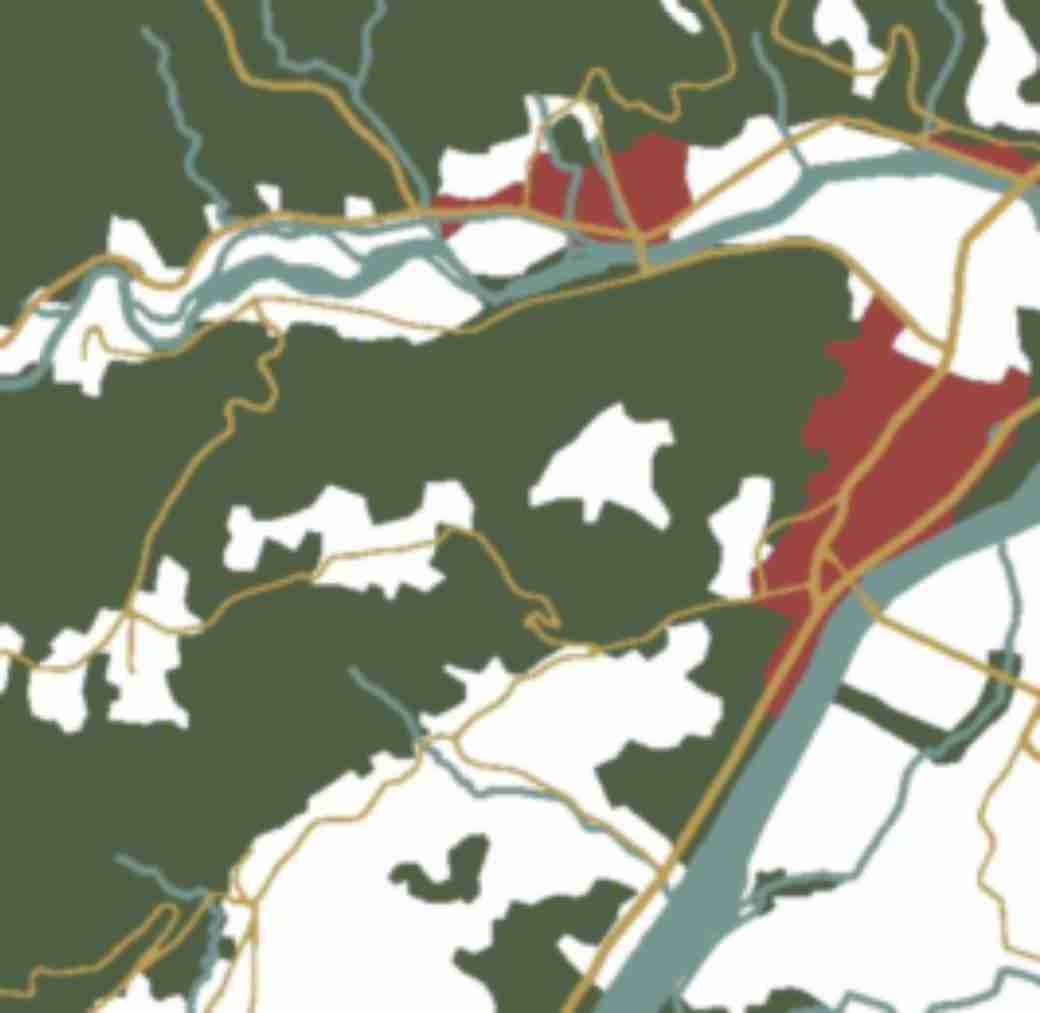} &
    \includegraphics[width=0.21\linewidth]{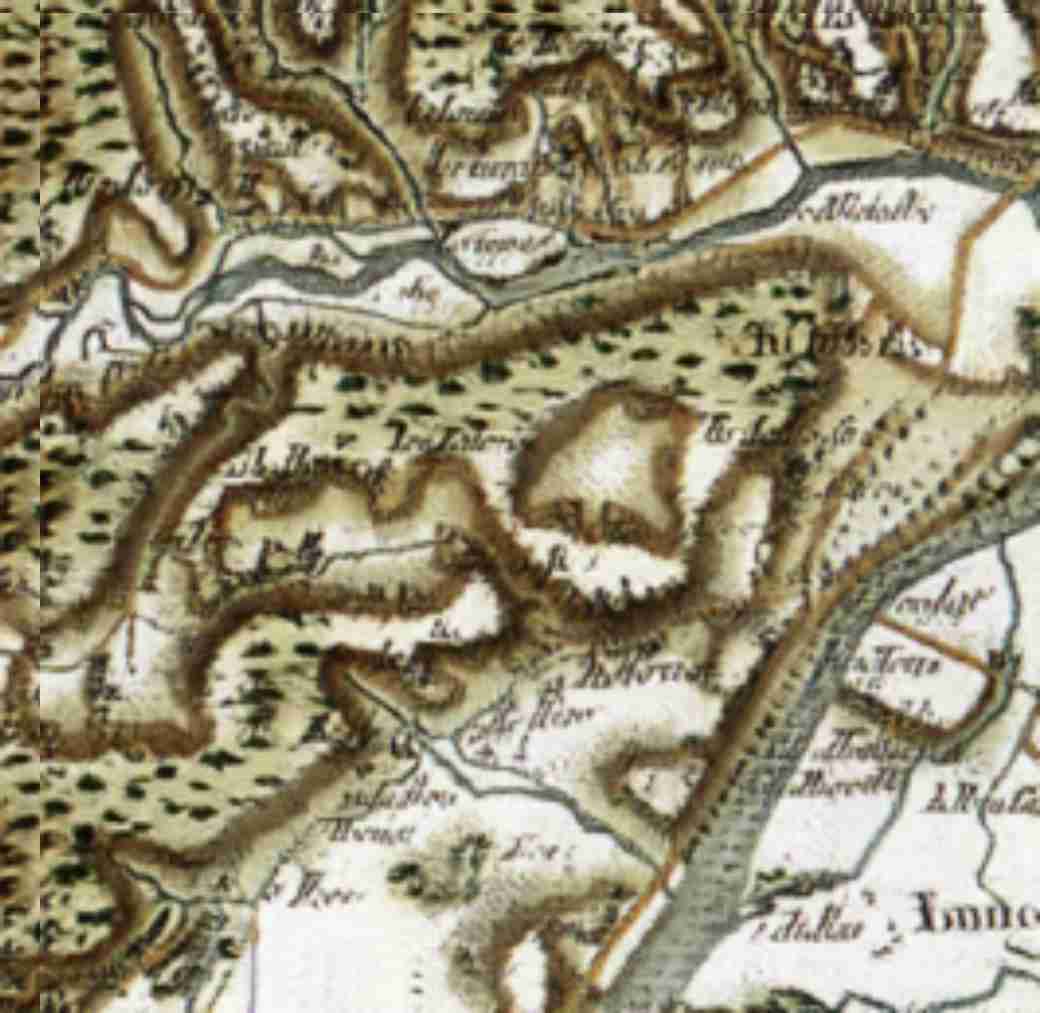} \\
    
    \raiseandrotate{.2}{\textbf{État-Major}} &
    \includegraphics[width=0.21\linewidth]{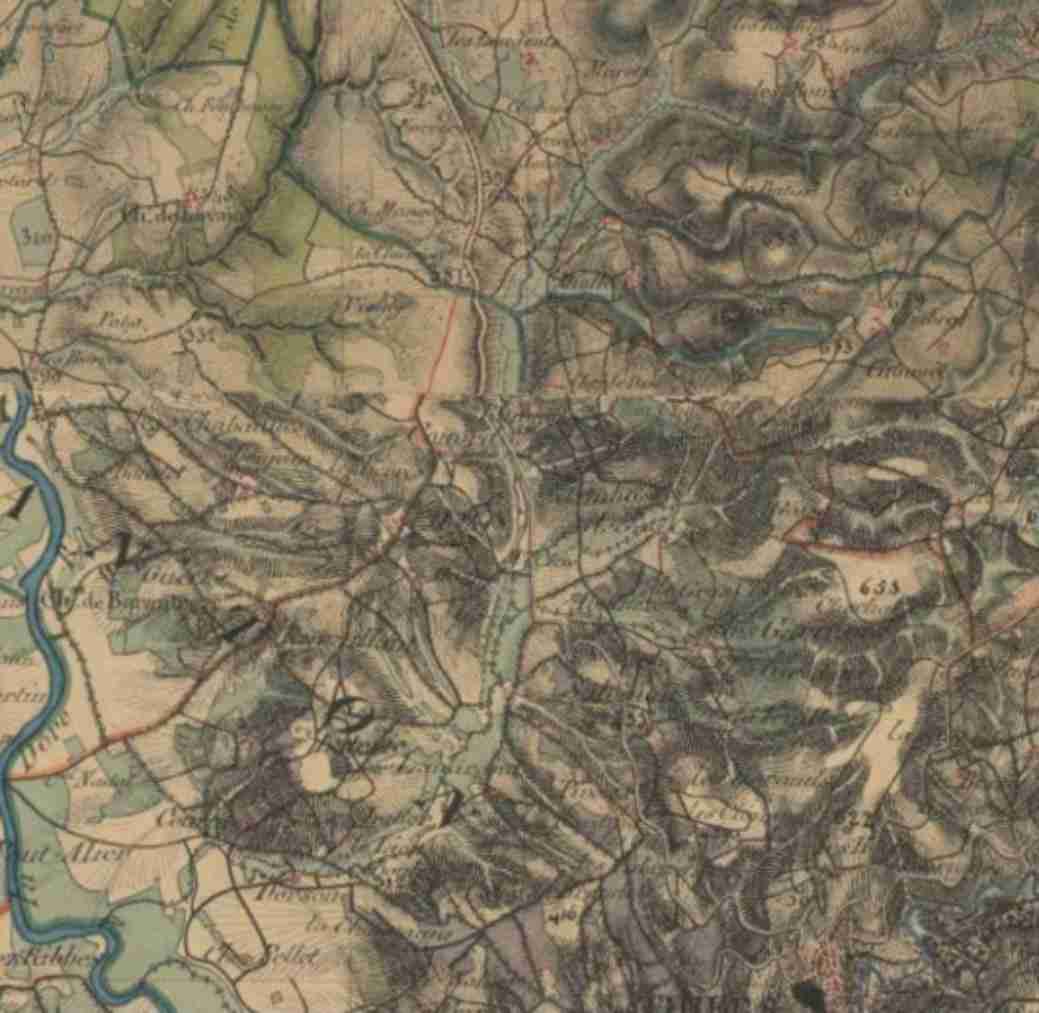} & 
    \includegraphics[width=0.21\linewidth]{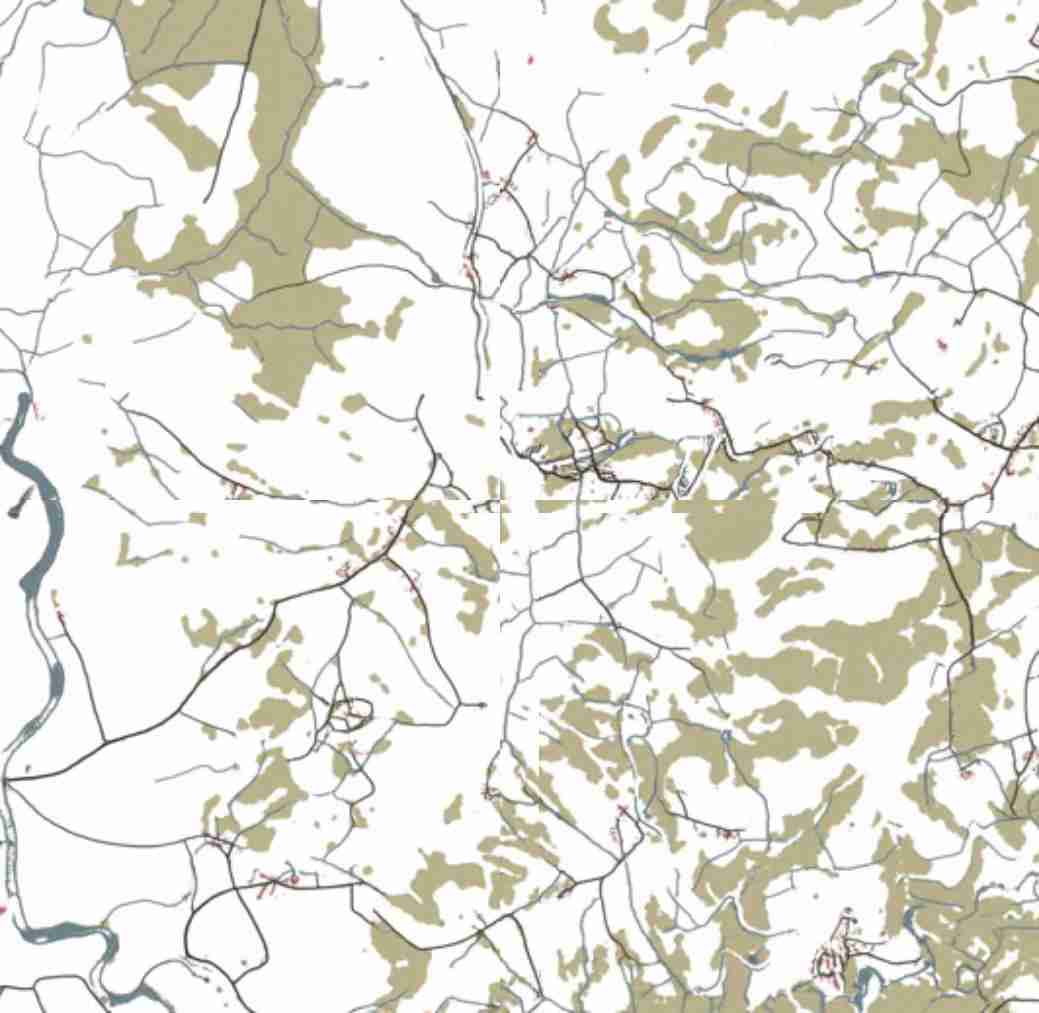} &
    \includegraphics[width=0.21\linewidth]{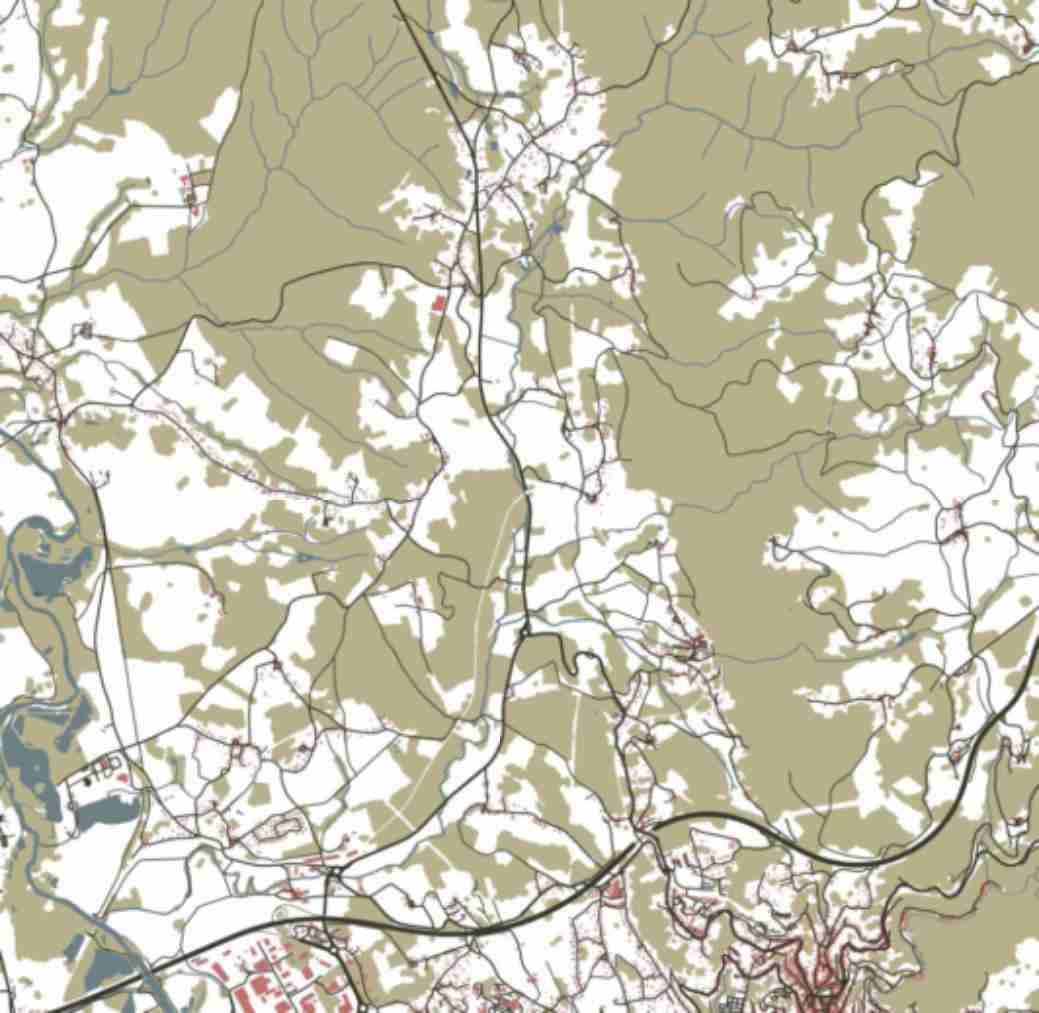} &
    \includegraphics[width=0.21\linewidth]{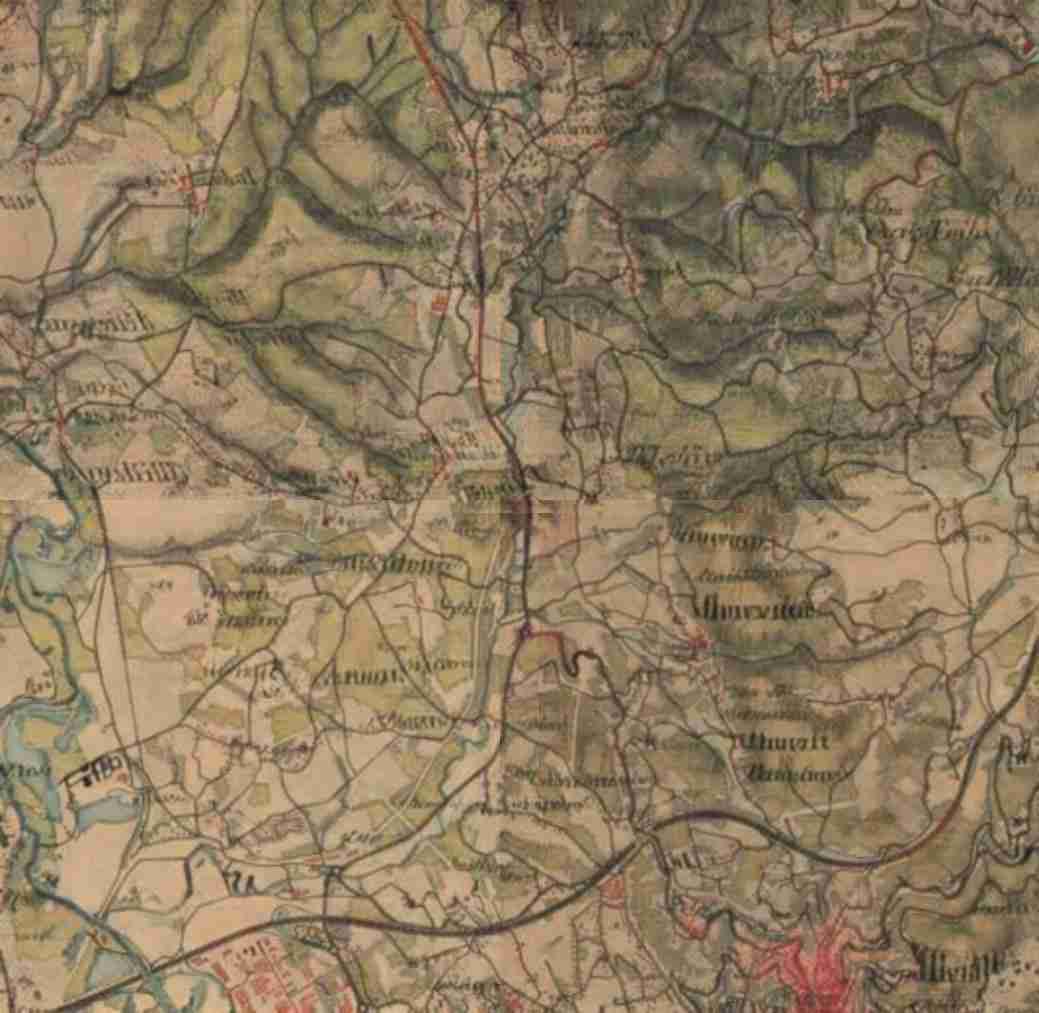} \\
    
    \raiseandrotate{.35}{\textbf{SCAN50}} &
    \includegraphics[width=0.21\linewidth]{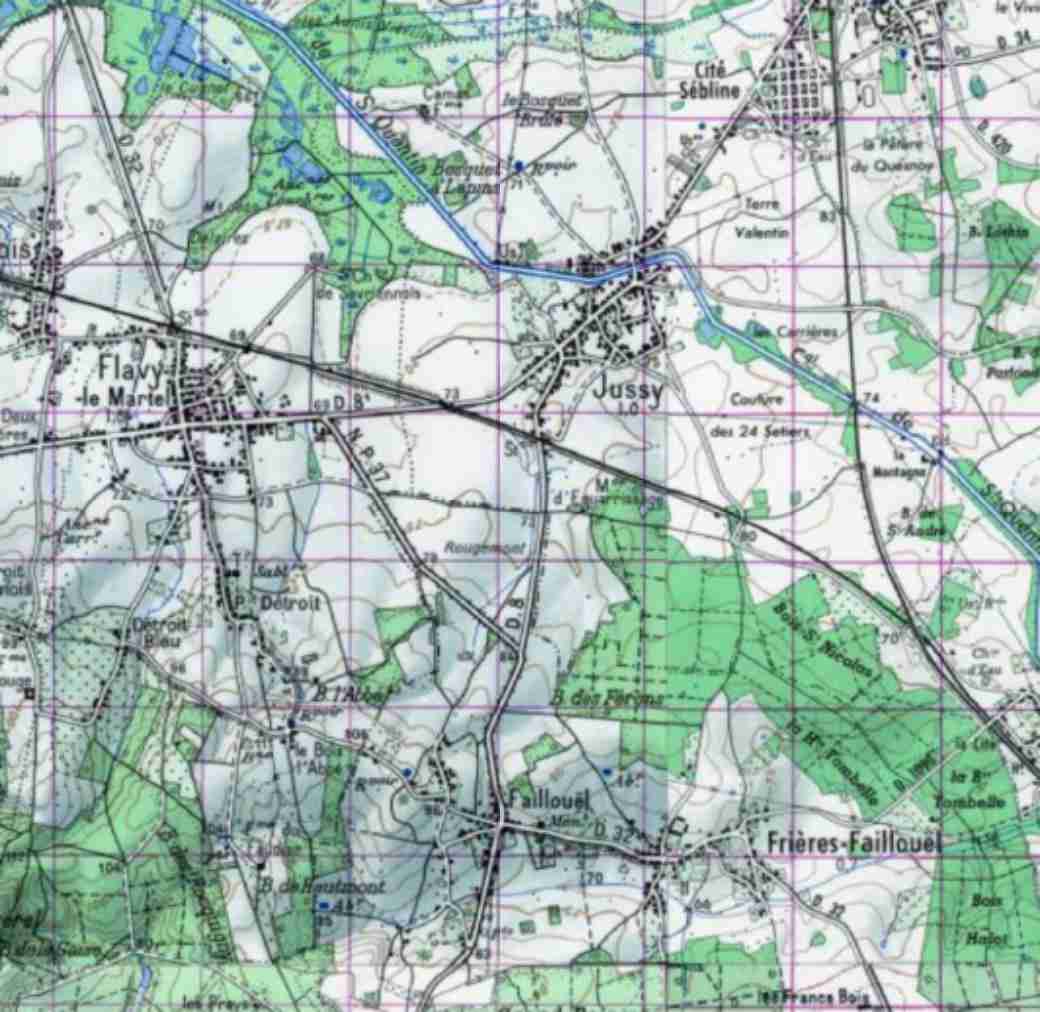} &
    \includegraphics[width=0.21\linewidth]{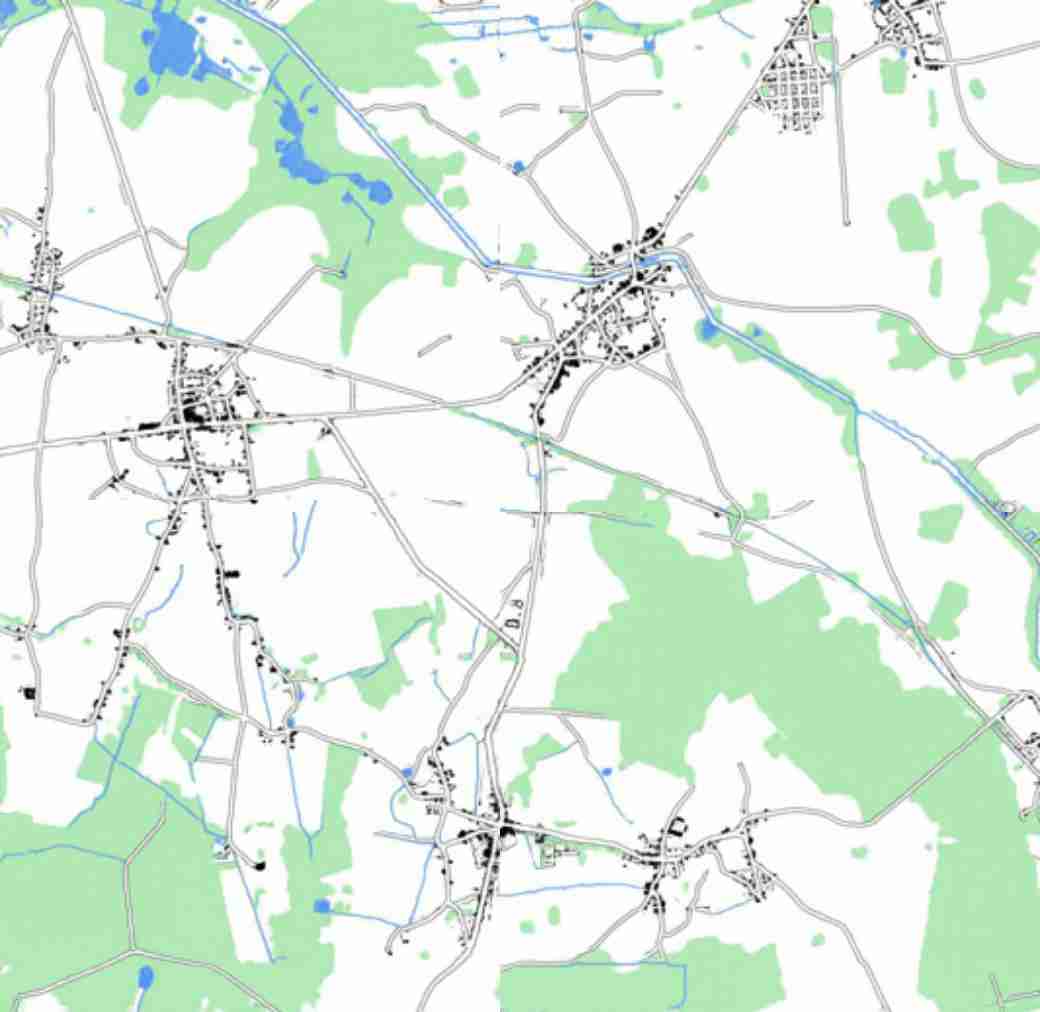} & 
    \includegraphics[width=0.21\linewidth]{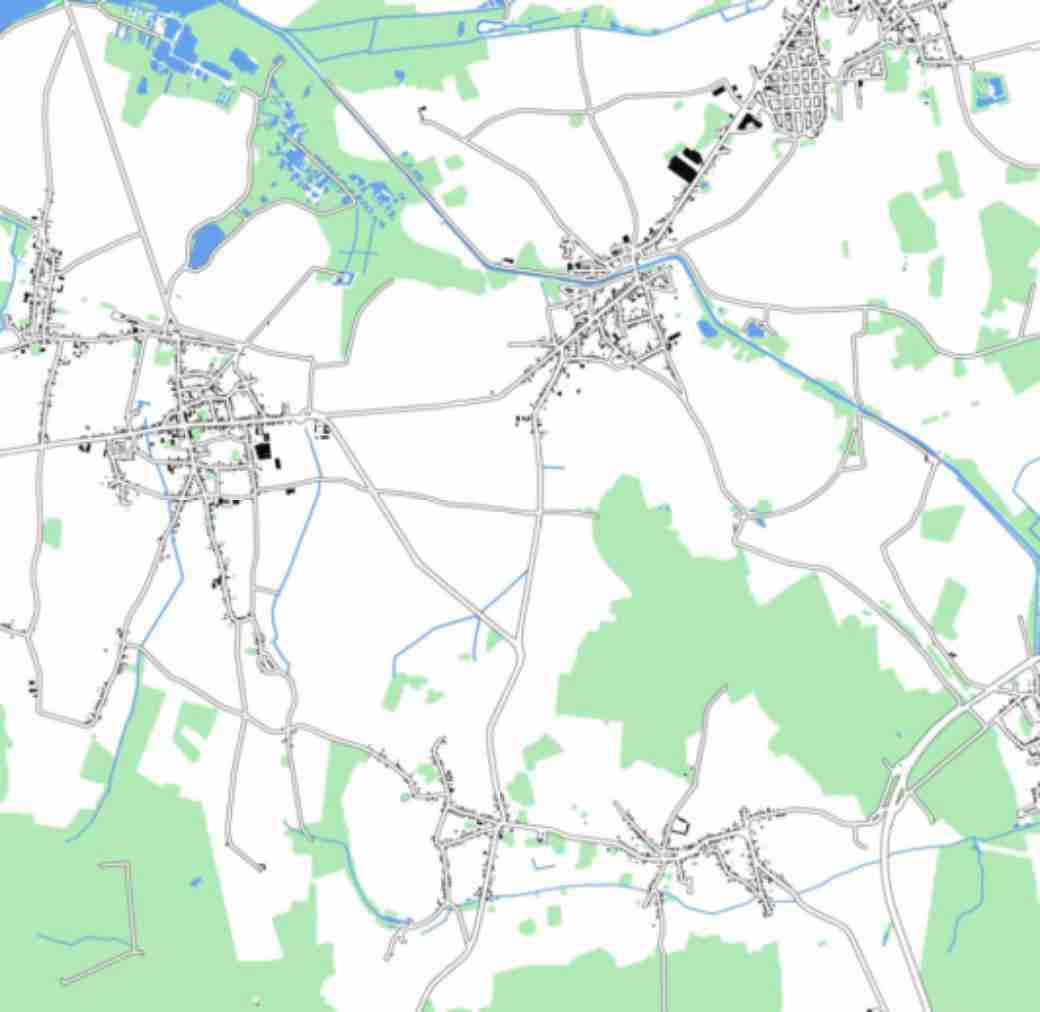} &
    \includegraphics[width=0.21\linewidth]{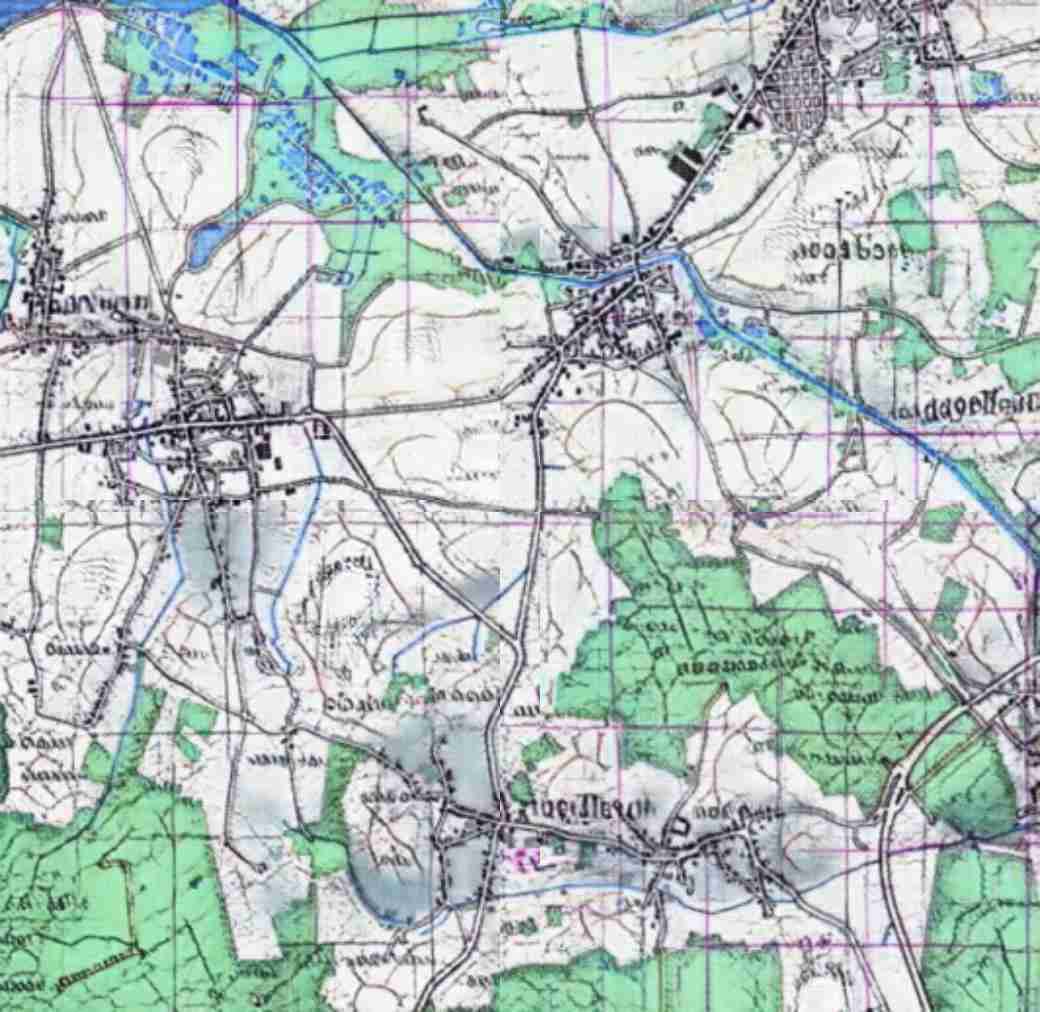} \\ \bottomrule
\end{tabular}

%% file: ablation_seg_quantitative.tex
\begin{tabular}{l@{\;\;}c@{\;\;}ccc@{\;\;}c@{\;\;}c@{\;\;}c@{\;\;}c@{\;\;}c}
    \toprule
    \multirow{2}{*}{\textbf{Map}} &
    \multicolumn{2}{c}{\textbf{Version}} & \; &
    \multirow{2}{*}{\textbf{OA}}& \textbf{Mean} &
    \multicolumn{4}{c}{\textbf{Per-class dIoU}} \\ \cline{2-3} \cline{7-10}
    & Match
    & $\mathcal{L}_{\text{tran}}$
    & & & \textbf{dIoU}
    & \raisebox{-0.7mm}{\tree}
    & \raisebox{-0.7mm}{\building}
    & \raisebox{-0.7mm}{\water}
    & \raisebox{-0.7mm}{\road} \\ \midrule
    
    \multirow{4}{*}{\textbf{Cassini}}
    & \no & \no & & 61.7 & 17.6 & 22.2 & 4.7 & 29.8 & 13.6 \\
    & \yes & \no & & 67.9 & 20.2 & 31.4 & 3.4 & 32.4 & \textbf{13.6} \\
    & \no & \yes & & 83.8 & 31.8 & 46.4 & \textbf{10.8} & 57.6 & 12.5 \\
    & \yes & \yes & & \textbf{85.3} & \textbf{36.1} & \textbf{56.1} & 4.7 & \textbf{70.9} & 12.7 \\ \midrule
    
    \multirow{4}{*}{\textbf{État-Major}} 
    & \no & \no & & 51.6 & 10.6 & 22.2 & 8.8 & 9.8 & 1.6 \\
    & \yes & \no & & 62.2 & 11.6 & 23.1 & 12.8 & 5.7 & 4.9 \\
    & \no & \yes & & \textbf{79.6} & \textbf{30.4} & \textbf{51.7} & \textbf{22.5} & 44.8 & 2.5 \\
    & \yes & \yes & & 78.4 & 28.7 & 43.0 & 15.6 & \textbf{51.0} & \textbf{5.3} \\ \bottomrule
\end{tabular}

%% file: ablation_seg_qualitative.tex
\begin{tabular}{ccccccc}
    \toprule
    & \multirow{2}{*}{\textbf{Input}}
    & Match\;\no
    & Match\;\yes
    & Match\;\no
    & Match\;\yes
    & \textbf{Ground} \\ 
    & & $\mathcal{L}_{\text{tran}}$\;\no
    & $\mathcal{L}_{\text{tran}}$\;\no
    & $\mathcal{L}_{\text{tran}}$\;\yes
    & $\mathcal{L}_{\text{tran}}$\;\yes
    & \textbf{Truth} \\ \midrule
    \raiseandrotate{.25}{\small \textbf{Cassini}} &
    \includegraphics[width=0.15\linewidth]{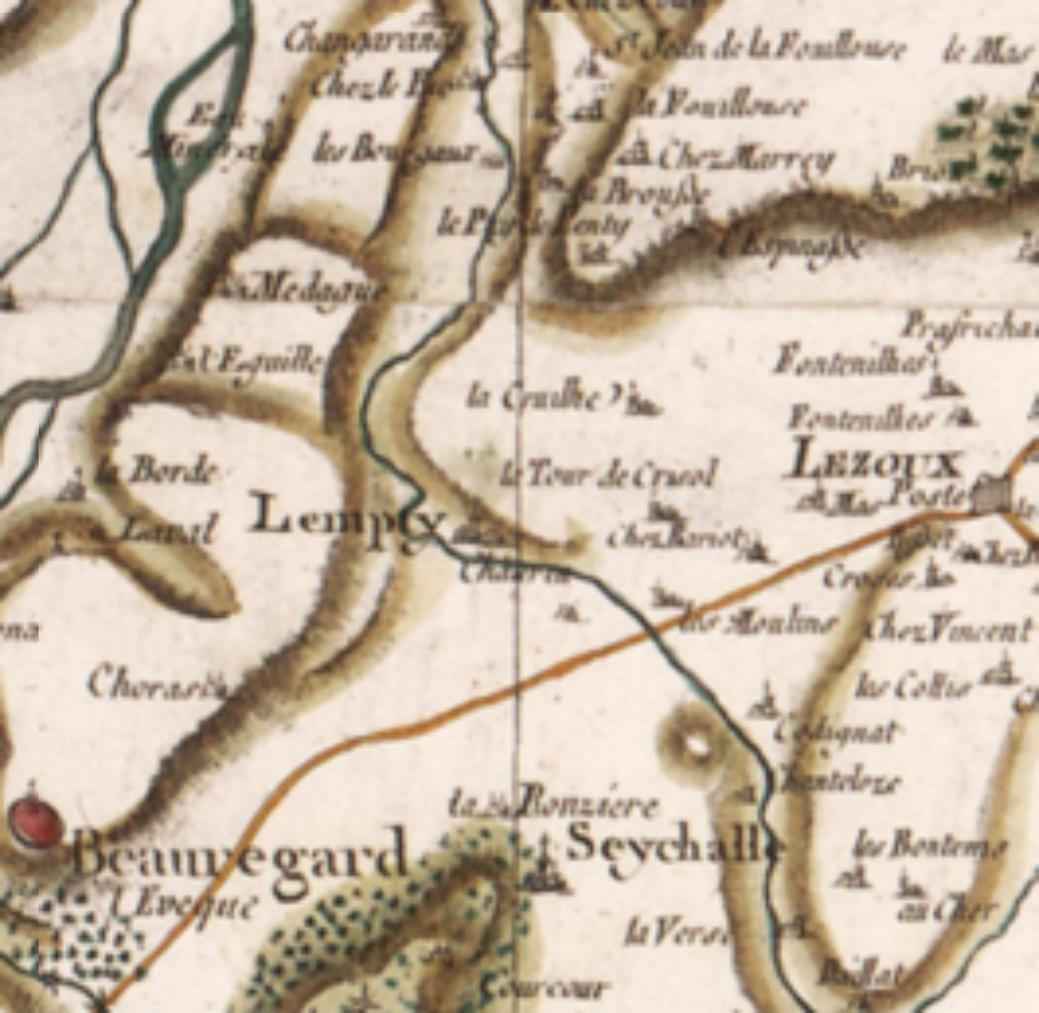} &
    \fbox{\includegraphics[width=0.15\linewidth]{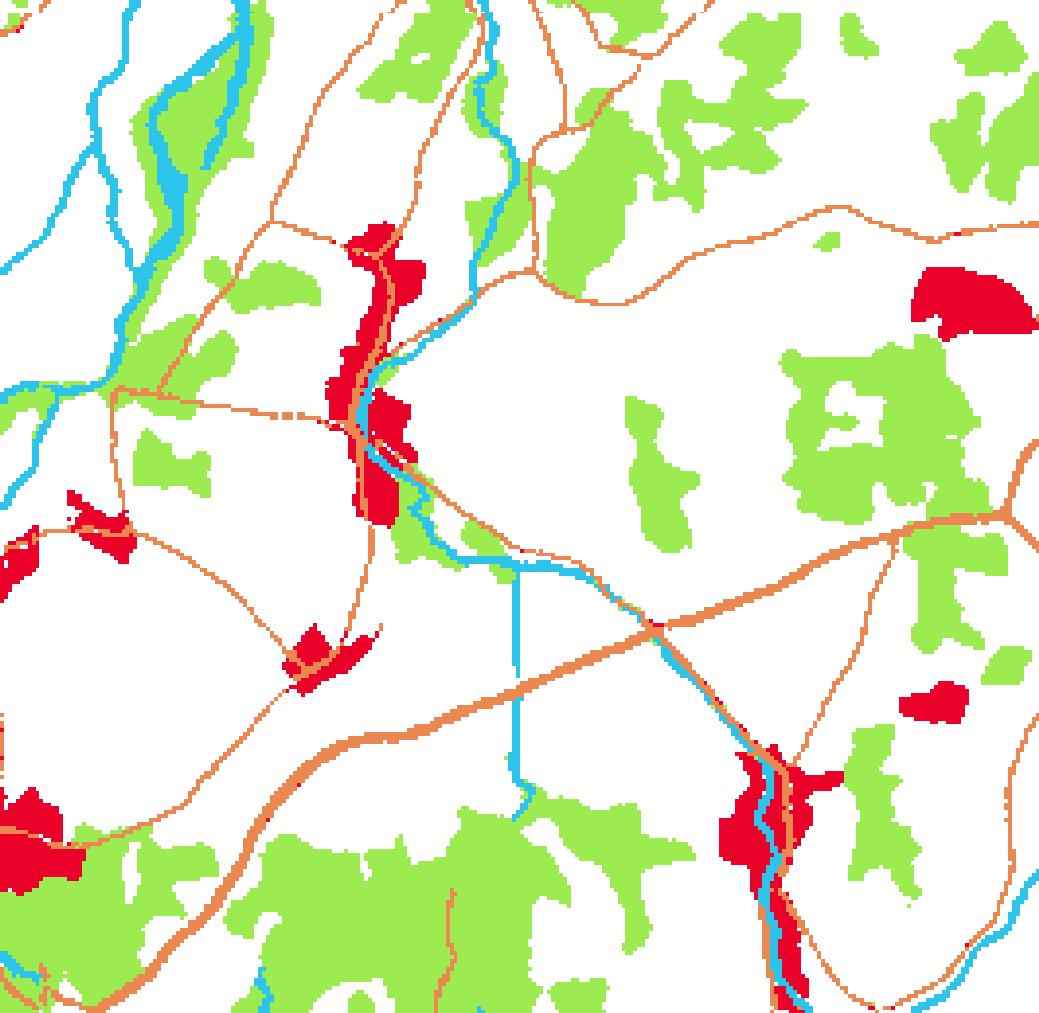}} &
    \fbox{\includegraphics[width=0.15\linewidth]{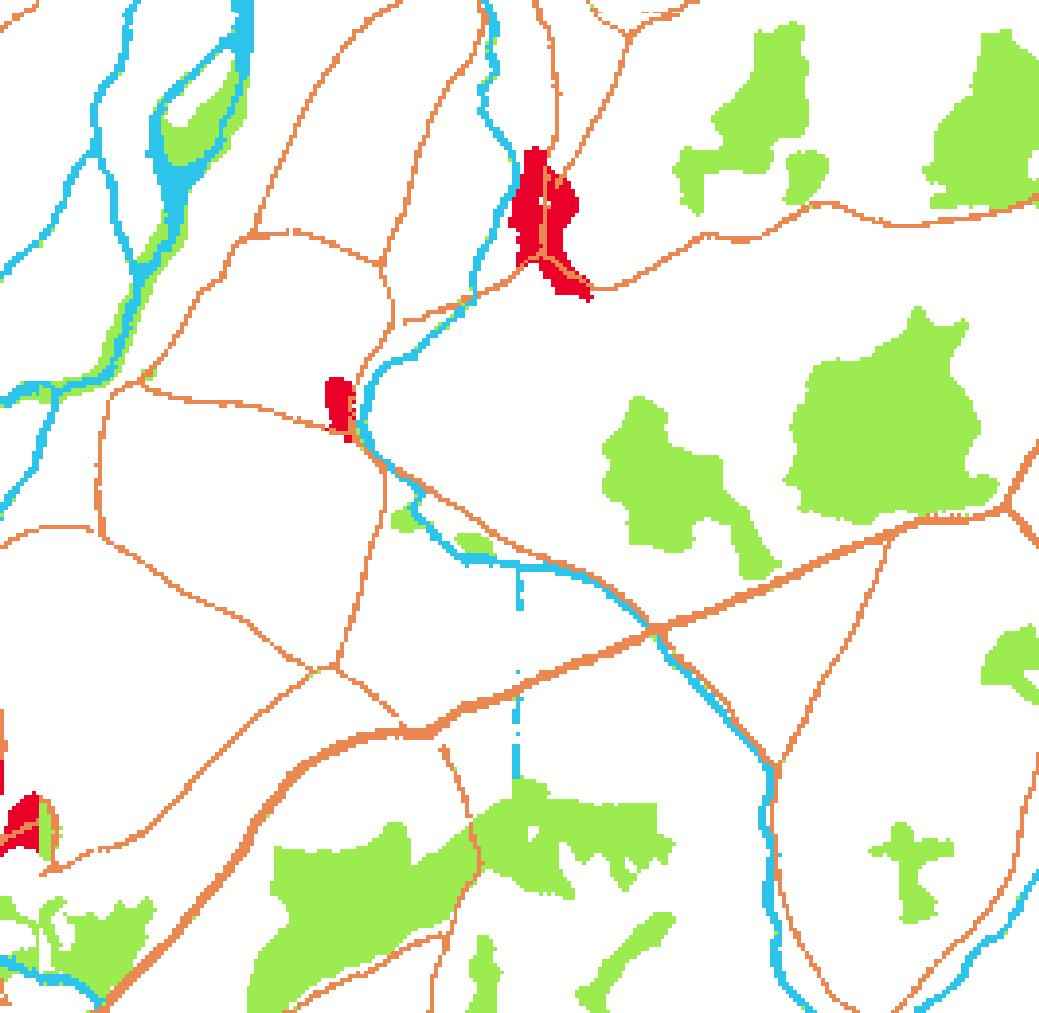}} &
    \fbox{\includegraphics[width=0.15\linewidth]{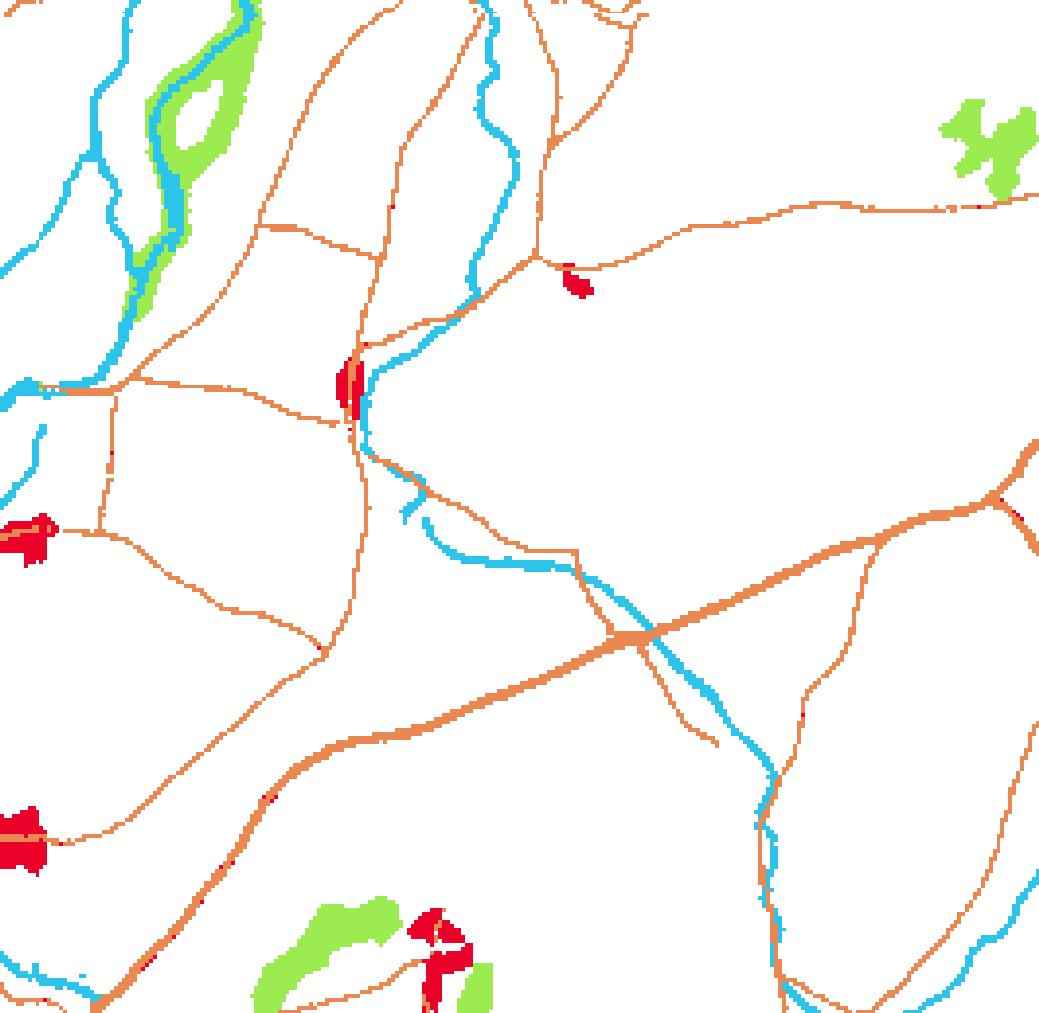}} &
    \fbox{\includegraphics[width=0.15\linewidth]{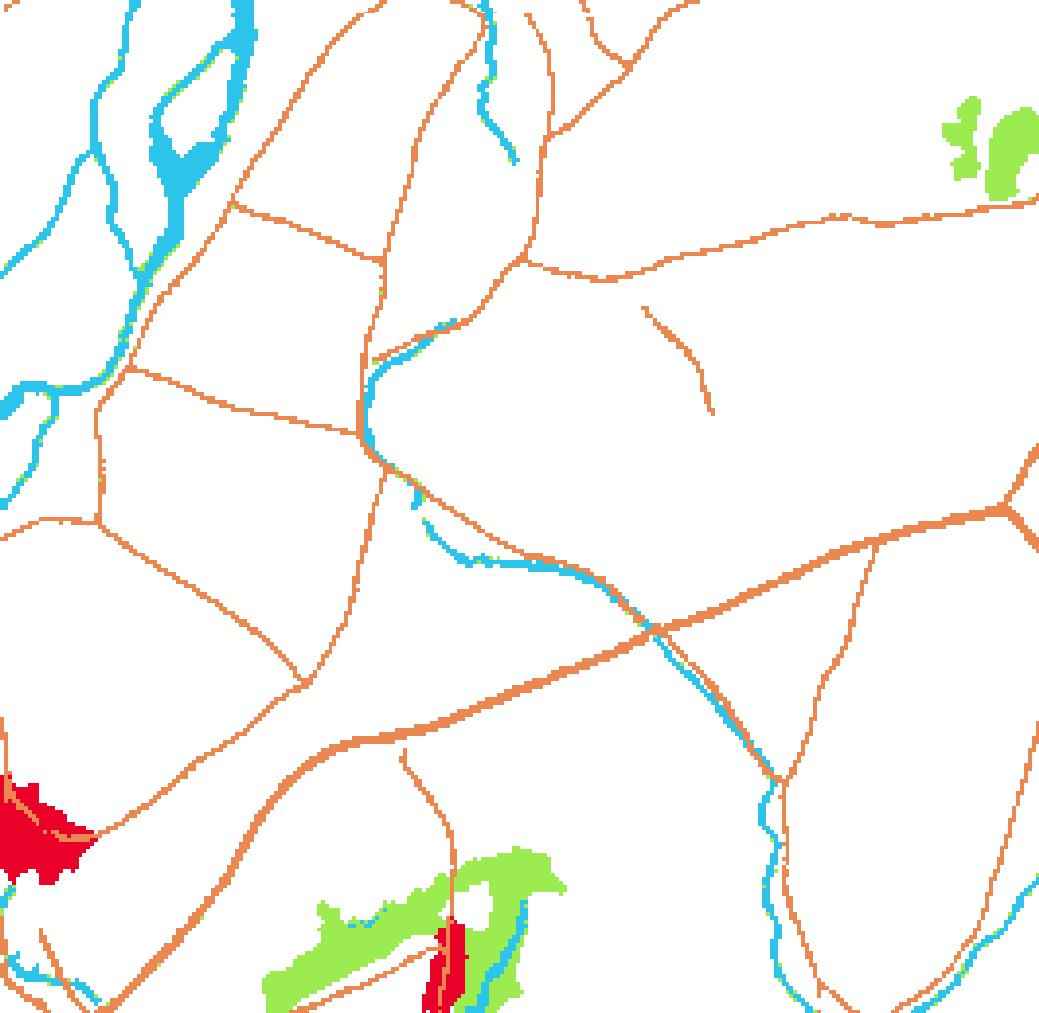}} & 
    \fbox{\includegraphics[width=0.15\linewidth]{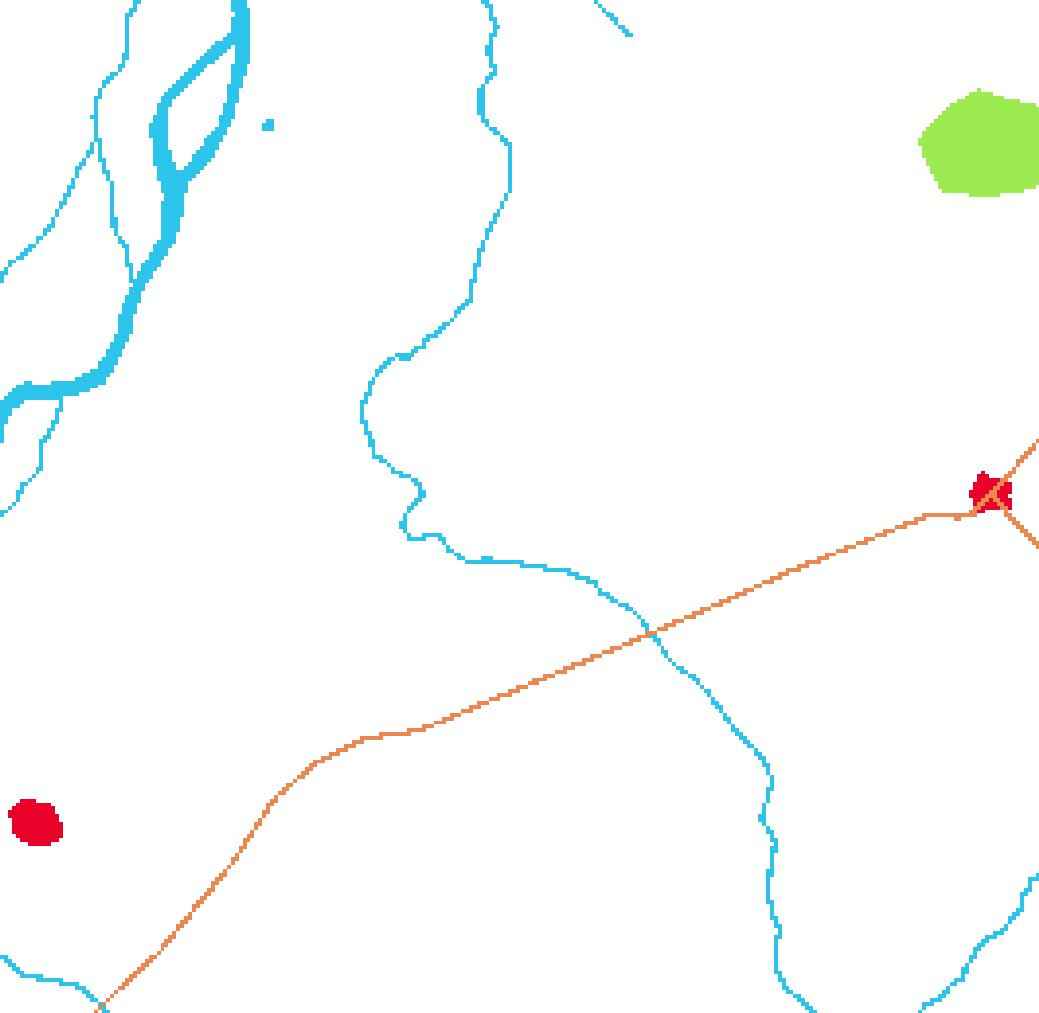}} \\
    
    \raiseandrotate{-.025}{\small \textbf{État-Major}} &
    \includegraphics[width=0.15\linewidth]{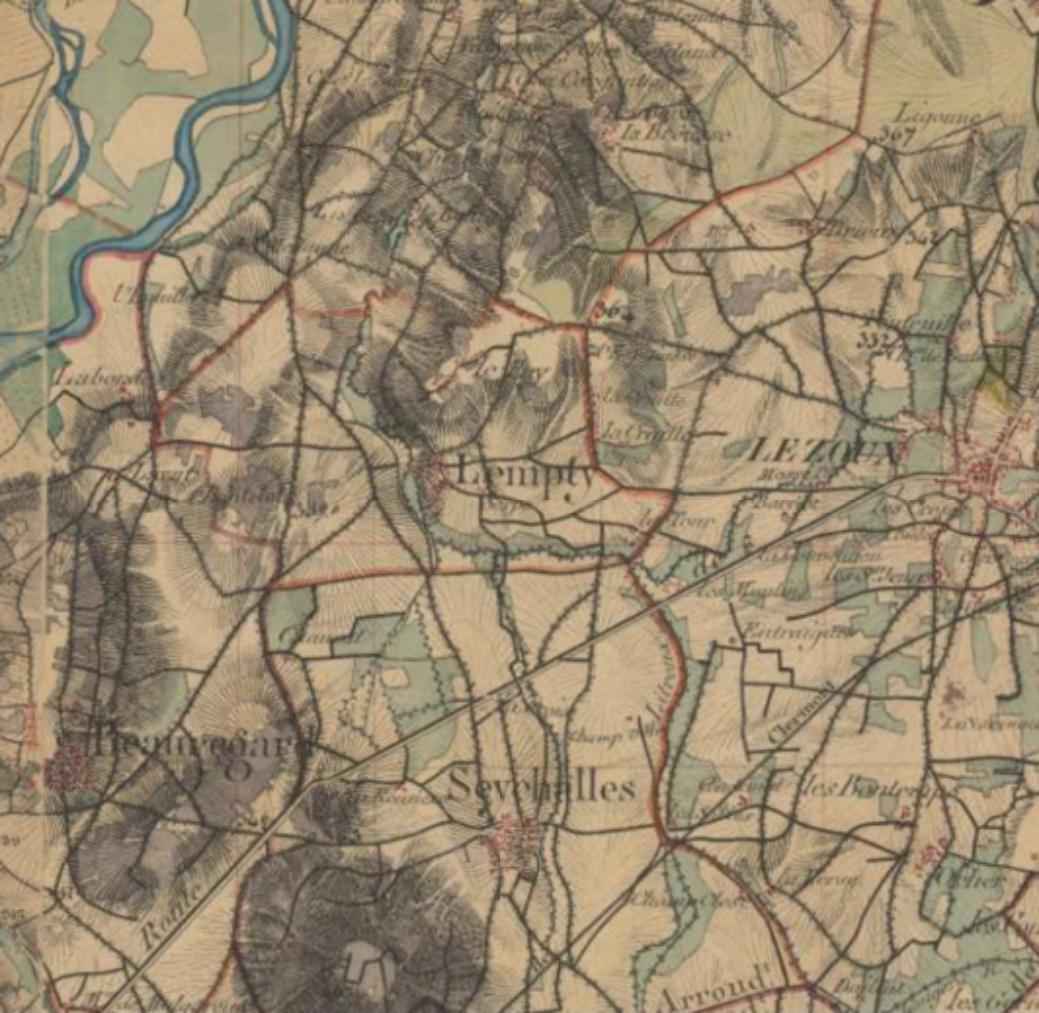} &
    \fbox{\includegraphics[width=0.15\linewidth]{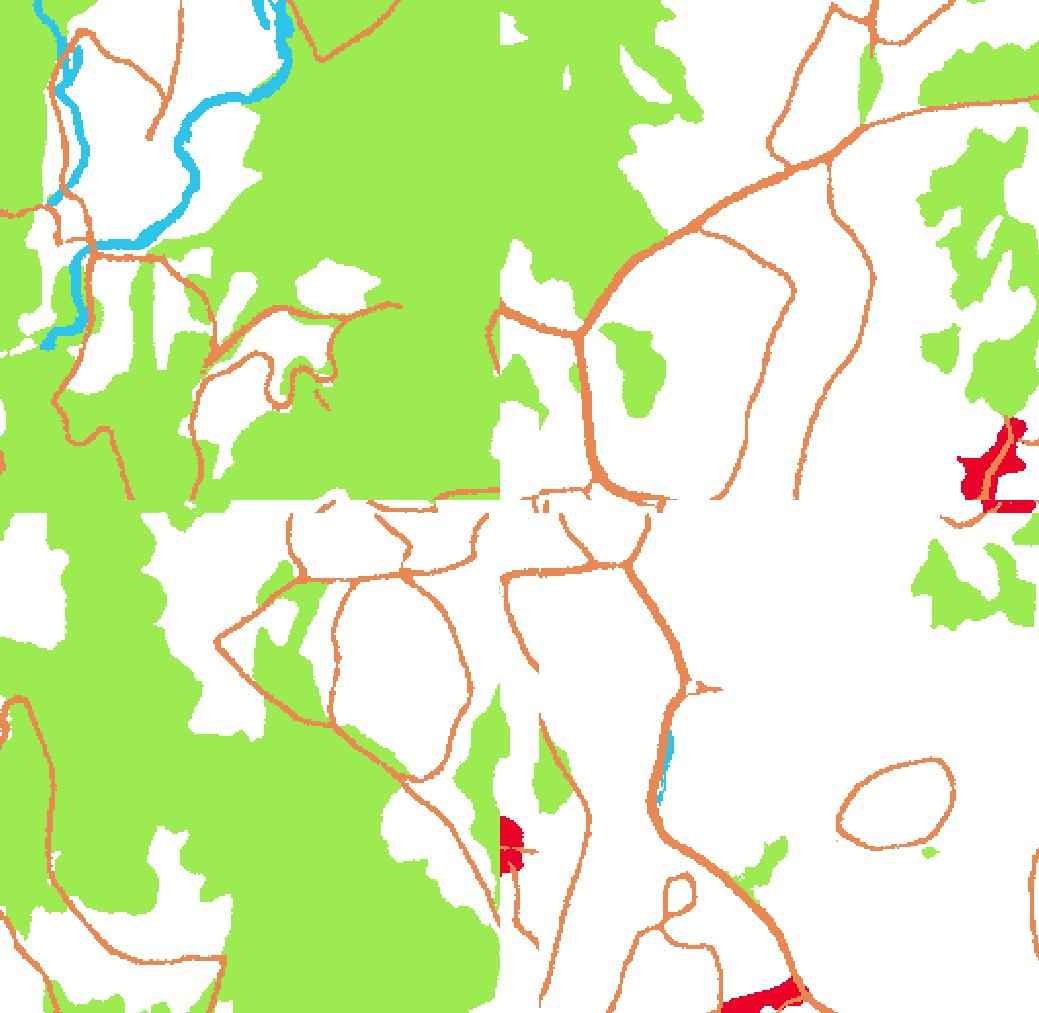}} &
    \fbox{\includegraphics[width=0.15\linewidth]{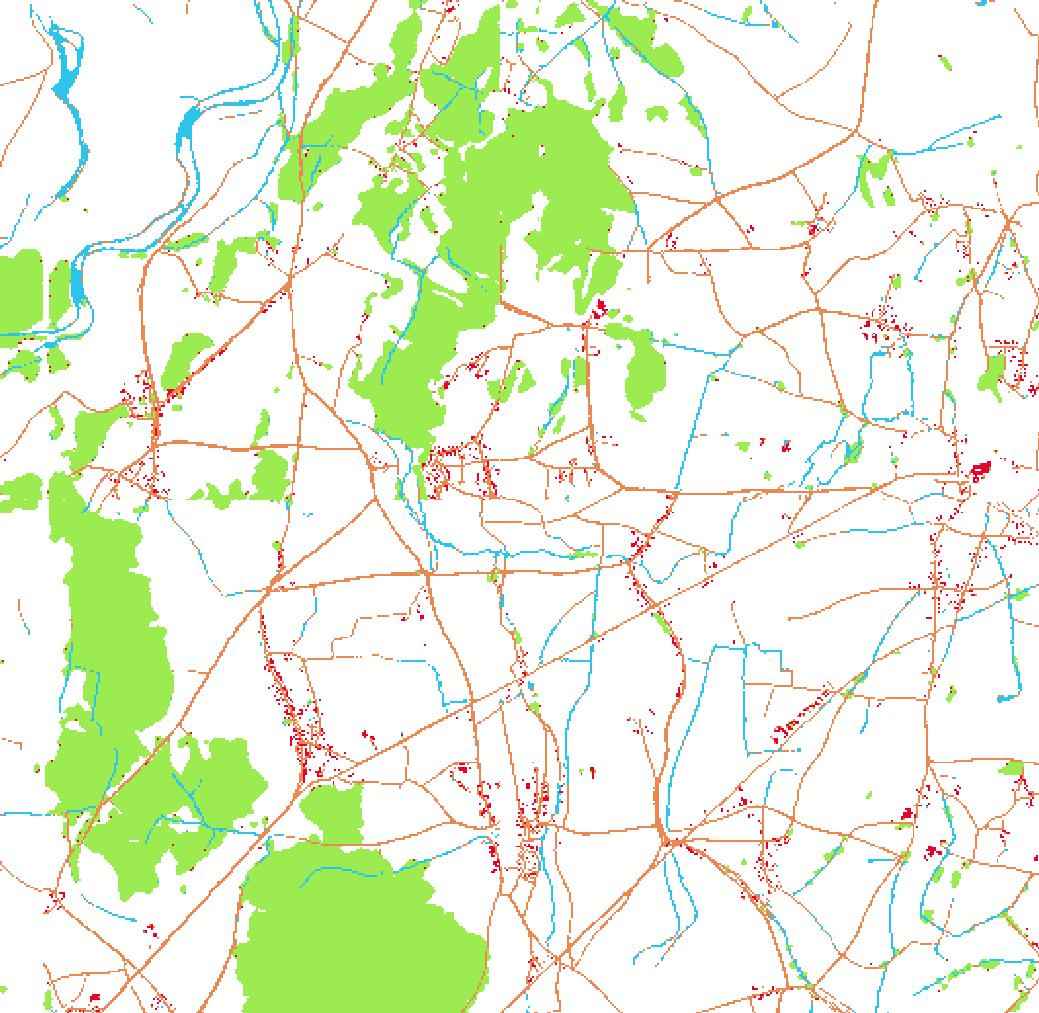}} &
    \fbox{\includegraphics[width=0.15\linewidth]{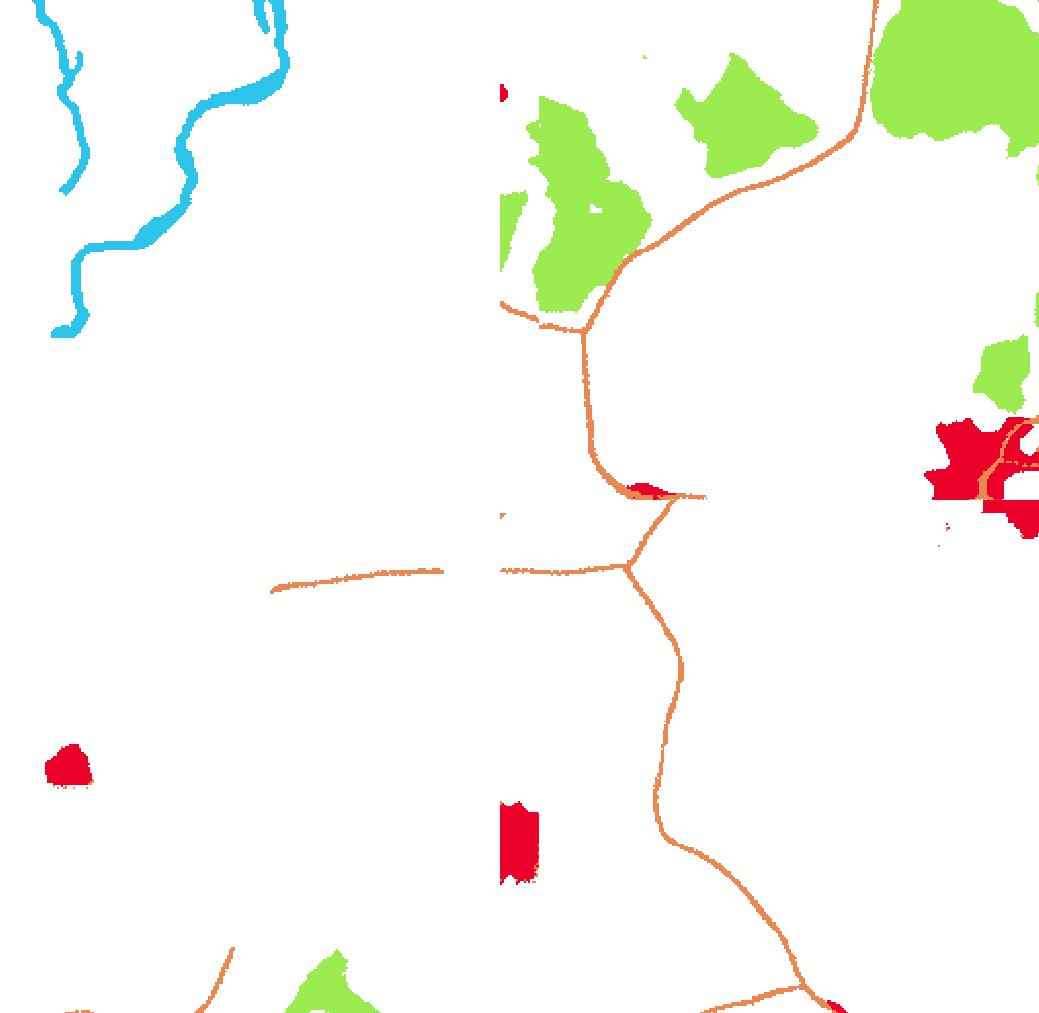}} &
    \fbox{\includegraphics[width=0.15\linewidth]{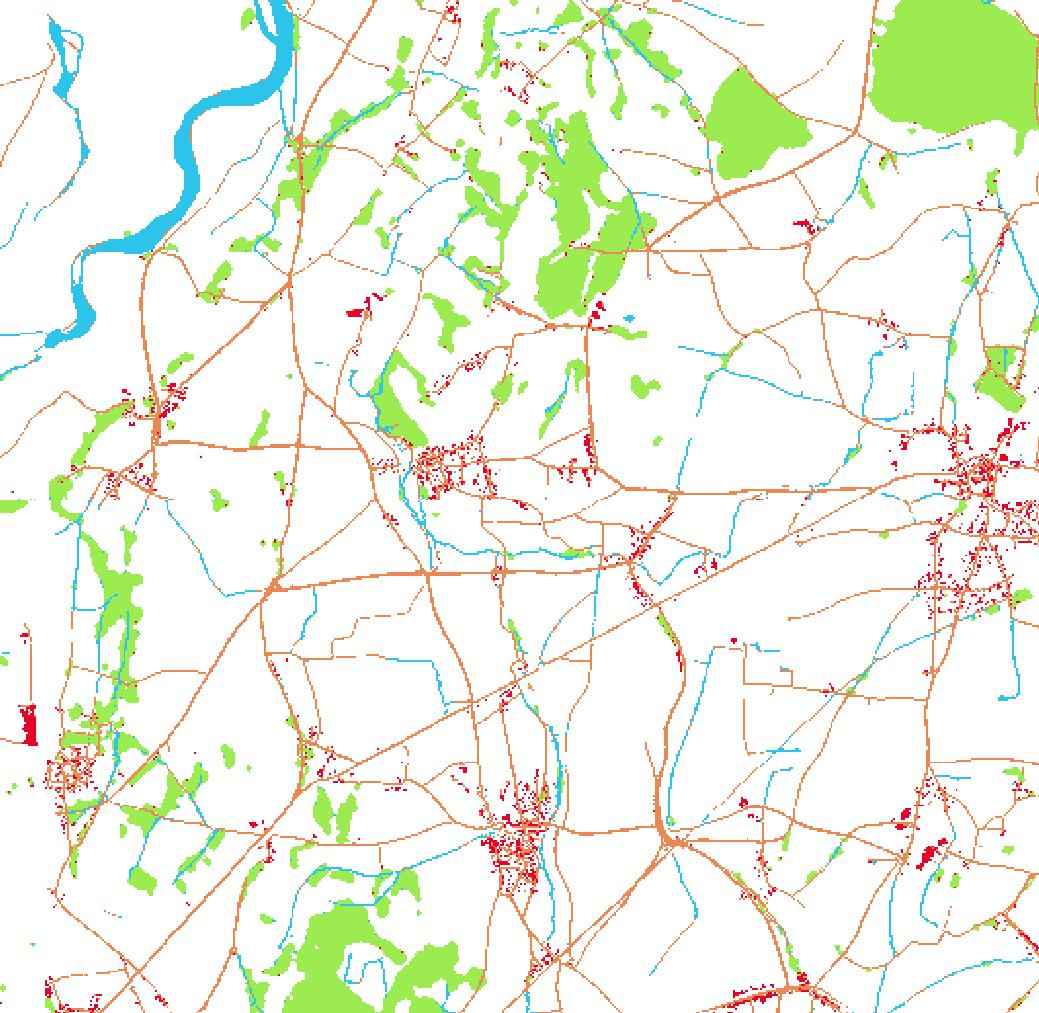}} &
    \fbox{\includegraphics[width=0.15\linewidth]{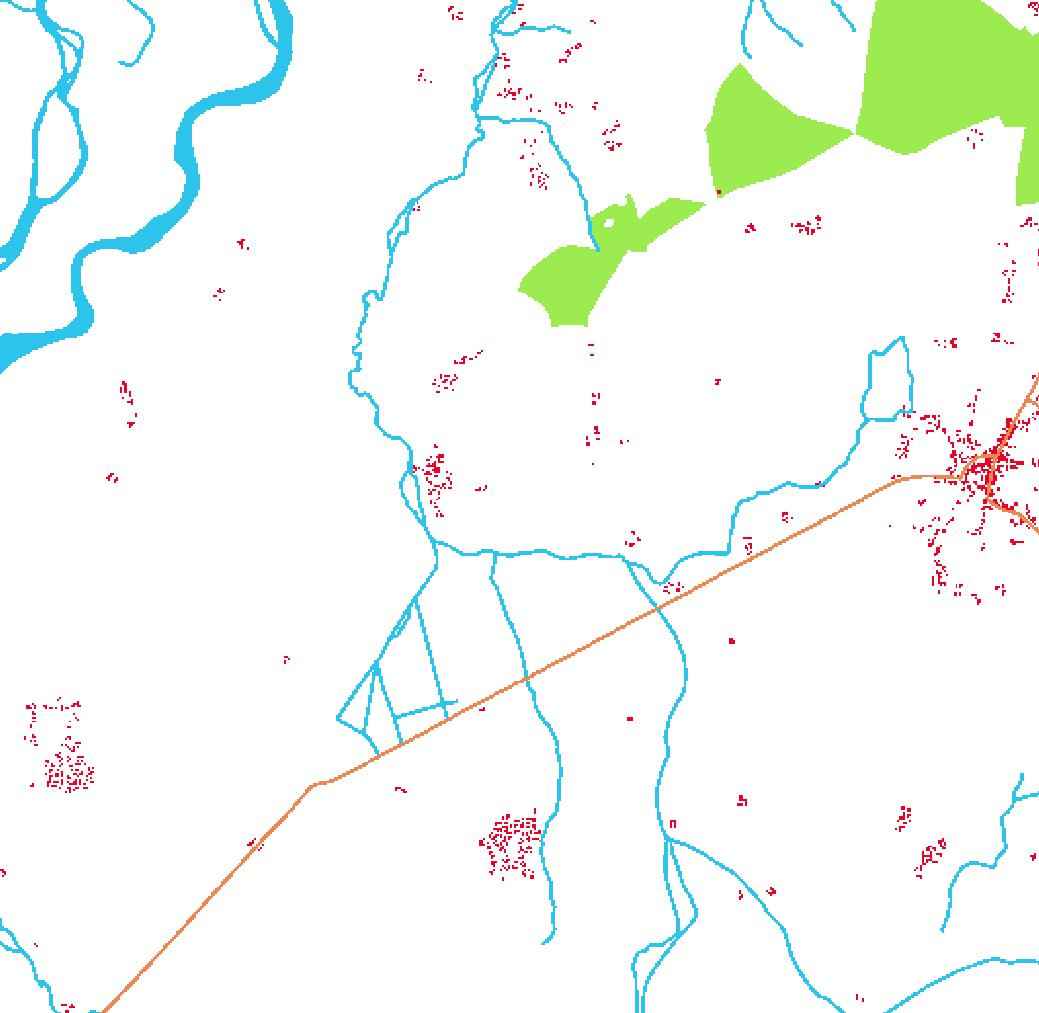}} \\
    
    \raiseandrotate{.125}{\small \textbf{SCAN50}} &
    \includegraphics[width=0.15\linewidth]{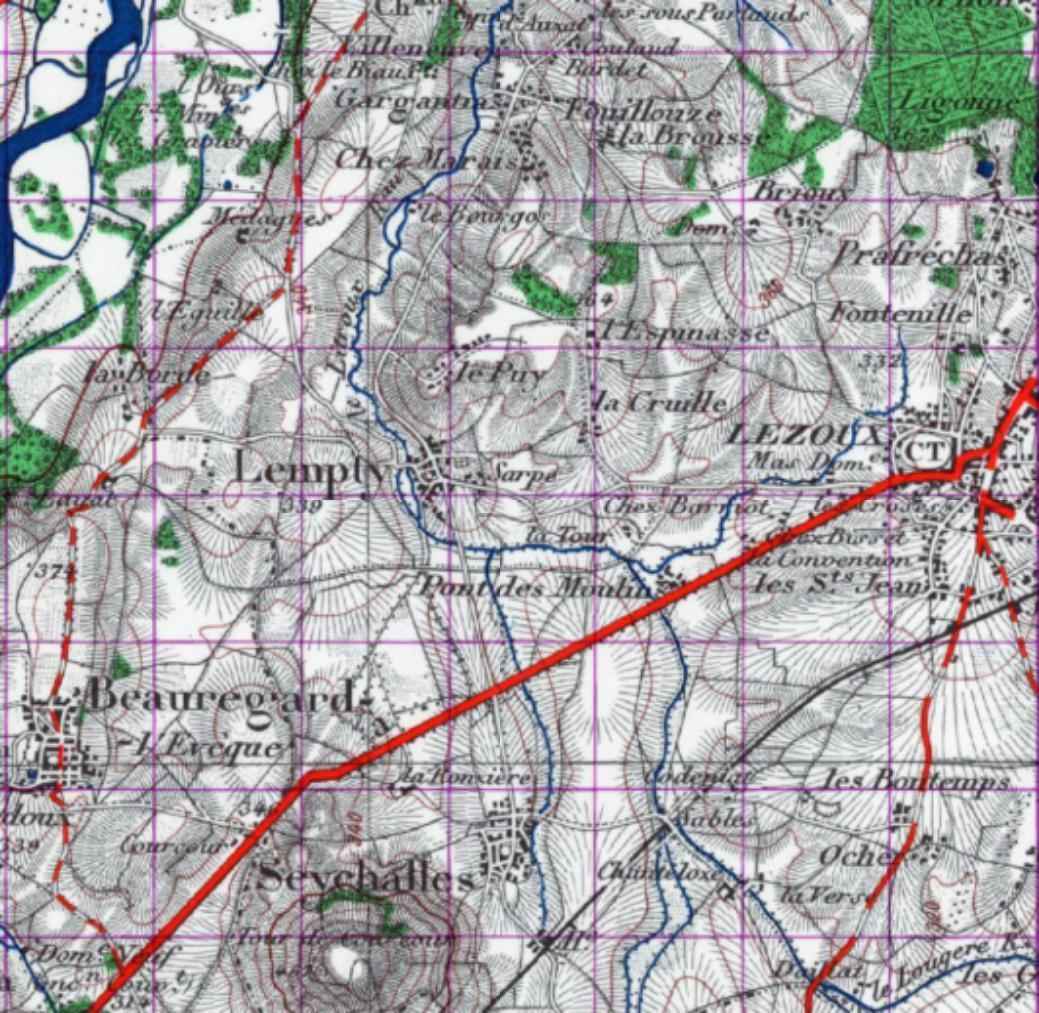} &
    \fbox{\includegraphics[width=0.15\linewidth]{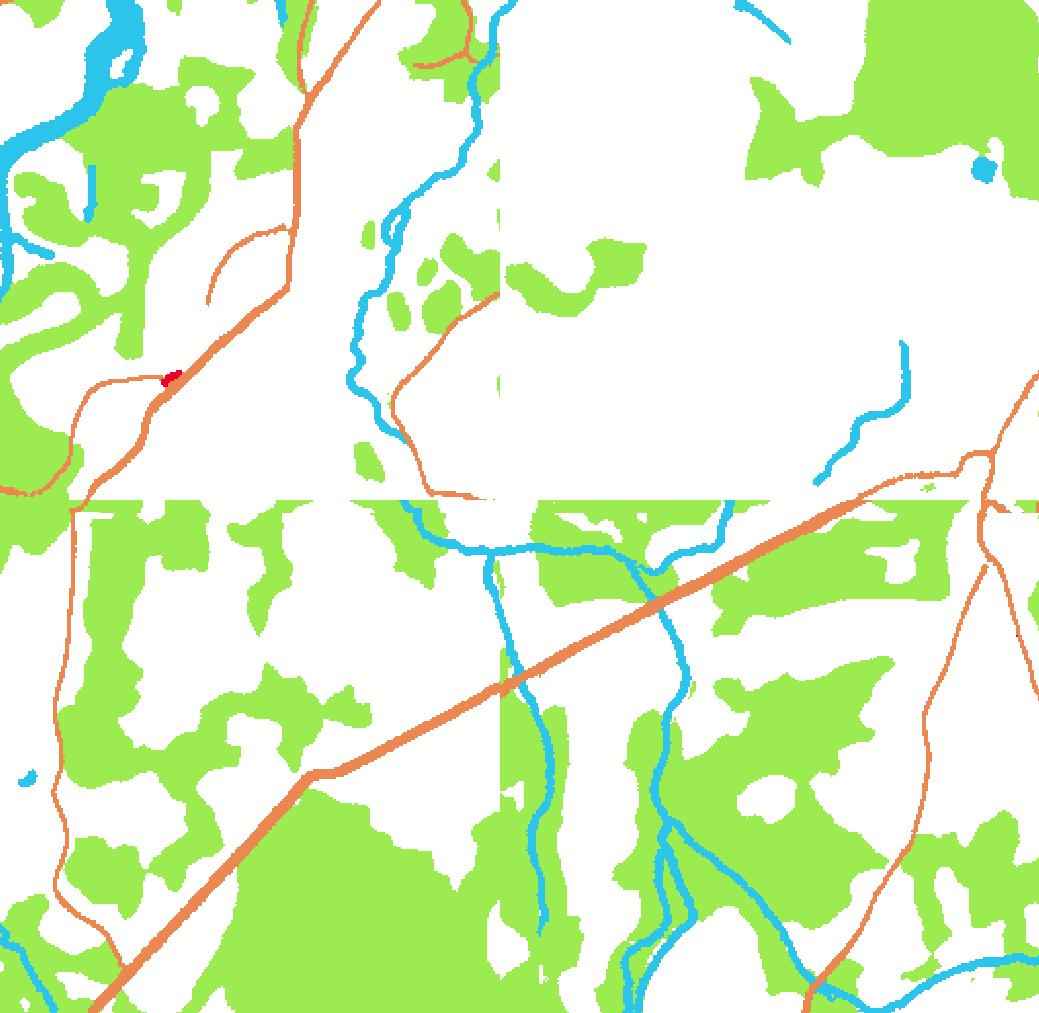}} &
    \fbox{\includegraphics[width=0.15\linewidth]{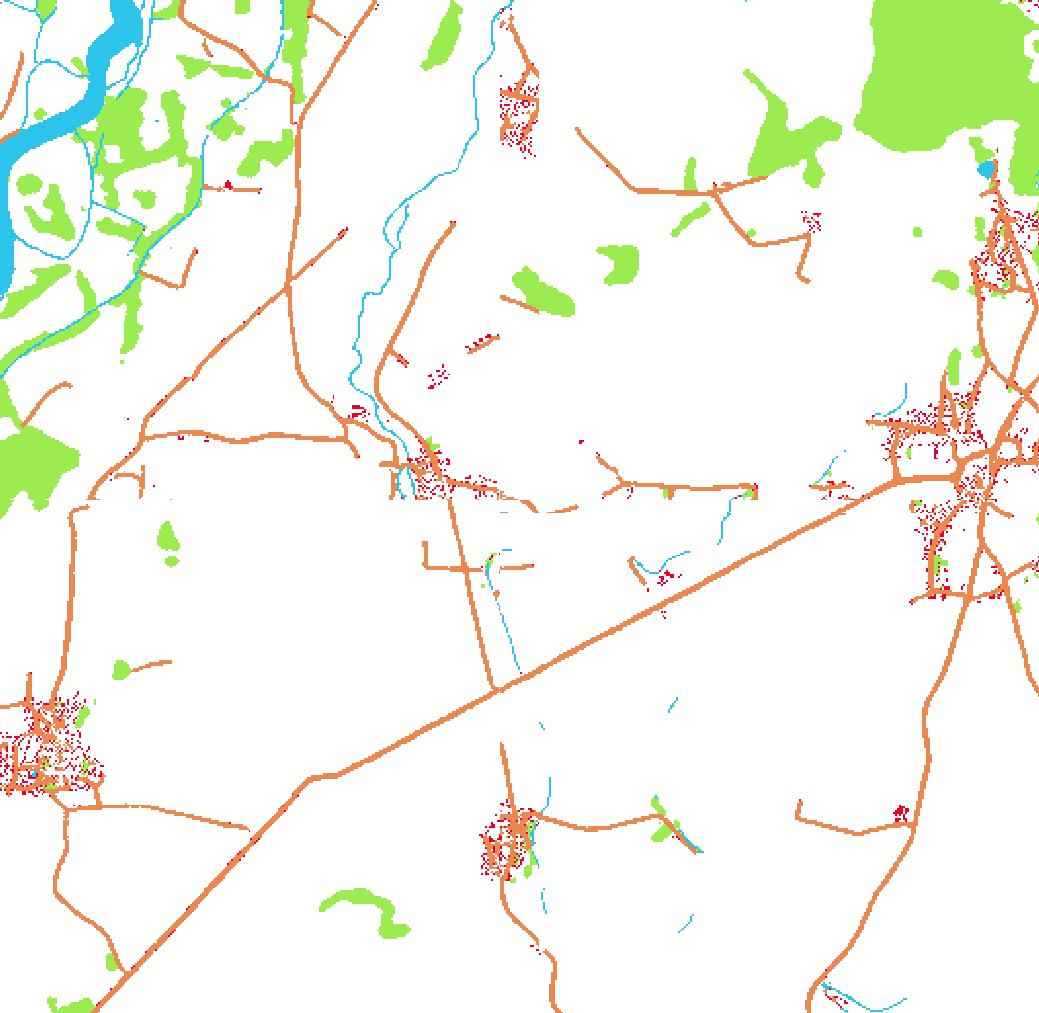}} &
    \fbox{\includegraphics[width=0.15\linewidth]{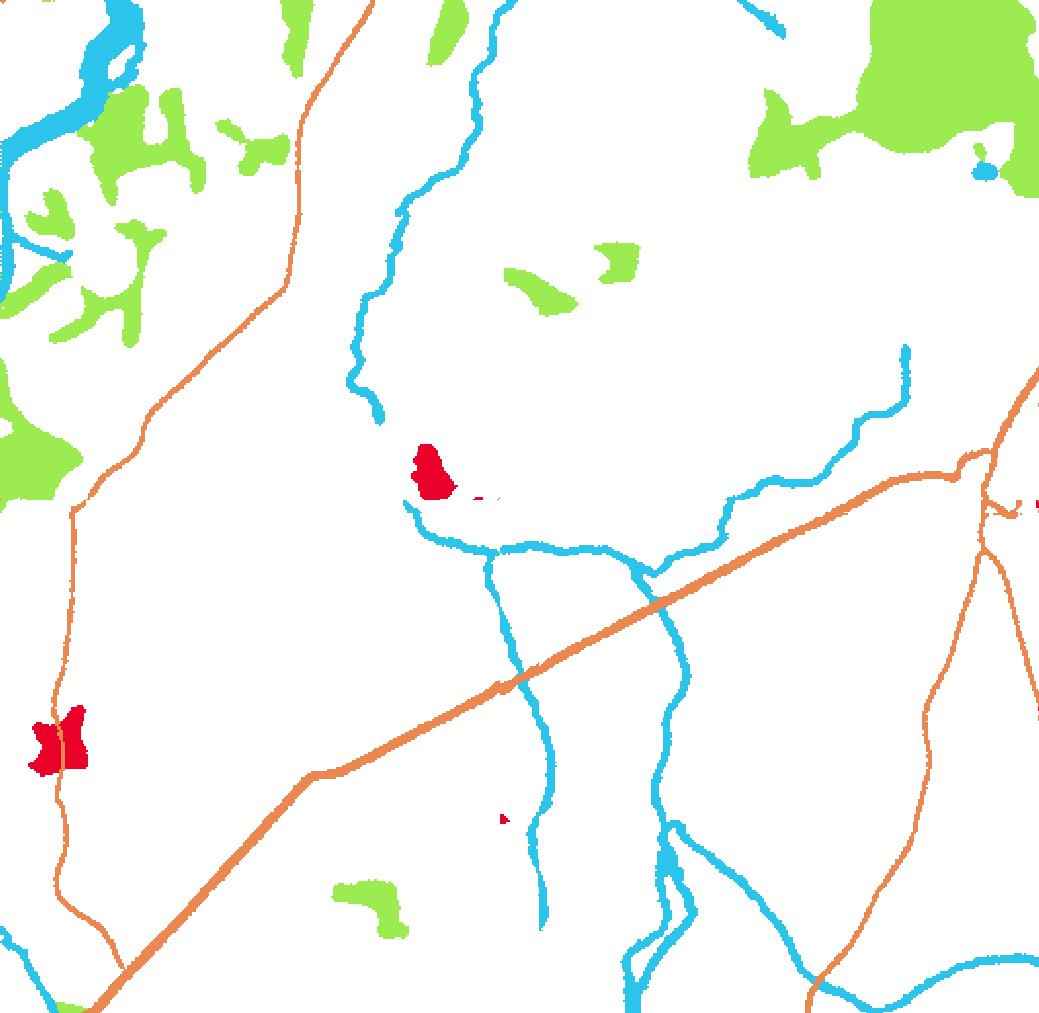}} &
    \fbox{\includegraphics[width=0.15\linewidth]{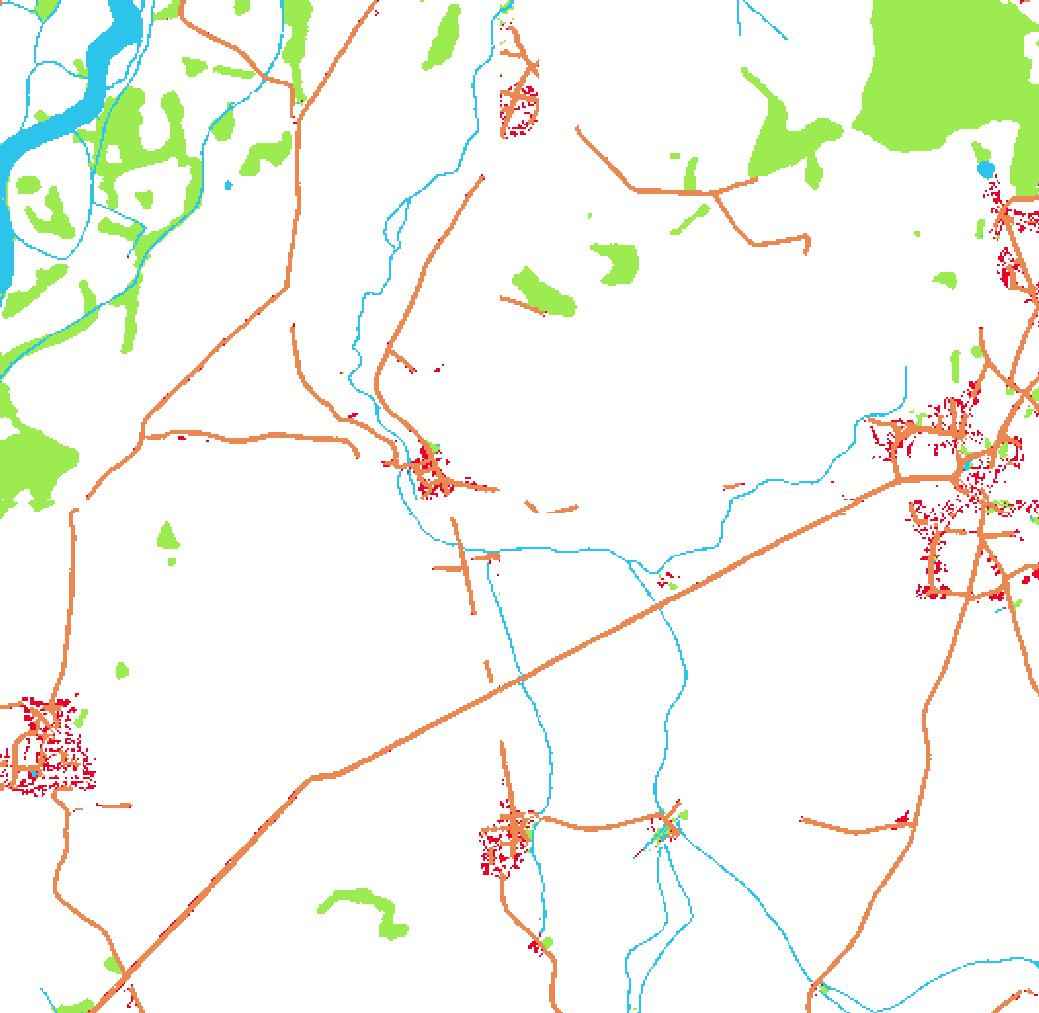}} &
    \fbox{\begin{tikzpicture}[inner sep=0pt]
        \draw [draw=gray!50] (0,0) -- (1.82,1.77);
    \end{tikzpicture}} \\ \bottomrule
\end{tabular}

%% file: forest_density.tex
\begin{tabular}{c@{\;}c@{\;\;}ccccc}
    \toprule
    \multicolumn{7}{c}{\textbf{Forest Density Prediction}} \\
    \toprule
    & & \multirow{2}{*}{\textbf{Historical Map}}
    & Direct weakly-
    & Translation +
    & \multirow{2}{*}{Supervised}
    & \multirow{5}{*}{\raisebox{-1.07\height}{\def\colormapheight{7.3cm}\input{greenbar}}} \\ 
    & & &supervised & segmentation & & \\ \midrule
    \raiseandrotate{.45}{\textbf{Cassini}} &
    \raiseandrotate{.3}{1747-1815} &
    \includegraphics[width=0.2\linewidth]{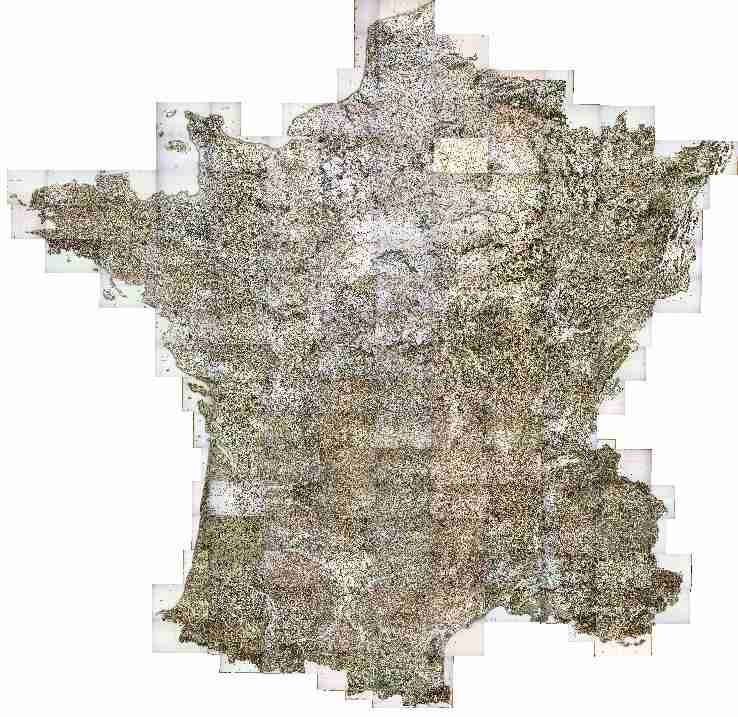} &
    \includegraphics[width=0.2\linewidth]{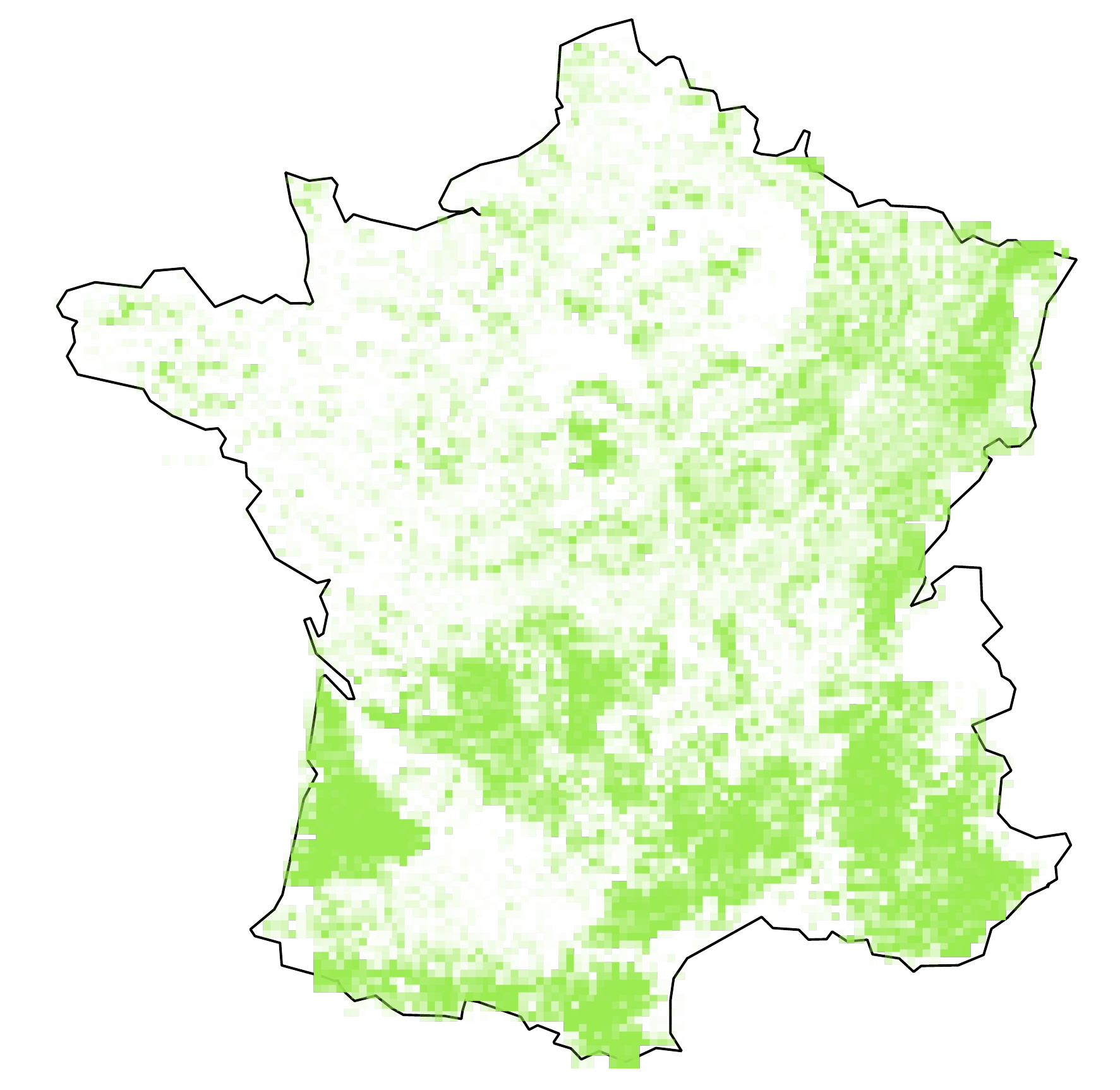} &
    \includegraphics[width=0.2\linewidth]{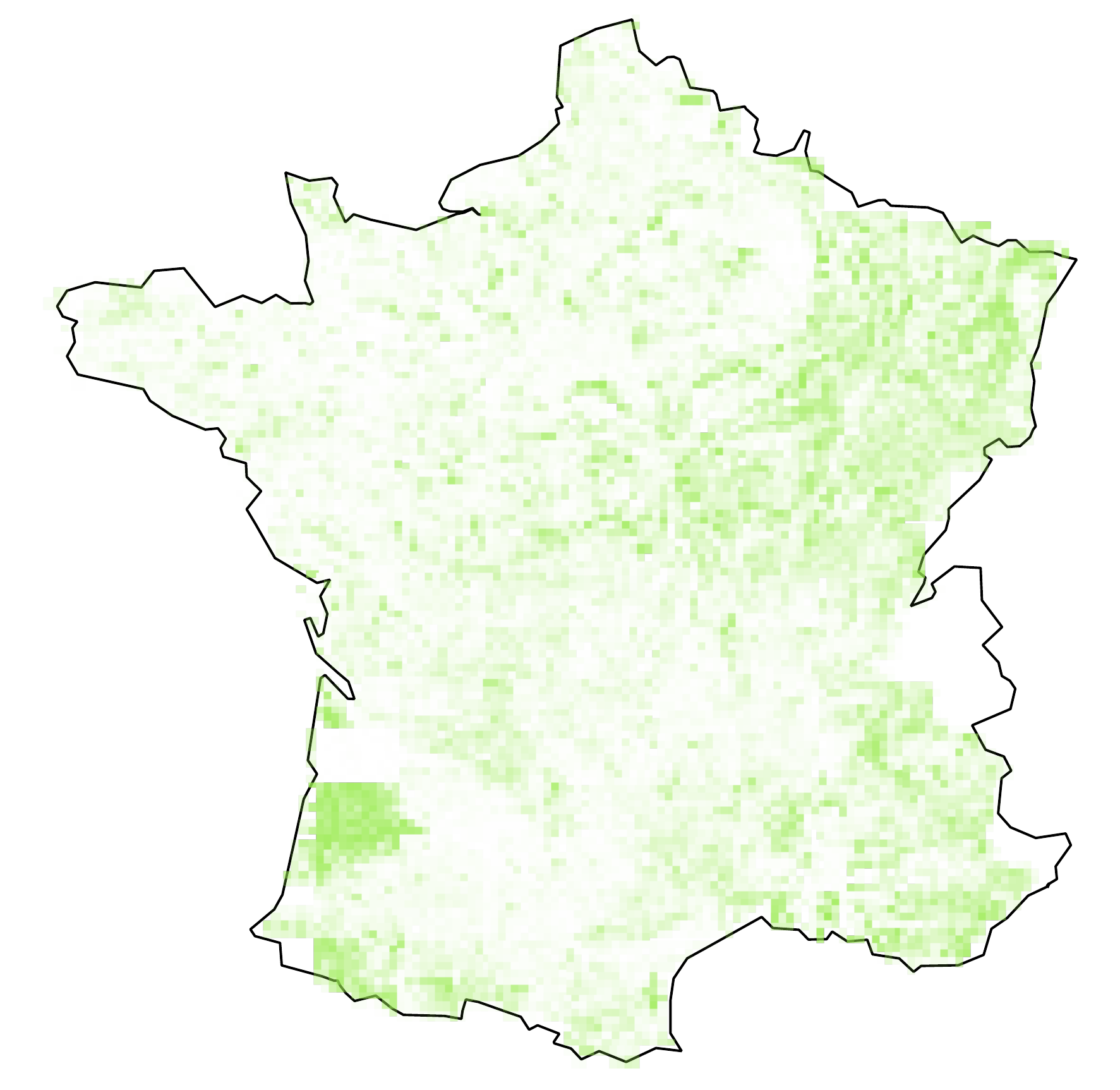} & 
    \includegraphics[width=0.2\linewidth]{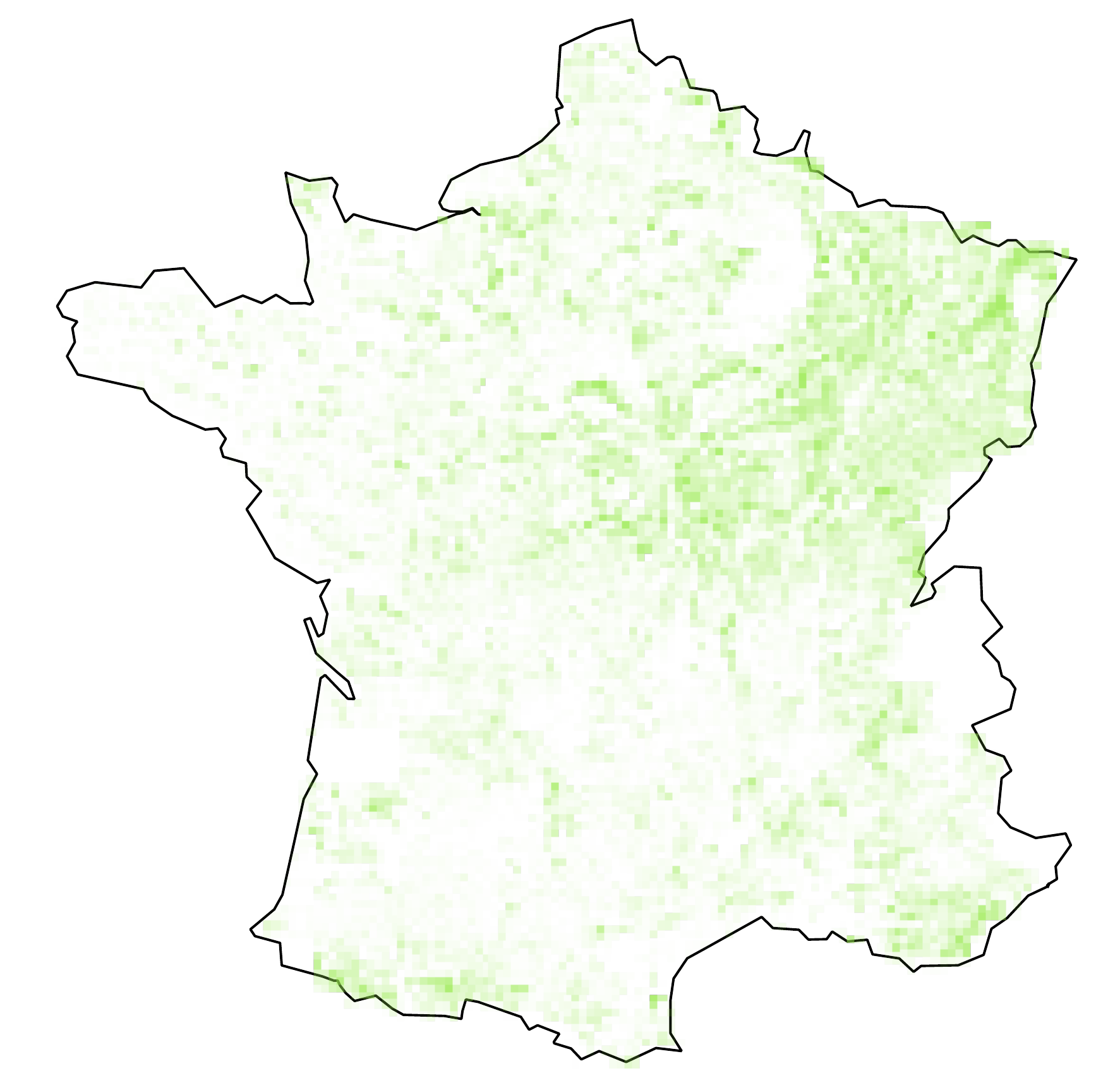} & \\
    
    \raiseandrotate{.1}{\textbf{État-Major}} &
    \raiseandrotate{.3}{1817-1866} &
    \includegraphics[width=0.2\linewidth]{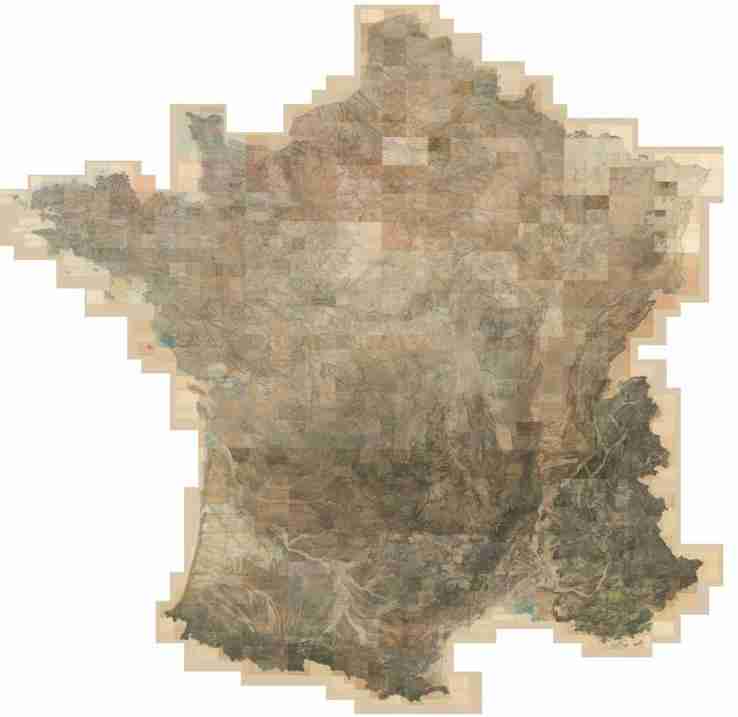} &
    \includegraphics[width=0.2\linewidth]{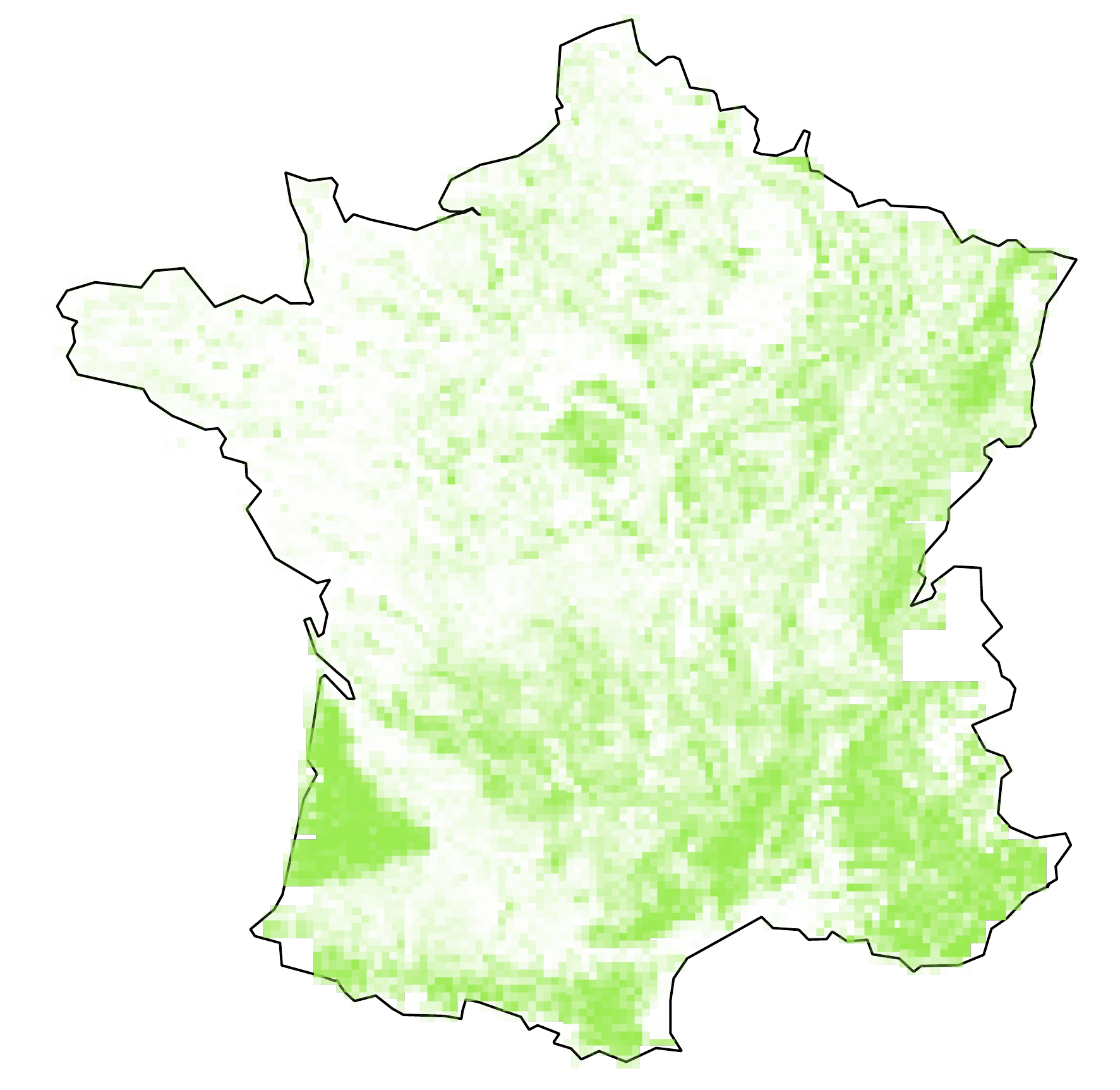} &
    \includegraphics[width=0.2\linewidth]{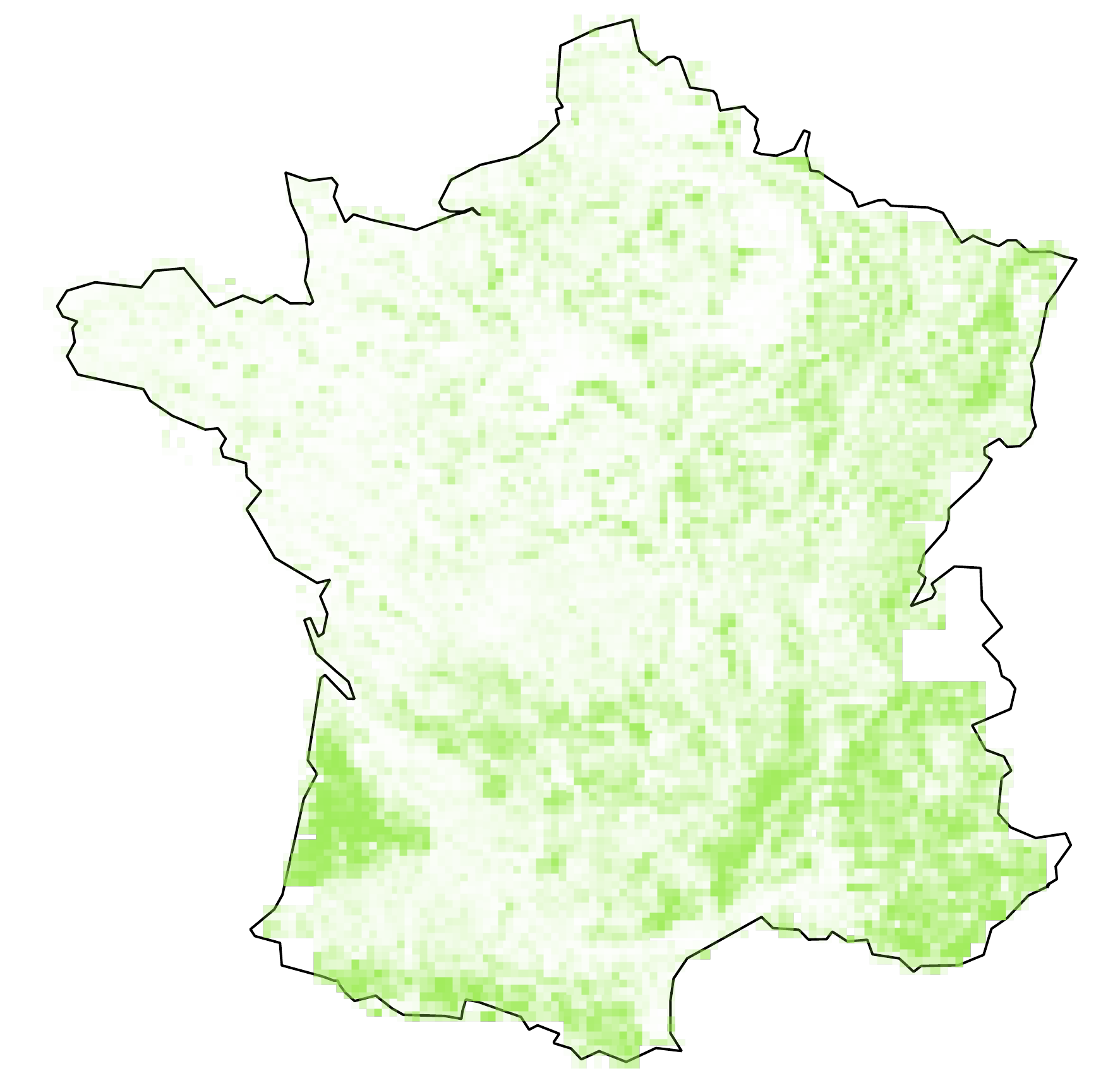} &
    \includegraphics[width=0.2\linewidth]{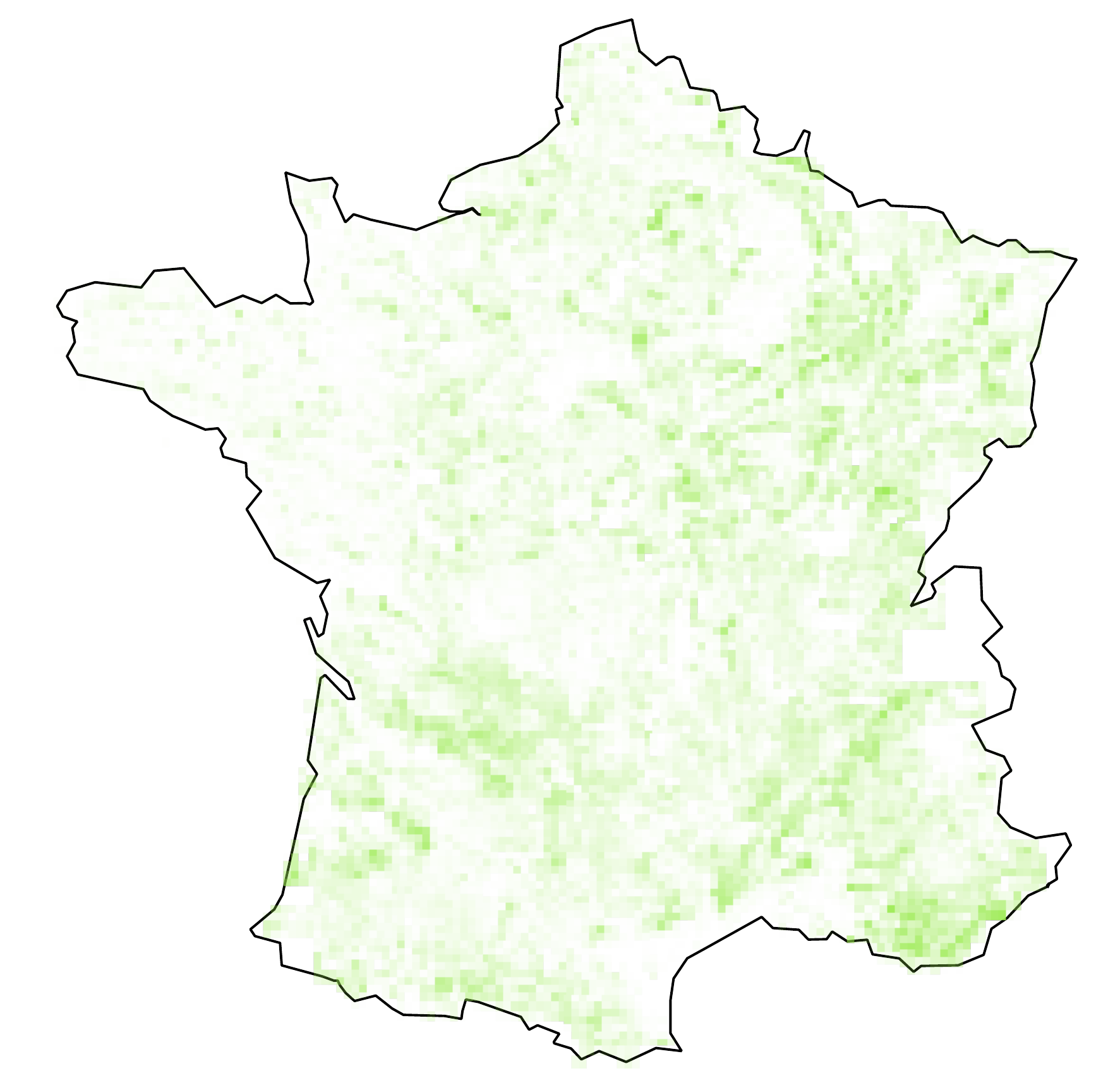} & \\
    
    \raiseandrotate{.3}{\textbf{SCAN50}} &
    \raiseandrotate{.3}{1889-1922} &
    \includegraphics[width=0.2\linewidth]{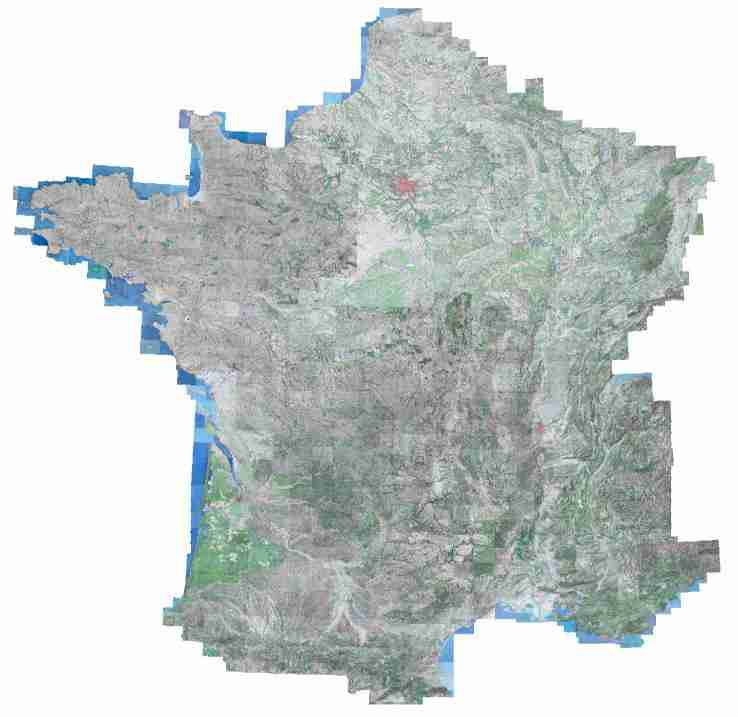} &
    \includegraphics[width=0.2\linewidth]{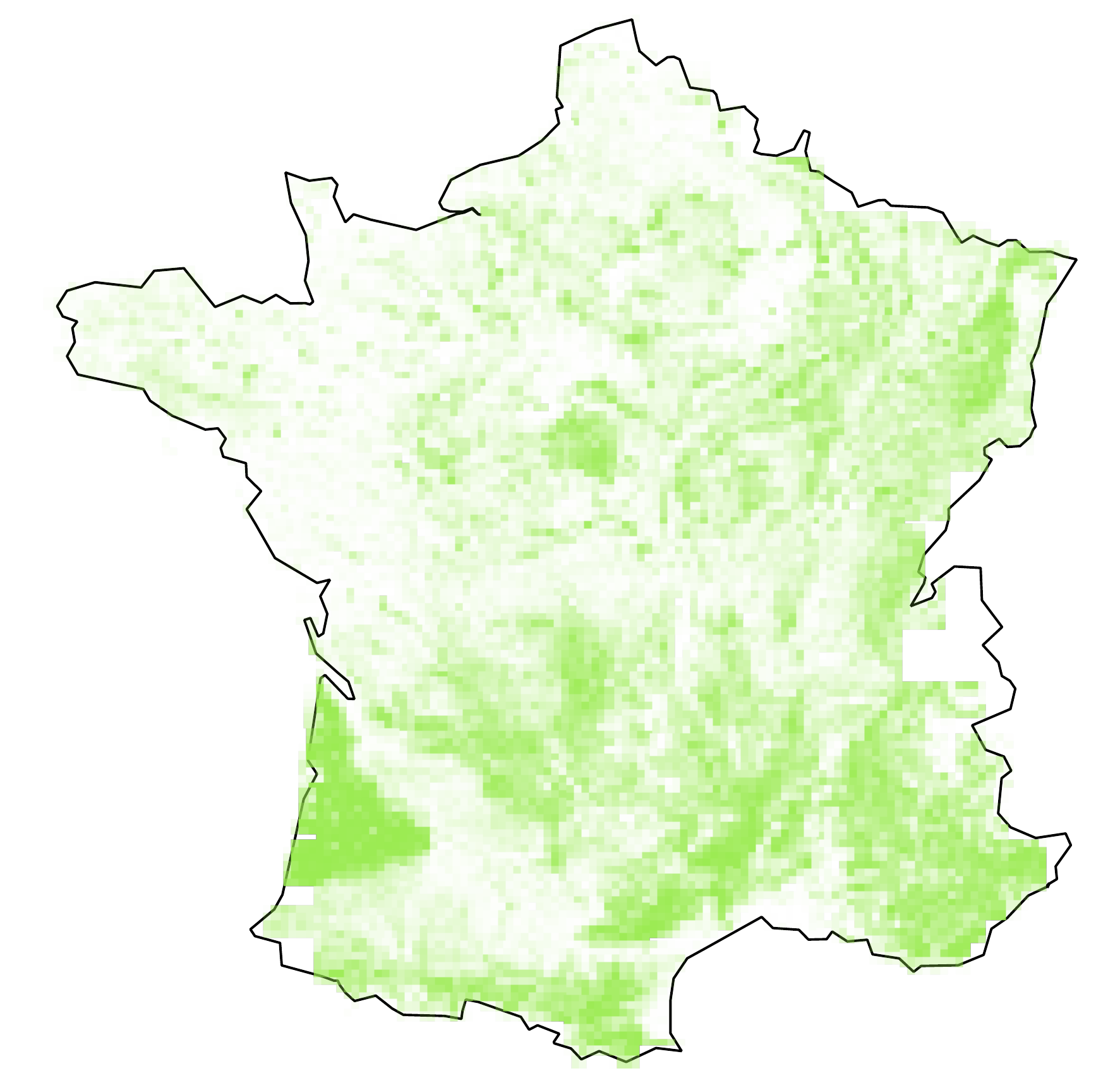} &
    \includegraphics[width=0.2\linewidth]{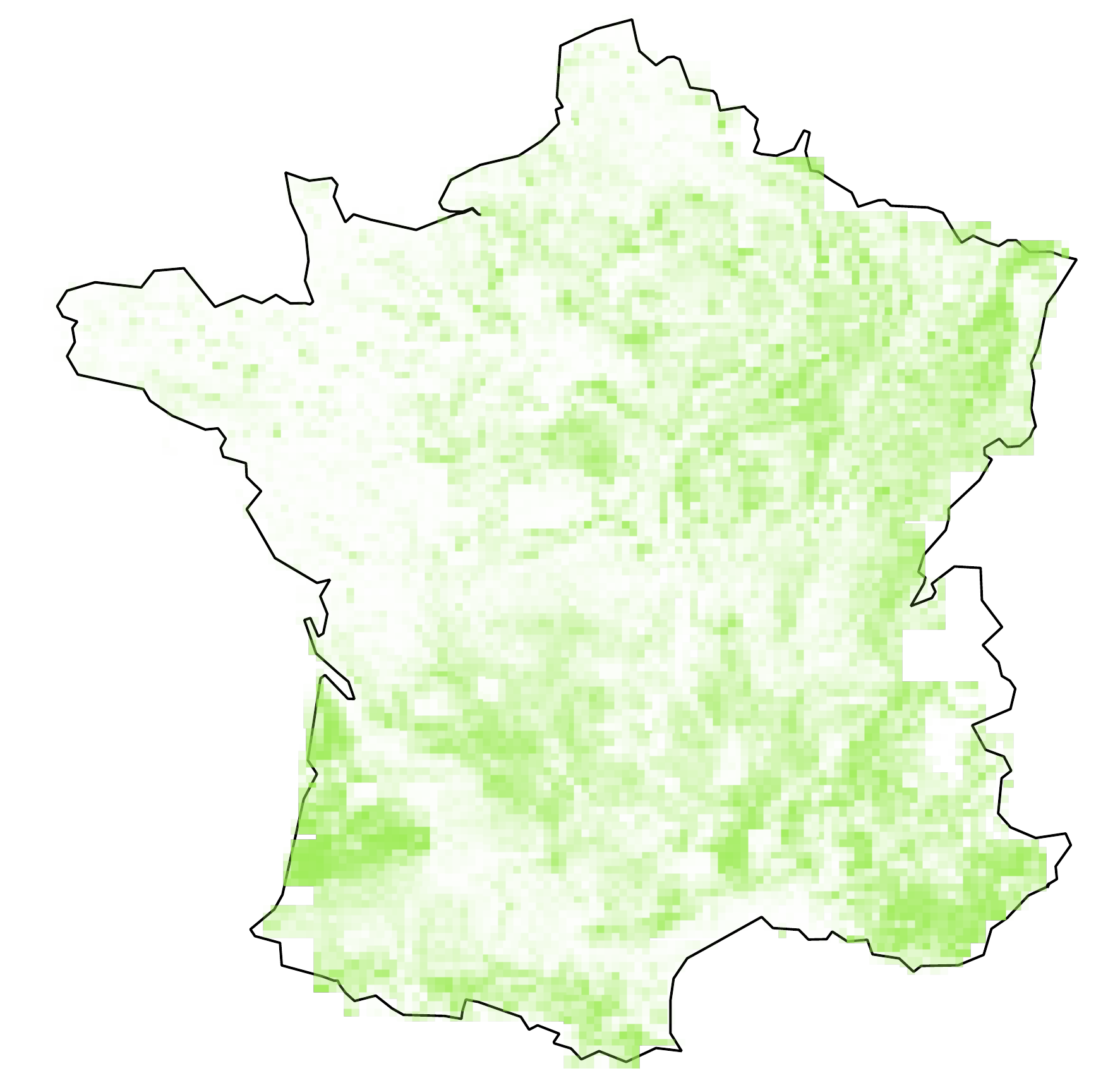} &
    \begin{tikzpicture}[framed,outer sep=0,inner sep=0]
        \node[anchor=south west,label={above:{\scriptsize Modern Map}}] (image) at (0,0) {\includegraphics[width=0.16\linewidth]{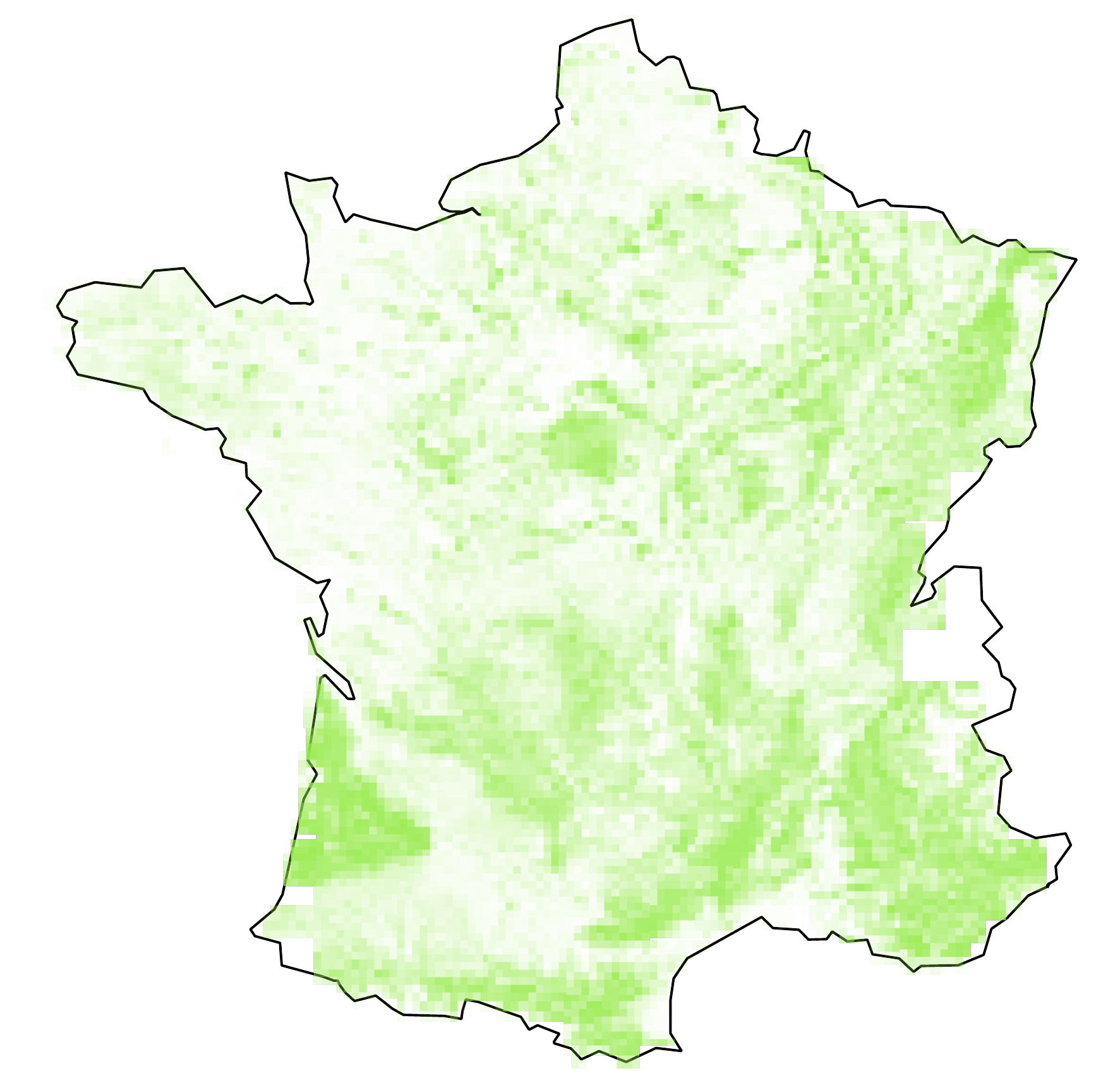}};
        \begin{scope}[x={(image.south east)},y={(image.north west)}]
            \draw[thick,draw=nice_brown] (0.3,0.26) circle (2mm);
            \draw[thick,draw=nice_orange,rotate=-8,anchor=north] (0.4,0.15) ellipse (3.9mm and 1.4mm);
            \draw[thick,draw=nice_blue] (0.55,0.3) circle (3.9mm);
            \draw[thick,draw=nice_purple,rotate=5,anchor=north] (0.85, 0.16) ellipse (2.5mm and 3.1mm);
        \end{scope}
    \end{tikzpicture} & \\ \bottomrule
\end{tabular}

%% file: greenbar.tex
%auto-ignore
\pgfplotsset{
    /pgfplots/colormap={treeheights}{
        rgb255(0cm)=(255,255,255)   % White
        rgb255(1cm)=(230,250,212)   % Light green
        rgb255(2cm)=(205,245,168)   % Medium green
        rgb255(3cm)=(180,240,125)   % Darker green
        rgb255(4cm)=(155,235, 81)   % Dark green
    }
}

\begin{tikzpicture}
    \begin{axis}[
        hide axis,
        scale only axis,
        xlabel={in m},
        colormap name=treeheights,
        colorbar,
           colorbar,
            colorbar style={
            width=.2cm,
            height=\colormapheight,
            ytick={0,2,4},
            yticklabels={0\%,50\%,100\%},
            yticklabel style={font=\tiny},
            major tick length=1.5pt, % Customize the length of the ticks
            line width=.05mm,
            grid style={draw=none} % Ensure no grid lines are drawn
        },
        point meta min=0,
        point meta max=4
    ]
    \addplot [draw=none] coordinates {(0,0)};
    \end{axis}
 
\end{tikzpicture}

%% file: 6_conclusion.tex
\section{Conclusion}

In this paper, we introduced a dataset for historical map segmentation that spans four centuries of metropolitan France. Our dataset provides extensive contemporary annotations as well as partial historical labels, enabling the evaluation of both fully- and weakly-supervised approaches. In particular, we demonstrated that translating historical maps into a modern style prior to applying a model trained solely on modern maps can significantly enhance segmentation performance. This highlights the promises of using modern labels to learn segmentation of unlabeled maps and enable long-term monitoring of landscape features, such as forests, across centuries. Furthermore, we outlined how these methods can be compared with other historical sources of information.

Since historical maps offer many crucial insights into a territory’s past and its future trajectory, we hope our contributions will inspire further research into historical map segmentation with weak annotations. By mitigating the high cost of manual labeling, such methods could advance large-scale, long-term studies of landscape change based on machine learning approaches.

%% file: 7_acknowledgments.tex
\section{Acknowledgments}

This work was supported by the European Research Council (ERC project DISCOVER, number 101076028) and by ANR project sharp ANR-23-PEIA-0008 in the context of the PEPR IA. This work was also granted access to the HPC resources of IDRIS under the allocation 2024-AD011015314 made by GENCI.

We would like to thank Ségolène Albouy, Raphaël Baena, Sonat Baltacı, Syrine Kalleli and Yannis Siglidis for their constructive manuscript feedback, as well as Gül Varol and Alexei Efros for sharing valuable insights.